\documentclass[10pt,journal,compsoc]{IEEEtran}
\ifCLASSOPTIONcompsoc
  \usepackage[nocompress]{cite}
\else
  \usepackage{cite}
\fi
\ifCLASSINFOpdf
  \usepackage[pdftex]{graphicx}
\else
\fi
\usepackage{amsmath}
\usepackage{array}

  \usepackage[caption=false,font=footnotesize,labelfont=sf,textfont=sf]{subfig}

\usepackage{multirow}
\usepackage{amssymb}
\usepackage{booktabs}
\usepackage{color}

\RequirePackage{xspace}
\makeatletter
\DeclareRobustCommand\onedot{\futurelet\@let@token\@onedot}
\def\@onedot{\ifx\@let@token.\else.\null\fi\xspace}
\def\eg{\emph{e.g}\onedot} 
\def\ie{\emph{i.e}\onedot} 
\def\etc{\emph{etc}\onedot} \def\vs{\emph{vs}\onedot}
\def\etal{\emph{et al}\onedot}

\begin{document}
%
\title{Evaluating the Generalization Ability of Super-Resolution Networks}
%
%
%
%

\author{Yihao Liu,
	Hengyuan Zhao,
	Jinjin Gu,
	Yu Qiao,~\IEEEmembership{Senior Member,~IEEE,}
	and Chao Dong
	\IEEEcompsocitemizethanks{\IEEEcompsocthanksitem Y. Liu is with Shanghai Artificial Intelligence Laboratory, Shanghai, 200232, China. He is also with Shenzhen Institute of Advanced Technology, Chinese Academy of Sciences, Shenzhen, 518055, China. E-mail: liuyihao14@mails.ucas.ac.cn. \protect
		\IEEEcompsocthanksitem H. Zhao is with National University of Singapore, 119077, Singapore. E-mail: hengyuan.z@u.nus.edu.
		\IEEEcompsocthanksitem J. Gu is with The University of Sydney, Camperdown, NSW 2006, Australia. He is also with Shanghai Artificial Intelligence Laboratory, Shanghai, 200232, China. E-mail: jinjin.gu@sydney.edu.au.
		\IEEEcompsocthanksitem  Y. Qiao is with Shanghai Artificial Intelligence Laboratory, Shanghai, 200232, China. E-mail: qiaoyu@pjlab.org.cn.
		\IEEEcompsocthanksitem C. Dong is with Shenzhen Institute of Advanced Technology, Chinese Academy of Sciences, Shenzhen, 518055, China. He is also with Shanghai Artificial Intelligence Laboratory, Shanghai, 200232, China.  \protect
		\IEEEcompsocthanksitem Corresponding author: Chao Dong. E-mail: chao.dong@siat.ac.cn.
	}
}

%
%

\markboth{Journal of \LaTeX\ Class Files,~Vol.~14, No.~8, August~2015}%
{Liu \MakeLowercase{\textit{et al.}}: Evaluating the Generalization Ability of Super-Resolution Networks}
%



\IEEEtitleabstractindextext{%
	\begin{abstract}
		Performance and generalization ability are two important aspects to evaluate the deep learning models. However, research on the generalization ability of Super-Resolution (SR) networks is currently absent. Assessing the generalization ability of deep models not only helps us to understand their intrinsic mechanisms, but also allows us to quantitatively measure their applicability boundaries, which is important for unrestricted real-world applications. To this end, we make the first attempt to propose a Generalization Assessment Index for SR networks, namely SRGA. SRGA exploits the statistical characteristics of the internal features of deep networks to measure the generalization ability. Specially, it is a non-parametric and non-learning metric. To better validate our method, we collect a patch-based image evaluation set (PIES) that includes both synthetic and real-world images, covering a wide range of degradations. With SRGA and PIES dataset, we benchmark existing SR models on the generalization ability. This work provides insights and tools for future research on model generalization in low-level vision.
	\end{abstract}
	
	}

\maketitle

\IEEEdisplaynontitleabstractindextext

%
\IEEEpeerreviewmaketitle

\IEEEraisesectionheading{\section{Introduction}\label{sec:introduction}}

%
%
%
%
\IEEEPARstart{D}{eep} learning has achieved great success in constrained environment, and we have steadily moved our attention to its generalization ability. Generalization determines whether an algorithm can work well on unseen data. However, due to the data-driven nature, deep learning can easily overfit the training data, leading to unpredictable generalization behavior. \textcolor{black}{In this paper, we address the problem of generalization in the context of image super-resolution (SR) and restoration, which are classic low-level vision problems. Conventional SR models are typically trained under known degradation types and downsampling kernels, limiting their performance severely on real-world images \cite{srresnet,ikc}. Improving the generalization ability of SR models is crucial for developing future methods. Notably, in this work, we extend the concept of the SR network beyond the classical bicubic super-resolution task to include simultaneous restoration. This broader scope allows us to investigate the generalization capability of models that perform both restoration and SR tasks, which are more relevant to real-world scenarios.}

The problem of super-resolving images with unknown degradations is also called blind SR or real SR. According to a recent blind SR survey \cite{liu2021blind}, existing methods can be categorized into four classes, including degradation specific methods \cite{realesrgan,bsrgan}, kernel estimation methods \cite{ikc,dan}, unsupervised methods \cite{cincgan,maeda2020unpaired}, and internal statistical methods \cite{zssr,kernelgan}. Remarkably, recent works based on synthetic data have made significant progress. For example, BSRGAN \cite{bsrgan} and Real-ESRGAN \cite{realesrgan} show that when we can synthesize abundant degradations, the model can be applied in a wide range of real-world scenarios. This phenomenon is also observed in blind face restoration task, like GFP-GAN \cite{wang2021towards} and GPEN \cite{yang2021gan}. They all show impressive results on some real-world images. Recently, Dropout \cite{kong2021reflash,srivastava2014dropout} is also introduced into SR networks to improve the generalization performance. 

However, here comes the problem: how to compare their generalization performance? We may also wonder: what are the failure cases? Whether extending the degradation types is a correct direction? All these require us to objectively evaluate the generalization ability. Existing works can only show some visual examples, but do not provide any feasible evaluation strategies. Without a standard evaluation metric and dataset, we cannot fairly compare different models, restricting the progress of their development.

Nevertheless, the evaluation of generalization is by no means easy! There is no specific assessment for generalization ability. In high-level vision tasks, like classification, they usually use the prediction accuracy on an unseen dataset or category as the generalization measure \cite{keskar2016large,zhang2021understanding,NeyshaburTS14,NIPS2017_10ce03a1,novak2018sensitivity}. But in low-level vision tasks, like image restoration, there are no appropriate strategies. Can we use image quality assessment (IQA), such as PSNR and NIQE \cite{mittal2012making}, to take the place of generalization assessment (GA)? The answer is NO, and there are three main reasons. First, IQA is designed to evaluate the image quality, yet image quality is not equal to generalization ability. For example, traditional interpolation or filtering methods get lower IQA values than deep models in most cases, but they have a stable performance (good generalization ability) on all kinds of data \cite{DDR}. Generalization should be a \textbf{relative} notion that is correlated with the method itself, not only the output. Second, IQA is highly sensitive to image \textit{content}, thus will have different absolute values on different images. While in image restoration, we need to give a stable measurement on unseen \textit{degradations} but not specific datasets. Third, IQA itself is not perfect. The reference-based IQA, like PSNR and SSIM, cannot be used in real-world images without ground truth, while the non-reference IQA, like NIQE and PI \cite{blau20182018}, cannot accurately evaluate the image quality. Gu et.al \cite{gu2021ntire} have proved that existing IQA methods all have low correlation rates with human subjective scores on PIPAL dataset \cite{jinjin2020pipal}. The above three issues have stopped IQA from being a qualified GA. More detailed explanations and experiments about IQA can be found in Section \ref{sec:comparing}. 

Additionally, we should note that GA is not proposed to take the place of IQA. They are two evaluation aspects, and both have great values. In general, we can first adopt IQAs to evaluate the model performance. If a model has much inferior performance than others, it is of little significance to extraly evaluate its generalization. For models with similar IQA performance, we can exploit GA to evaluate their generalization ability. This helps us to comprehensively appraise the models in a multi-dimensional way. Hence, \textit{IQA and GA are different but complementary with each other}. They each describe a different aspect of the model. Nevertheless, the reasearch on model generalization in low-level vision is rarely discussed.

To fill in the gap, \textbf{we make the first attempt to propose a complete evaluation protocol, including a GA index, a series of test datasets, and a benchmark}, which could provide a comprehensive evaluation of the generalization ability. Before presenting our method, we clarify the basic definitions of generalization in SR task, and provide some general principles for the new index design. Afterwards, we introduce the first Generalization Assessment index for SR task -- SRGA. SRGA is based on the statistical characteristics of internal features of the model, not output images. It is calculated on the test dataset, but is not sensitive to the data selection. More interestingly, it is a non-parametric and non-learning metric, which is guaranteed to have a good generalization ability itself. Moreover, the proposed SRGA does not require paired ground-truth images. \emph{Therefore, it can be used to evaluate the model generalization ability on real-world data, not just on synthetic data.} To better validate our method, we collect a patch-based image evaluation set (PIES) that includes both synthetic data and real-world images, covering a wide range of degradations. On this basis, we can benchmark existing SR models on generalization ability. The benchmarking results are mostly consistent with our common sense. For instance, BSRGAN and Real-ESRGAN have a good generalization performance on most datasets, and are superior to other blind SR methods. We also have some surprising discoveries, like SwinIR \cite{swinir} generalizes better on heavier noisy degradation, which could provide further insights on these methods. We hope our SRGA and datasets can help promote the development of blind SR methods, as well as other low-level vision problems.

\begin{figure}[htbp]
	\begin{center}
		
		\includegraphics[width=1\linewidth]{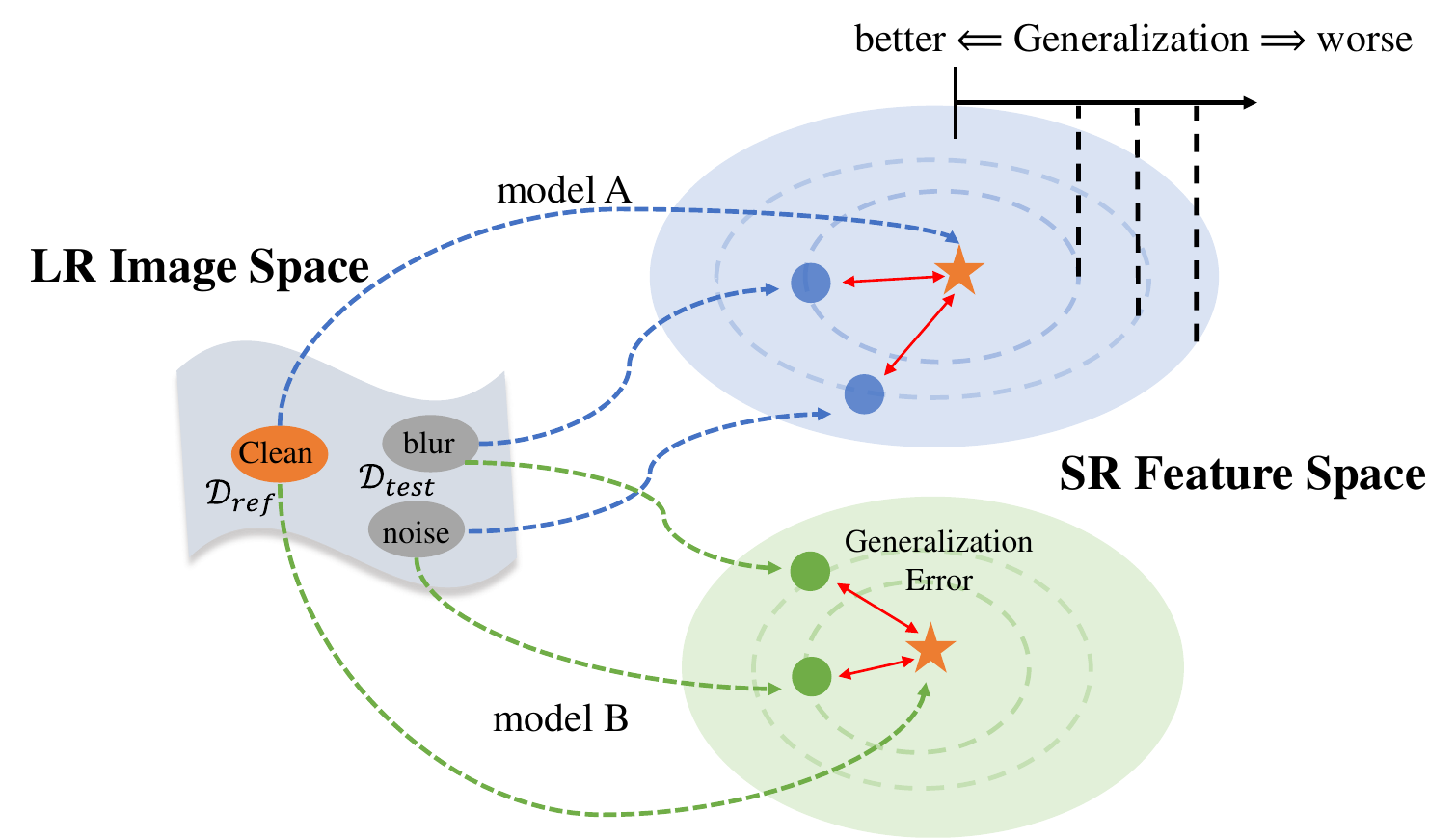}
		
	\end{center}
	\vspace{-10pt}
	\caption{Illustration of measuring the generalization ability in model internal feature space.}
	\label{fig:illustration}
\end{figure}

\section{Preliminaries}\label{sec:pre}
Designing a generalization assessment (GA) method is a completely new task for low-level vision, thus we need to reach an agreement on some basic definitions and general principles. We achieve this goal by answering common questions, which could lay the foundation of the follow-up methodology. Actually, if you find this section tedious, it is also OK to read the Experiments Section first. There you may raise some questions, and you can find the answers here.

\textbf{1. How to define the generalization ability in SR task?}
As a commonly-used definition, generalization is the ability of your model, after being trained to digest new data and make accurate predictions \cite{neyshabur2017exploring,zhou2018understanding}. There are three key components -- model, data and prediction, which all have specific meanings in SR task. (1) Model in this paper refers to the SR network. Generalization should be a property of the network, NOT output images. (2) Data refers to the input images with specific degradations. Generalization should be evaluated on degradations, NOT individual images. (3) Prediction refers to the processing effect of the output images. Generalization should measure the consistency of the processing effects across different input degradations, NOT absolute quality evaluation values. For example, an appropriate description would be: if an SR-net trained with data-A (well-performing domain) can achieve similar processing performance on data-B, then SR-net generalizes well on data-B. 

\textbf{2. How to measure the generalization ability in SR task?}
Generally, we use the generalization error to measure the generalization ability \cite{neyshabur2017exploring,zhou2018understanding}. The generalization error is easy to define in high-level vision tasks, which can be calculated by the distance between ground-truth (GT) labels and the model predictions. However, in low-level vision, there are no GT images in most cases, and one input image can correspond to multiple GT images, due to the ill-posed nature. It is hard to directly measure the restoration accuracy. It is acknowledged that deep model could achieve the best performance on in-distribution data (same distribution as the training data), but can hardly generalize to out-distribution data (different distribution from the training data). Thus, it is reasonable to define the generalization error in SR as the performance gap between the in-distribution and out-distribution data. As shown in Figure \ref{fig:illustration}, the in-/out-distribution inputs are distributed in two separate image spaces. After processing, they are expected to lie in the same space, indicating a better generalization ability. The gap between these two output spaces is the generalization error, and GA index is exactly the measurement of such a generalization error.

\textbf{3. What are the general principles for GA design?}
From the above descriptions, we can summarize four unique characteristics of GA and introduce the corresponding design principles. 
(1) GA index should be correlated with both the model and the test data. Thus, it is better to take advantage of the model, not only the output image. 
(2) \textcolor{black}{GA index should be a relative measure that quantifies the distance between the reference dataset (e.g., in-distribution data) and the test dataset (e.g., out-of-distribution data). Therefore, it is preferred to select an appropriate reference dataset beforehand. However, in certain situations, identifying a suitable in-distribution dataset can be a daunting task or even impossible. In such cases, utilizing a pairwise metric between all test data may be more suitable for constructing the GA index. This approach offers a flexible and robust way of evaluating the generalization relationship across multiple datasets.}
(3) GA index in low-level vision should be sensitive to degradation, but insensitive to image content. Thus, it is important to disentangle image degradation from content.   
(4) GA index should have good generalization ability itself. Thus, it is preferable to devise a non-learning (not rely on training data) and non-parametric (not rely on human settings) metric.

The rest of the paper is organized as follows. In Section \ref{sec:formulation}, we give the formulation of the classic image degradation model, and discuss the difference between evaluating model performance and generalization ability. The proposed SRGA index is presented in Section \ref{sec:method}. Then, in Section \ref{sec:PIES}, we describe the collected PIES dataset. In Section \ref{sec:benchmark}, we demonstrate the effectiveness of the proposed SRGA and measure the generalization ability of several representative SR models.

\section{Formulation}\label{sec:formulation}
\noindent\textbf{Model Performance.}
Given a trained SR model $G$ and a set of test input images $S_{\mathcal{D}}=\{I_n\}_{n=1}^{N}$ with degradation $\mathcal{D}$, the predicted SR results are obtained by: $I_n^{SR} = G(I_n)$.
To evaluate the performance of model $G$, we can quantify the distance between the predicted output and the ground truth (GT) image $I^{HR}$:
\begin{equation}\label{equ:performance}
Perf(G, S_{\mathcal{D}}) = \sum_{n=1}^{N}Dist(I_n^{SR}, I_n^{HR}),
\end{equation}
where $Dist(\cdot, \cdot)$ is a distance or similarity function, such as $L_2$ error, PSNR, SSIM \cite{ssim}, LPIPS \cite{lpips} or other image quality evaluation metrics. The Equ. (\ref{equ:performance}) describes the average performance of model $G$ on the test set $S_{\mathcal{D}}$. Note that the model performance is actually affected by image content and degradation simultaneously. That is, datasets with the same degradation but different image content will result in different performance scores.

\noindent\textbf{Generalization Ability.} Unlike model performance, generalization ability should characterize the consistency of the model's processing effects across different types of input data, rather than absolute performance values. A model with good generalization ability should have similar processing effects for different types of inputs.

Formally, given two different input sets $S_{\mathcal{D}_1}$ and $S_{\mathcal{D}_2}$, generalization measures the difference between the processing effects of $G$ on $S_{\mathcal{D}_1}$ and $S_{\mathcal{D}_2}$:

\begin{equation}\label{equ:generalizability}
Gen(G, S_{\mathcal{D}_1}, S_{\mathcal{D}_2}) = Dist(\mathcal{F}(G, S_{\mathcal{D}_1}), \mathcal{F}(G, S_{\mathcal{D}_2})),
\end{equation}
where $\mathcal{F}(G, S_{\mathcal{D}})$ is a function that represents the processing effct of model $G$ on dataset $S_{\mathcal{D}}$. It should concentrate more on the image degradation $\mathcal{D}$ instead of image content. In this paper, we explore the intrinsic statistics of deep features of SR networks, and propose a statistical method for evaluating the generalization ability.

\noindent\textbf{Blind Super-resolution.} Blind super-resolution aims at recovering and super-resovling an input low-resolution (LR) image with unknown degradation to a high-resolution (HR) version. Formally, a basic image degradation model is formulated as follows:

\begin{equation}\label{equ:degradation}
I^{LR} = (I^{HR} \otimes k)\downarrow_s + n,
\end{equation}
where $I^{HR}$ is the HR image, $I^{LR}$ is the degraded LR image, $\otimes$ denotes the convolution operation. There are mainly three types of degradation in this model, \ie, blur kernel $k$, additive noise $n$ and the downsampling operation $\downarrow_s$. For blind SR setting, the degradation information is unavailable during the training process. Generally, most current methods follow this degradation model or its variants \cite{ikc,dan,realesrgan}. In this paper, we also follow the basic degradation formulation as Equ. (\ref{equ:degradation}) to make the PIES dataset and validate the proposed generalization index.

\begin{figure*}[htbp]
	\begin{center}
		
		\includegraphics[width=1\textwidth]{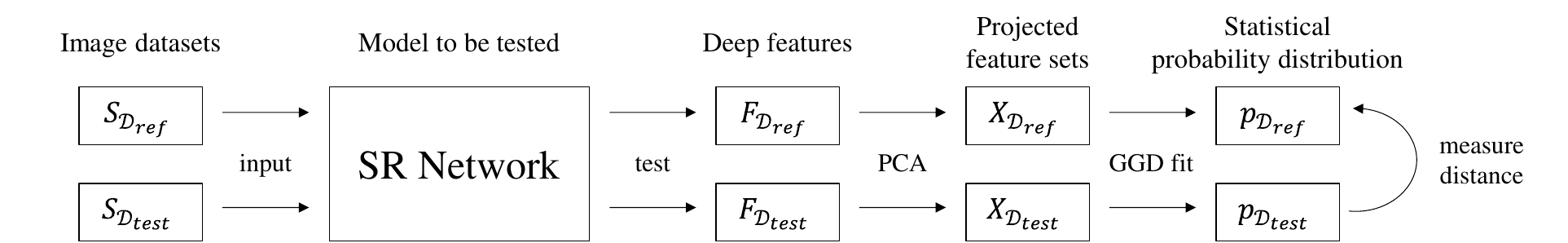}
		
	\end{center}
	\vspace{-10pt}
	\caption{Schematic pipeline for calculating the proposed generalization index SRGA with unpaired reference dataset.}
	\label{fig:overview}
	
\end{figure*}

\section{Statistical Modeling of Deep SR Networks}\label{sec:method}
\subsection{Overview of the Proposed SRGA}
The overview of the proposed generalization ability metric SRGA is depicted in Figure \ref{fig:overview}. SRGA is built upon the statistical modeling of deep features of SR networks. It is a relative measurement that computes the distance of the feature distributions between the reference dataset and the candidate test dataset. The reference dataset used in SRGA is typically the one on which the model performs well, and is usually within the training distribution. We first obtain the corresponding deep features of the input datasets. These features are then compressed using principal component analysis (PCA), and the resulting projected feature sets are modeled using a generalized Gaussian distribution. Finally, the generalization error is measured by the Kullback-Leibler divergence (KLD) between the two probability distributions, leading to the proposed generalization index SRGA.

\textcolor{black}{
	When a model performs well on a specific dataset, we can use it as a reference and compare the distance between other candidate datasets and the reference dataset using SRGA. However, if an appropriate reference dataset cannot be identified, we can treat all candidate datasets equally and calculate the SRGA score between each pair to obtain a generalization matrix. By using SRGA in this way, we can assess a model's generalization performance beyond its training set and make informed decisions about its suitability for real-world applications.
}
In the following, we will elaborate on the details of the adopted statistical methodology.

\subsection{Revisit Natural Image Statistics}
For decades, lots of efforts have been made to explore the statistics of natural images \cite{ruderman1994statistics,ruderman1994statistics2,wainwright1999scale,moorthy2010statistics}. For example, Mallat \cite{mallat2009theory} discovered that coefficients of multi-scale and orthonormal wavelet decompositions of images could be described by the generalized Gaussian model. Moulin and Liu \cite{moulin1999analysis} analyzed the multi-resolution (wavelet) image denoising schemes using generalized Gaussian and complexity priors. Among these works, Mittal \etal \cite{mittal2012no} discovered that the image mean subtracted contrast normalized (MSCN) coefficients in the spatial domain have characteristic statistical properties that are strongly correlated to the image distortion. They adopted the generalized Gaussian distribution to model the MSCN coefficients, and successfully proposed the widely-used no-reference image quality assessment (NR-IQA) BRISQUE \cite{mittal2012no} and NIQE \cite{mittal2012making}. A recent work \cite{DDR} further discovered the deep degradation representations (DDR) hidden in the SR networks: the deep features of SR networks are spontaneously discriminative to image degradations rather than image contents.

Inspired by these works, we present a new perspective for studying the generalization ability of deep models. Specifically, we explore the statistics of deep features of SR networks and build a statistical model to quantitatively measure the generalization ability.

\begin{figure*}[htbp]
	\begin{center}
		
		\includegraphics[width=1\textwidth]{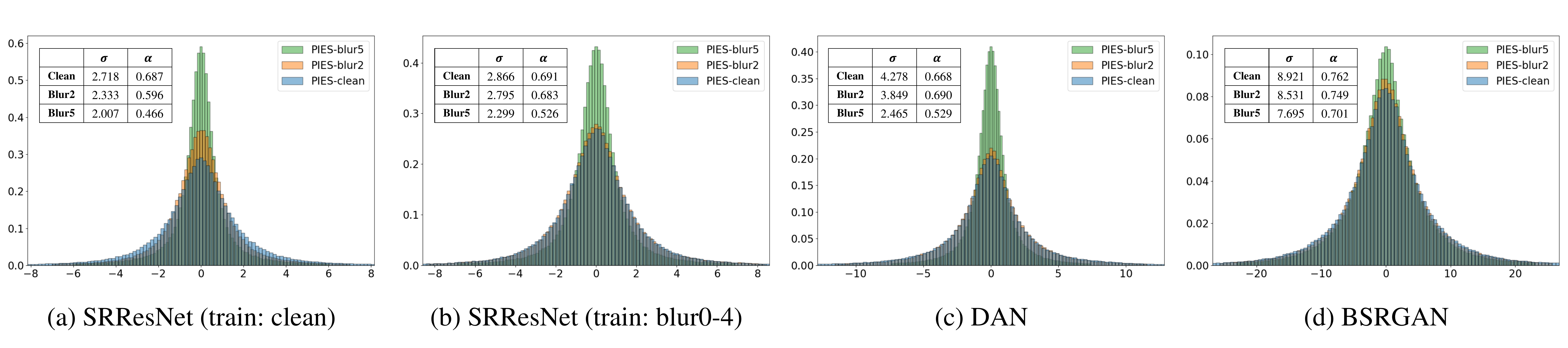}
		
	\end{center}
	\vspace{-10pt}
	\caption{Statistics of deep features of SR networks. Different distributions will lead to different parameters. The distribution differences within the training data (well generalized region) are usually small, while the distribution outside the training data deviates significantly. This potentially reveals the difference of the processing effects for various degradations. By measuring the distribution differences, we can approximately evaluate the generalization ability.}
	\label{fig:statistics}
	
\end{figure*}

\subsection{Statistics of Deep Features of SR Networks} \label{sec:Statistics}

Given a set of $N$ input images $\{I_n\}_{n=1}^{N}$ with degradation $\mathcal D$, for arbitrary super-resolution model $G$, we can obtain its corresponding deep features $\{F_n\}=\{G_{F}(I_n)\}$, where $F_n \in \mathbb{R}^{H \times W \times C}$, $H$, $W$ and $C$ are the height, width and depth, respectively. $G_{F}$ denotes the model containing all the layers before the last output layer, \ie, $F_n$ is the extracted deepest feature map before the output layer.

Firstly, we apply PCA \cite{wold1987principal} to reduce the dimension of $\rm vec(F_n)$ \footnote{Flatten the spatial feature map into a one-dimensional vector.} from $H W C$ to $D$, to alleviate the calculation cost and compress information.\footnote{In our experiments, $D$ is set to 300. Other reasonable values ($>300$) also work fine.} Specifically, $Y=\left[\rm vec(F_1); \rm vec(F_2);...;\rm vec(F_N)\right] \in \mathbb{R}^{N \times HWC}$. The PCA method will find a projection matrix $P \in \mathbb{R}^{HWC \times D}$, resulting in  $X=YP, X\in \mathbb{R}^{N \times D}$. Denote $\rm set(X) = \{X(j,k)| j \in \{1,2,...,N\}, k \in \{1,2,...,D\}\}$. The cardinal of $\rm set(X)$ is $N \times D$. Our finding is that the elements $x \in \rm set(X)$ have the statistical properties that are determined by the model $G$ and the degradation type $\mathcal D$. Formally, $x$ obeys probability distribution $p_{G,\mathcal D}$: $x \sim p_{G,\mathcal D}(x)$.

\begin{figure}[htbp]
	\begin{center}
		\includegraphics[width=0.85\linewidth]{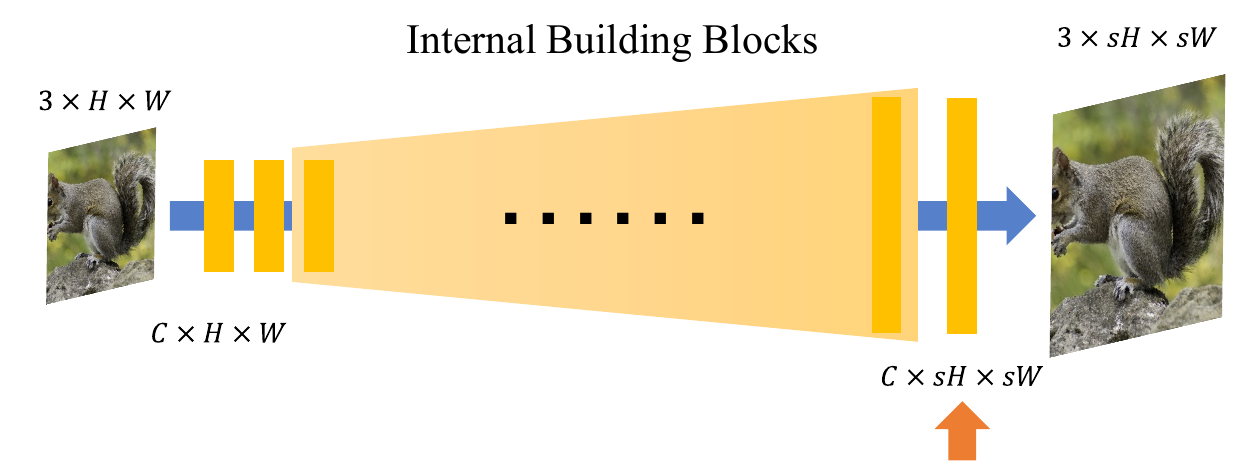}
	\end{center}
	\vspace{-10pt}
	\caption{Simplified architecture for arbitrary SR networks. We extract the deepest features (pointed by the arrow) to calculate the SRGA index.}
	\label{fig:network}
\end{figure}

Notbly, for all the SR models, we extract the output features of the penultimate layer to calculate the SRGA index. A simplified illustration is depicted in Figure \ref{fig:network}. This deepest layer contains all the processed information to produce the final results. There is no more processing layer afterwards. Thus, it is reasonable to use this layer to compute SRGA. To make an analogy, IQA adopts the three-channel output images to measure the model performance (image quality), while we utilize the penultimate multi-channel output feature maps to evaluate the generalization ability of deep models. Another reason is that different models have different architectures, it is convenient and universal to adopt the last layer.

In order to visualize how the distributions $p_{G,\mathcal D}$ vary as a function of the model $G$ and input degradation $\mathcal D$, Figure \ref{fig:statistics} plots the histograms of $x \in \rm set(X)$ with different $G$ and $\mathcal D$ on PIES dataset (described in Section \ref{sec:PIES}). From the statistical results, we can draw two important observations: 1) For a given model $G$, different input degradation types $\mathcal D_1$ and $\mathcal D_2$ will lead to different probability distributions $p_{G,\mathcal D_1}$ and $p_{G,\mathcal D_2}$. By quantifying the distance between these two resulting probability distributions, it is possible to measure the difference in the processing effect on different input distributions. 2) For degradation type $\mathcal D$, different model $G$ will produce different distributions, implying that each model has created its own feature manifold, which determines the generalization ability.

\subsection{Generalized Gaussian Distribution}
To model and analyze the statistics of the deep features of SR networks, we explore to fit the empirical distributions $p_{G,\mathcal D}$ with an appropriate statistical distribution. The Generalized Gaussian Distribution (GGD) is widely used for many relevant tasks, since its form is general and covers many common distributions like Gaussian distributions, Laplace distributions and uniform distributions. More importantly, extensive practice has demonstrated that GGD can fit natural image statistics, and the deep features extracted by SR networks inherit such properties. We discover that the statistical properties of deep features are changed by the presence of distortion. Thus, quantifying these changes can make it possible to measure the processing effect of the model on different input degradations. The GGD with zero mean is formulated as:
\begin{equation}
GGD(x;\alpha, \sigma^2) = \frac{\alpha}{2 \beta \Gamma(1/\alpha)}\rm{exp}\left(-\left(\frac{|x|}{\beta}\right)^\alpha\right),
\end{equation}
where
\begin{equation}
\beta=\sigma \sqrt{\frac{\Gamma(1/\alpha)}{\Gamma(3/\alpha)}},
\end{equation}
and $\Gamma(.)$ is the gamma function:
\begin{equation}
\Gamma(z)=\int_{0}^{\infty} t^{z-1}e^{-t}\ dt \ (z>0).
\end{equation}

There are two parameters $\alpha$ and $\sigma$ in GGD. The parameter $\alpha$ controls the ``shape'' of the distribution. $\alpha=1$ and $\alpha=2$ yield the Laplacian and the Gaussian density function, respectively. Smaller values of the shape parameter lead to more peaked distributions. Another parameter $\sigma$ controls the variance. Intuitively, the natural image contents are presented by the distribution, while the degradation changes the distribution parameters.

We try to match the empirical histograms of the feature sample values $x \in X$ with the best possible GGD pdf. Specifically, we adopt the moment-matching based method proposed in \cite{sharifi1995estimation} to estimate the parameters $\alpha$ and $\sigma$ of GGD. Table \ref{tab:GGD} summarizes the statistics of several representative models with different input degraded images. The complete results are in the supplementary materials.

\begin{table}[htbp]
	\centering
	\caption{The estimated GGD parameters of representative methods with different degraded input.}
	\begin{tabular}{c|ccccc}
		\hline
		Methods &  & Clean & Blur1 & Blur2 & Blur4 \\ \hline 
		
		\multirow{2}{*}{\begin{tabular}[c]{@{}c@{}}SRResNet\\ (train: clean)\end{tabular}} & $\sigma$ & 2.718 & 2.532 & 2.333 & 2.083 \\ 
		& $\alpha$ & 0.687 & 0.661 & 0.596 & 0.494 \\ \hline
		\multirow{2}{*}{\begin{tabular}[c]{@{}c@{}}SRResNet\\ (train: blur0-4)\end{tabular}} & $\sigma$ & 2.866 & 2.783 & 2.795 & 2.465 \\ 
		& $\alpha$ & 0.691 & 0.682 & 0.683 & 0.591 \\ \hline
		\multirow{2}{*}{SwinIR-GAN} & $\sigma$ & 5.178 & 5.140 & 5.069 & 4.929 \\ 
		& $\alpha$ & 0.740 & 0.742 & 0.742 & 0.733 \\ \hline
	\end{tabular}
	
	\label{tab:GGD}
\end{table}

\newcommand{\ftexttt}[1]{\texttt{\frenchspacing#1}}
\begin{table*}[ht]
	\caption{Description of Patch-based Image Evaluation Dataset (PIES). It includes both synthetic and real-world images with fine-grained degradation types.}
	\resizebox{\textwidth}{26mm}{
		\begin{tabular}{cccl}
			\hline
			\toprule[0.5pt]
			\textbf{Datasets} & \textbf{Has Ref?} & \textbf{Syn/Real?} & \multicolumn{1}{c}{\textbf{Description}} \\ \hline
			\textbf{PIES-Clean} & $\checkmark$ & Syn & \begin{tabular}[c]{@{}l@{}}Image patches collected from DIV2K-valid100, BSDS100, Urban100 and General100 datasets.\\ The corresponding LR patches are downsampled using matlab \ftexttt{bicubic} function.\end{tabular}  \\ \hline
			\textbf{PIES-Blur} & $\checkmark$ & Syn & \begin{tabular}[c]{@{}l@{}}Additionally apply Gaussian blur on PIES-Clean dataset. The blur kernel width is sampled in\\ \{0.5, 1.0, 1.5, 2.0, 2.5, 3.0, 3.5, 4.0, 4.5, 5.0, 5.5, 6.0, 6.5, 7.0, 7.5, 8.0\} (16 subsets in total).\end{tabular} \\ \hline
			\textbf{PIES-AnisoBlur} & $\checkmark$ & Syn & \begin{tabular}[c]{@{}l@{}}Additionally apply anisotropic Gaussian blur on PIES-Clean dataset.\\ The kernel size is $21$, the kernel width is uniformly sampled in $[0.6, 5]$ and the rotation is uniformly sampled in $[0, \pi]$.\end{tabular} \\ \hline
			\textbf{PIES-Noise} & $\checkmark$ & Syn & \begin{tabular}[c]{@{}l@{}}Add Gaussian noise on PIES-Clean dataset.\\ The noise level is sampled in \{5, 10, 15, 20, 25, 30, 35, 40, 45, 50\} (10 subsets in total).\end{tabular} \\ \hline
			\textbf{PIES-BlurNoise} & $\checkmark$ & Syn & \begin{tabular}[c]{@{}l@{}}Apply both Gaussian blur and Gaussian noise.\\ The blur kernel width is sampled in \{1, 2, 4, 6\}, and the noise level is sampled in \{10, 20, 30\} (12 subsets in total).\end{tabular}  \\ \hline
			\textbf{PIES-RealCam} & $\checkmark$ & Real & \begin{tabular}[c]{@{}l@{}}Image patches selected from RealSR dataset \cite{cai2019toward}. RealSR dataset consists of LR-HR image pairs obtained by\\ adjusting the lens of two digital single lens reflex (DSLR) cameras (Nikon D810 and Cannon 5D3).\end{tabular}  \\ \hline
			\textbf{PIES-RealLQ} & $\times$ & Real & Image patches collected from the Internet, containing real-world images of various distortion types and degrees. \\ \hline \toprule[0.5pt]
			
	\end{tabular}}
	
	\label{tab:PIES}
\end{table*}

\subsection{Measuring Model Generalization Ability}
In the previous sections, we have revealed that different input degradation types will lead to different feature distributions. Ideally, if model $G$ is robust enough and has perfect generalization ability, the outputs of different inputs should be of the same distribution or very close to each other. Hence, by quantifying the distance between these two probability distributions, we can measure the difference of processing effects between different input distributions, which reflects the model generalization performance. 

In practice, given two sets of input images with different degradation types $\{I_n\}^{\mathcal D_1}$ and $\{I_n\}^{\mathcal D_2}$, we first obtain their corresponding deep feature sets $\rm set(X^{\mathcal D_1})$ and $\rm set(X^{\mathcal D_2})$. Then, we adopt GGD to fit the datapoints to model the feature statistics: $p_{G,\mathcal D_1}(x)=GGD(x;\alpha_1, \sigma_1^2)$, $p_{G,\mathcal D_2}(x)=GGD(x;\alpha_2, \sigma_2^2)$. Once the distribution parameters have been estimated with the deep features, we can capture the distribution change by calculating the Kullback-Leibler divergence (KLD) between distributions. Fortunately, the KLD of two zero-mean Generalized Gaussian Distribution $GGD(x;\alpha_1, \sigma_1^2)$ and $GGD(x;\alpha_2, \sigma_2^2)$ has the closed-form solution \cite{xiong2019abnormality}:
\begin{equation}
\begin{split}
D_{kl} = \rm ln \frac{\alpha_1 \sigma_2 \Gamma(1/ \alpha_2) \sqrt{\Gamma(1/ \alpha_2) \Gamma(3/ \alpha_1)}}{\alpha_2 \sigma_1 \Gamma(1/ \alpha_1) \sqrt{\Gamma(1/ \alpha_1) \Gamma(3/ \alpha_2)}} \\
+ \left( \frac{\sigma_1 \sqrt{\Gamma(1/ \alpha_1) \Gamma(3/ \alpha_2)}}{\sigma_2 \sqrt{\Gamma(1/ \alpha_2) \Gamma(3/ \alpha_1)}} \right)^{\alpha_2} \frac{\Gamma(\alpha_2/\alpha_1 + 1/\alpha_1)}{\Gamma(1/ \alpha_1)}.
\end{split}
\end{equation}

Now, we have an analytical approach to quantitatively measure the difference in the output feature distributions caused by different input images. Specifically, to evaluate the generalization ability of model $G$, a reference dataset $S_{\mathcal{D}_{ref}}$ could be selected (\ie, known in-distribution data that the model can perform well), then we compute the feature distribution distance (FDD) between the reference dataset $S_{\mathcal{D}_{ref}}$ and the test dataset $S_{\mathcal{D}_{test}}$:

\begin{equation}
FDD =  D_{kl}(P(G, S_{\mathcal{D}_{ref}}), P(G, S_{\mathcal{D}_{test}})),
\end{equation}
\begin{equation}
SRGA =  \log_{10}(FDD+10^{-\delta})+\delta,\\
\end{equation}
where $P(G, S_{\mathcal{D}})$ represents the generalized Gaussian distribution fitted by the deep features of model $G$ with the input dataset $S_{\mathcal{D}}$. Smaller FDD means that $G$ has similar processing effects on these two different input degradations. $\delta$ is introduced to avoid zero point in logarithmic function and the min value is shifted to 0 by adding $\delta$ (we set $\delta=5$). The proposed SRGA is a plausible representation of $Dist$ and $\mathcal{F}$ in Equ. (\ref{equ:generalizability}).

In addition, if there are $N$ test datasets $\{S^i_{\mathcal{D}_{test}}\}$, we can calculate the mean value of the SRGA:
\begin{equation}
mSRGA = \frac{1}{N} \sum_i^N SRGA(P(G, S_{\mathcal{D}_{ref}}), P(G, S^i_{\mathcal{D}_{test}})).
\end{equation}

Notably, mSRGA describes the averaged generalization ability across multiple degradation datasets. It is actually averaging different \textbf{degradations} not the contents of the datasets. For the same degradation but different contents, SRGA is not sensitive; for different degradations, mSRGA can give a mean value.
It is worth noting that such a measurement method does not require any paired ground truth (GT) image or even the final output image. Further, even if the contents of the two sets of input images are completely different, we can still measure the difference between them. Because the probability distribution is derived from the statistics of the network deep features and is almost irrelevant to specific image content (see Section \ref{sec:comparing}). In summary, SRGA satisfies all four properties for GA design in Section \ref{sec:pre}. SRGA is correlated with both the model itself and the test data. It is a relative indicator that measures the difference of processing effects. It is not sensitive to specific image content but the degradation. Also, SRGA is a statistical approach without relying on learning process.

\textcolor{black}{
	\textbf{Generalization Matrix.} Identifying a reference dataset that accurately reflects a model's performance can be a tricky task. Fortunately, the use of SRGA offers a solution to this challenge by enabling us to evaluate a model's generalization ability without relying on a specific reference dataset. By calculating the SRGA score between candidate datasets, we can obtain an SRGA matrix that effectively illustrates the processing effects of the model on different datasets. This approach allows us to compare the generalization abilities of different datasets by examining their SRGA distances. Datasets with similar SRGA distances demonstrate similar generalization abilities, and the model is expected to perform consistently on such datasets. By leveraging the SRGA matrix, we can evaluate a model's performance in a more comprehensive manner and make informed decisions when selecting the most suitable model for a specific task. Therefore, SRGA is a valuable tool for assessing a model's generalization ability and can aid in selecting the best model for a given task.
}

\section{Patch-based Image Evaluation Set}\label{sec:PIES}
For SR task, classical test datasets include Set5 \cite{bevilacqua2012low}, Set14 \cite{zeyde2010single}, BSD100 \cite{martin2001database}, Urban100 \cite{huang2015single} and DIV2K-valid100 \cite{timofte2017ntire}. Most methods evaluate their performance on these public datasets. However, these datasets are mainly designed for evaluating the absolute model performance. As illustrated in Figure \ref{fig:IQA_content}, these datasets have severe bias on image contents, which cannot well reflect the model generalization on degradations. Further, these datasets do not provide sufficient fine-grained and continuous degradation types, making it difficult to precisely evaluate the generalization performance. Moreover, test images with large resolution require tremendous computational cost and storage to estimate the feature distributions. Therefore, we propose a new fine-grained Patch-based Image Evaluation Set (PIES). It contains a variety of test images with different degradation types and degrees, including both \textbf{synthetic} and \textbf{real-world} degradations. PIES dataset has the following characteristics: i) Patch-based. Instead of evaluating a whole image with large resolution, we focus on image patches with relatively small resolution ($128 \times 128$ for HR and $32 \times 32$ for LR, \ie $\times 4$ SR). The degradation type and degree in one patch can be considered homogeneous and spatially invariant, which can facilitate analysis. ii) Fine-grained degradation types. PIES dataset contains different types of common degradation and covers a wide range of degradation degrees. The descriptions for PIES are summarized in Table \ref{tab:PIES}. Each subset contains 800 patches. With PIES dataset, we can both evaluate the model performance and generalization ability on a unified platform.

Notably, researchers can also define their own test dataset and reference dataset according to the practical needs. \textbf{The proposed SRGA index is not restricted to specific datasets and does not require paired GT images. Thus, it can be applied in real-world images.}

\begin{figure*}[htbp]
	\begin{center}
		\begin{minipage}[t]{0.325\linewidth}
			\centering
			\includegraphics[width=0.92\linewidth,height=0.65\linewidth]{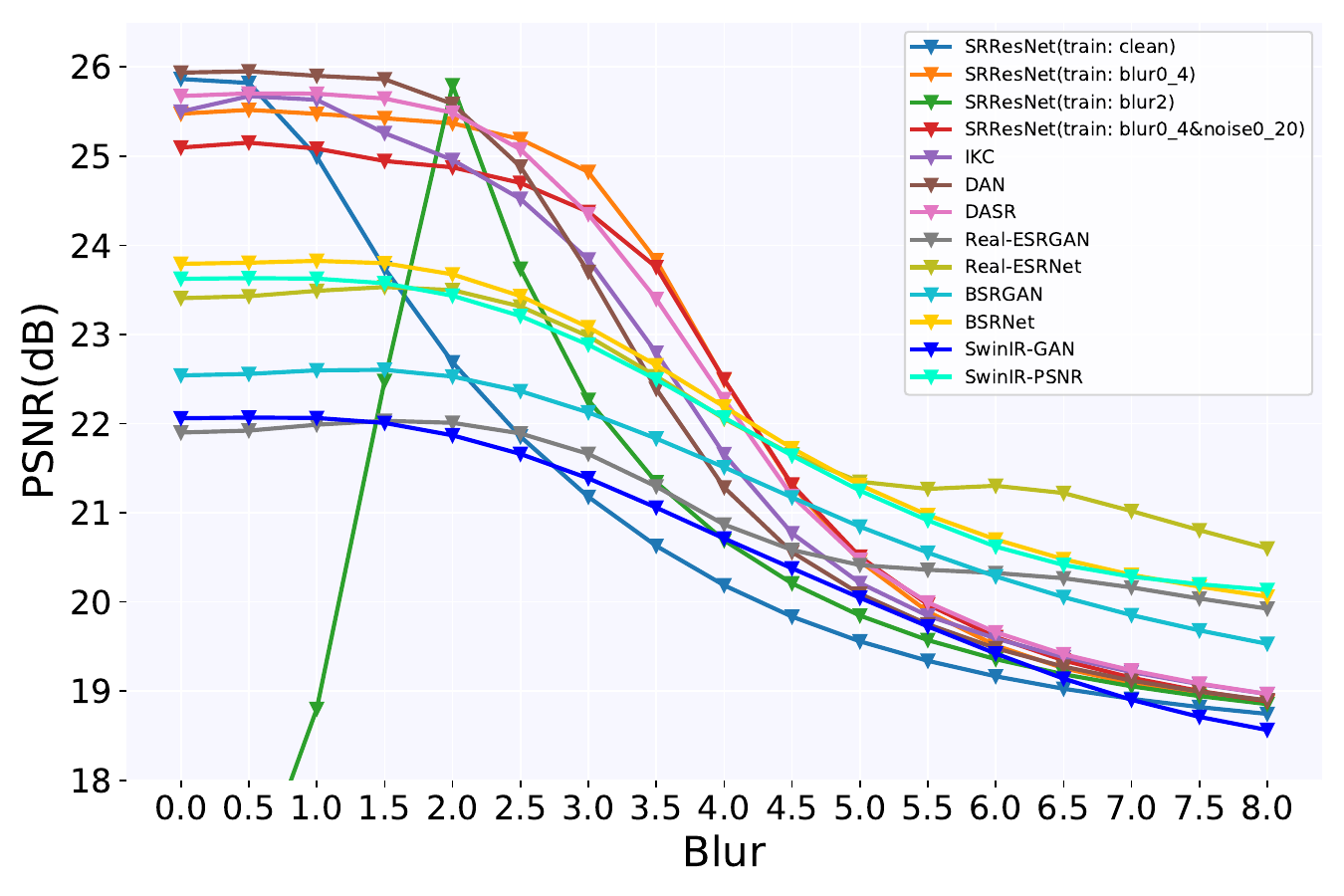}\\
			\scriptsize (a)
		\end{minipage}
		\begin{minipage}[t]{0.325\linewidth}
			\centering
			\includegraphics[width=0.92\linewidth,height=0.65\linewidth]{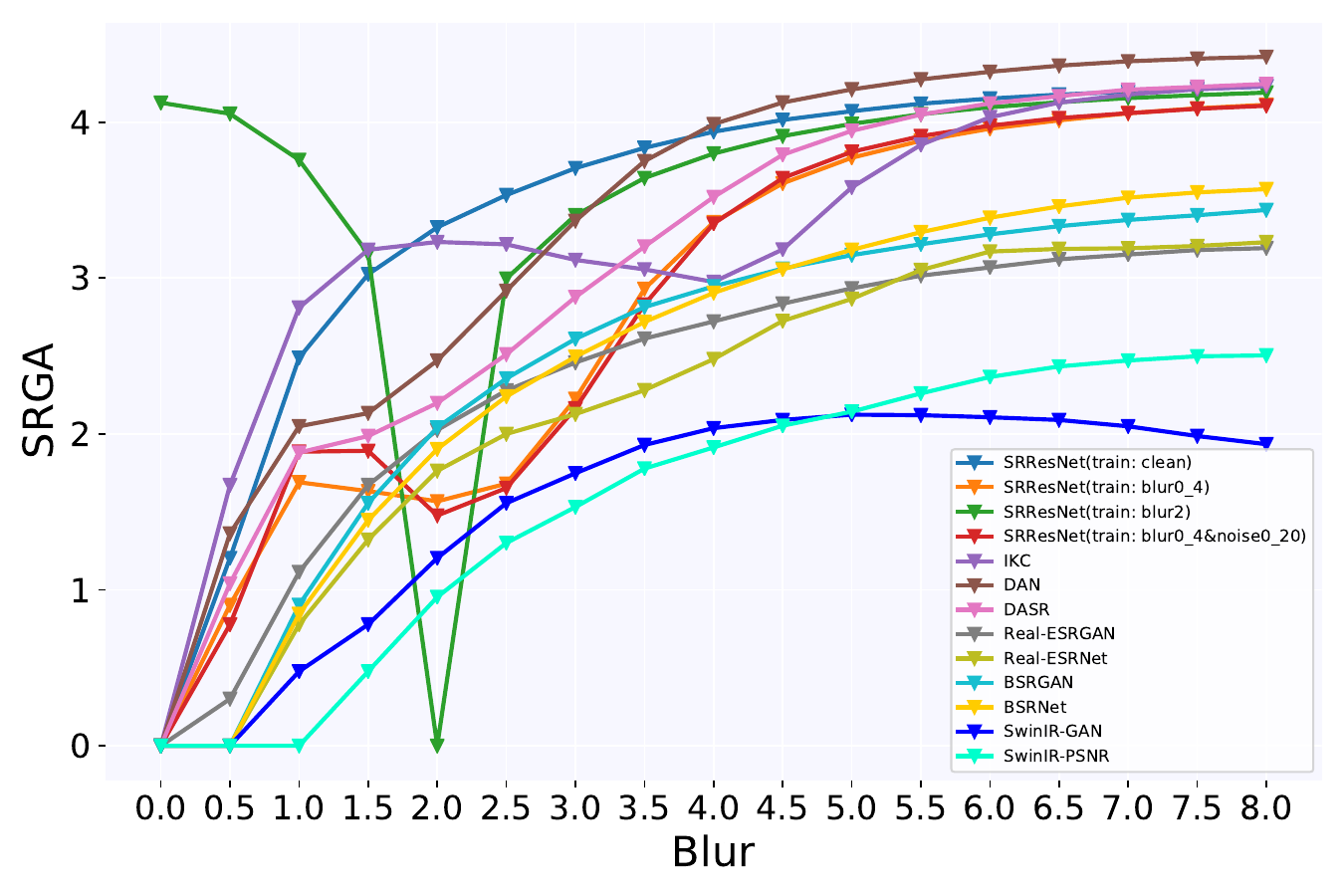}\\
			\scriptsize (b)
		\end{minipage}
		\begin{minipage}[t]{0.325\linewidth}
			\centering
			\includegraphics[width=0.92\linewidth,height=0.65\linewidth]{figures/Blur_NIQE}\\
			\scriptsize (c)
		\end{minipage}
		
		\begin{minipage}[t]{0.325\linewidth}
			\centering
			\includegraphics[width=0.92\linewidth,height=0.65\linewidth]{figures/Noise_PSNR}\\
			\scriptsize (d)
		\end{minipage}
		\begin{minipage}[t]{0.325\linewidth}
			\centering
			\includegraphics[width=0.92\linewidth,height=0.65\linewidth]{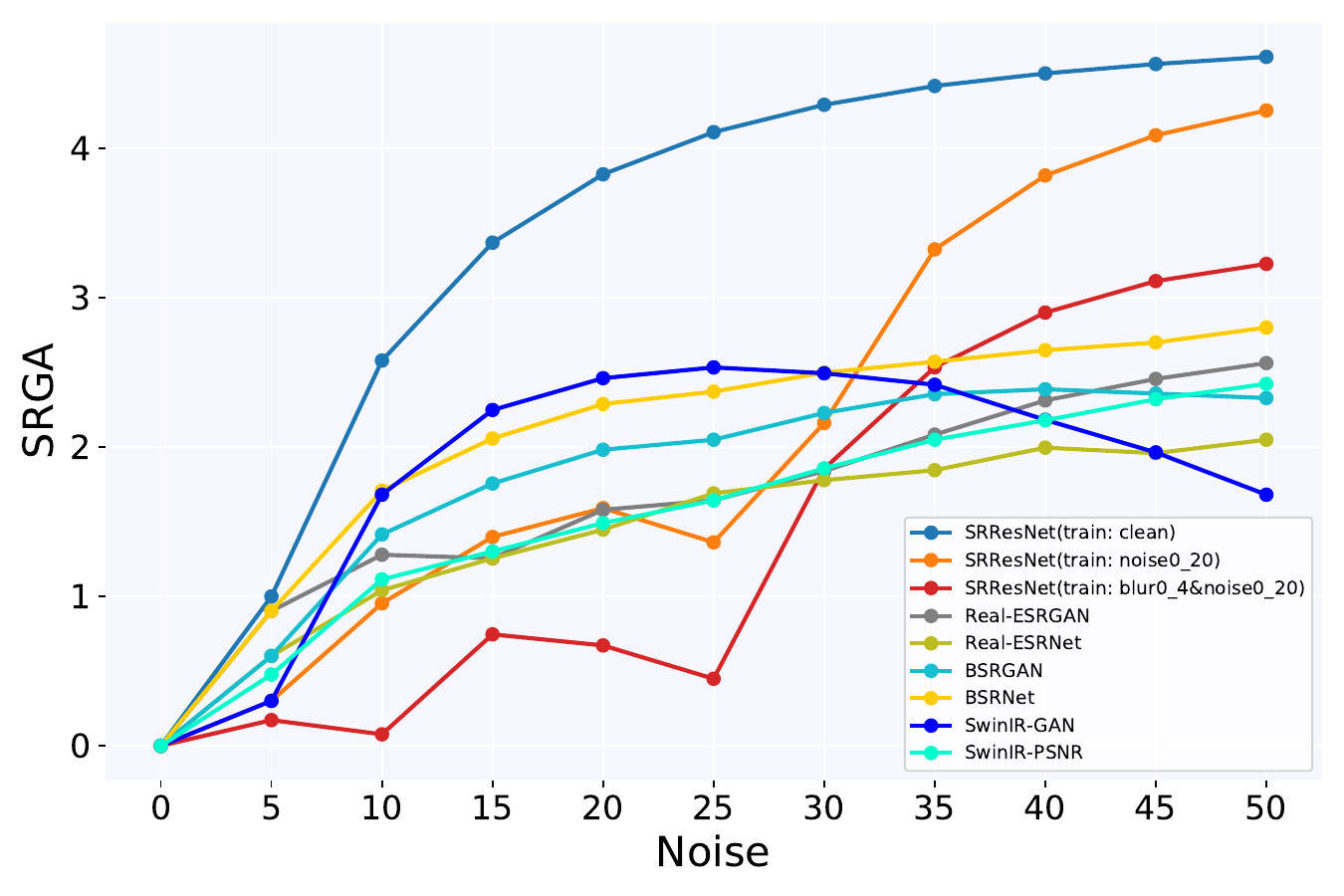}\\
			\scriptsize (e)
		\end{minipage}
		\begin{minipage}[t]{0.325\linewidth}
			\centering
			\includegraphics[width=0.92\linewidth,height=0.65\linewidth]{figures/Noise_NIQE}\\
			\scriptsize (f)
		\end{minipage}

	\end{center}
	\caption{The PSNR curves (a)\&(d), SRGA curves (b)\&(e) and NIQE curves (c)\&(f) of different models on blur and noise degradations, respectively. SRGA successfully quantifies the model generalization ability. Further, it can reflect more precise information about the model generalization ability. The detailed quantitative values are listed in the supplementary file.}
	
	\label{fig:blur_PSNR}
	\label{fig:blur_KL}
\end{figure*}

\section{Experiments}\label{sec:benchmark}
In this paper, we select representative SR methods to benchmark their generalization ability with PIES dataset, including SRResNet \cite{srresnet}, IKC \cite{ikc}, DAN \cite{dan}, DASR \cite{dasr}, Real-ESRGAN \cite{realesrgan}, BSRGAN \cite{bsrgan} and SwinIR \cite{swinir}. We train SRResNet with different training data as baselines. For other methods, we directly adopt their released pre-trained models. IKC, DAN and DASR methods mainly focus on blur degradation, thus their released model cannot deal with noisy images. Real-ESRGAN, BSRGAN and SwinIR-GAN (GAN version of SwinIR) are trained with multiple synthetic complex degraded data. We conduct experiments on $\times 4$ SR. Since the blind SR methods are supposed to perform well on clean input images, we select PIES-Clean dataset as the reference dataset, and then calculate the SRGA index between the reference and other test datasets. Note again that SRGA is not limited to synthetic datasets, but is applicable to real-world datasets. A smaller SRGA value suggests that the model can well generalize to the test dataset.\footnote{Note that model like SRResNet (train: blur2) performs well on blur2 images, thus, the reference dataset should be selected as PIES-Blur2 instead of PIES-Clean.}

In this section, we first conduct validation experiments to verify the correctness of the proposed SRGA index, and benchmark the generalization ability of existing models. Then, we compare SRGA with IQA as generalization ability measurements. Due to the space limit, the complete quantitative and qualitative results are included in the supplementary file.

\begin{figure*}[htbp]
	\begin{center}
		\begin{minipage}[t]{0.325\linewidth}
			\centering
			\includegraphics[width=0.95\linewidth]{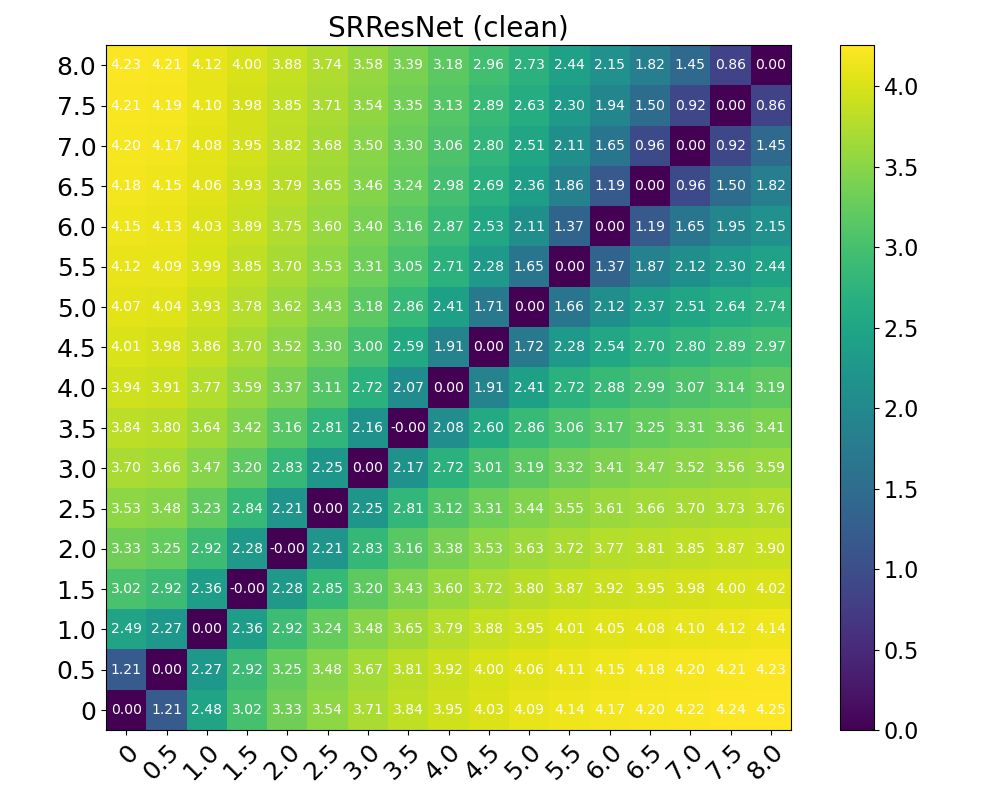}\\
			\scriptsize (a)
		\end{minipage}
		\begin{minipage}[t]{0.325\linewidth}
			\centering
			\includegraphics[width=0.95\linewidth]{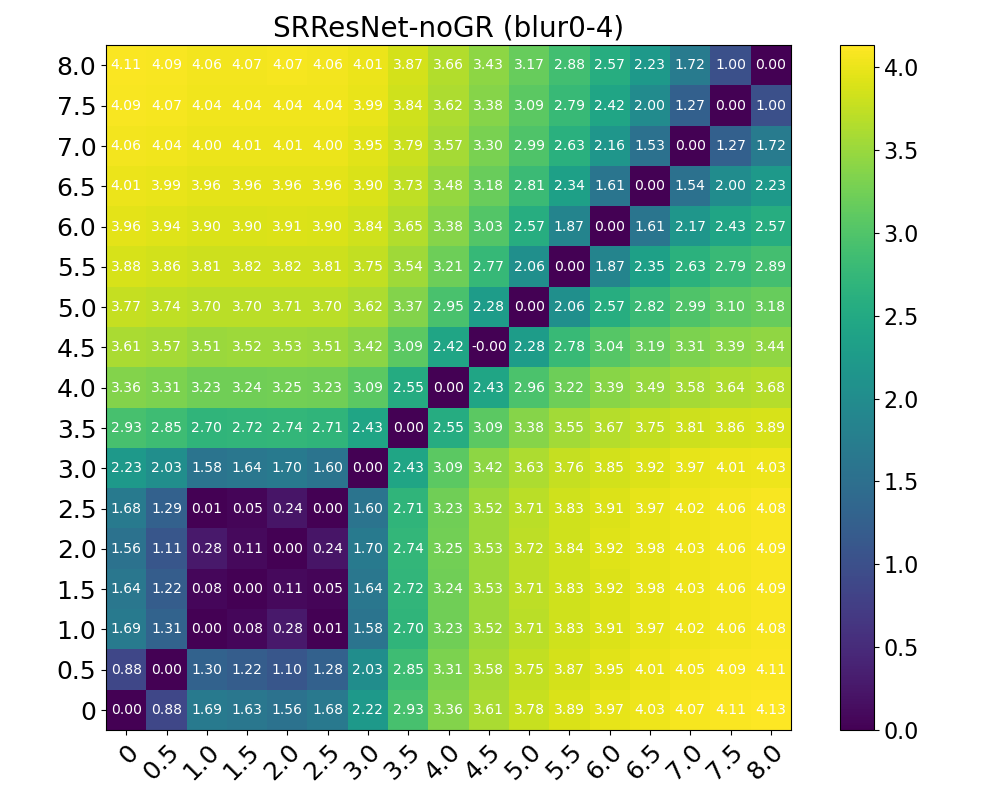}\\
			\scriptsize (b)
		\end{minipage}
		\begin{minipage}[t]{0.325\linewidth}
			\centering
			\includegraphics[width=0.95\linewidth]{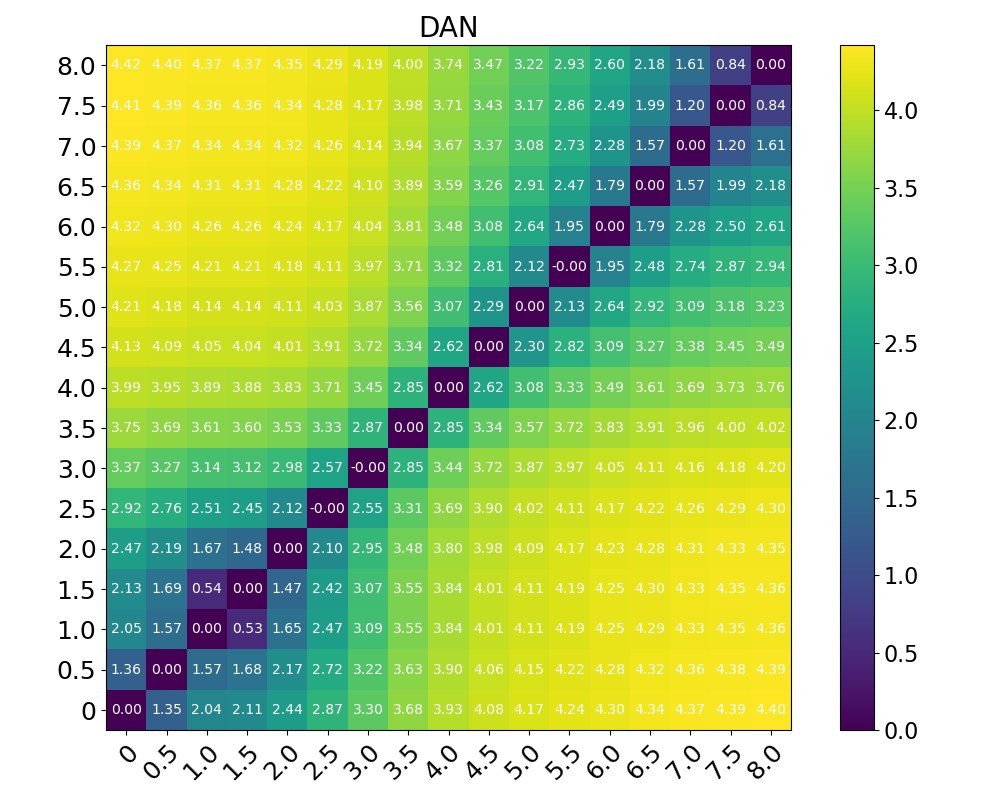}\\
			\scriptsize (c)
		\end{minipage}
		
		\begin{minipage}[t]{0.325\linewidth}
			\centering
			\includegraphics[width=0.95\linewidth]{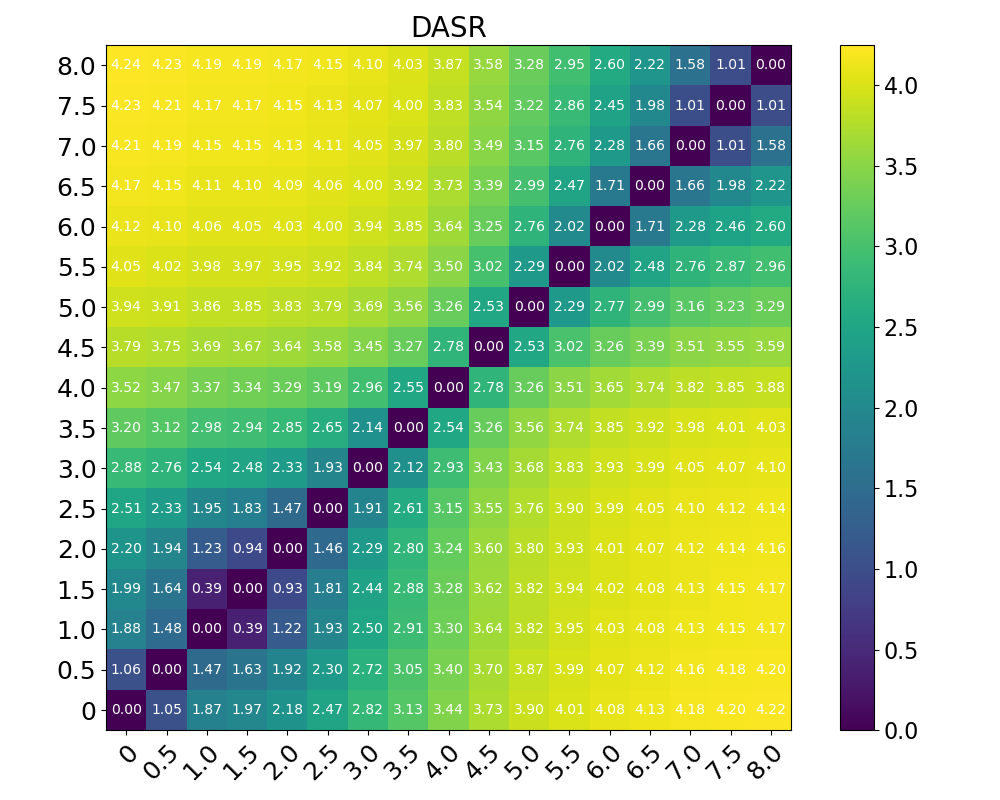}\\
			\scriptsize (d)
		\end{minipage}
		\begin{minipage}[t]{0.325\linewidth}
			\centering
			\includegraphics[width=0.95\linewidth]{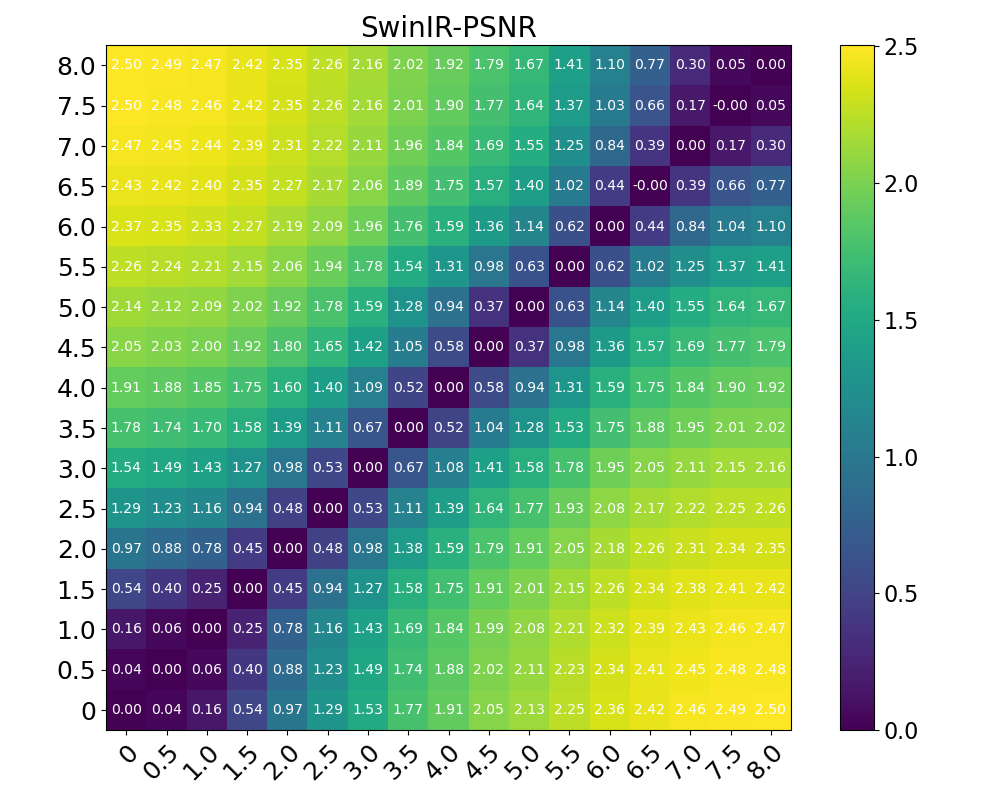}\\
			\scriptsize (e)
		\end{minipage}
		\begin{minipage}[t]{0.325\linewidth}
			\centering
			\includegraphics[width=0.95\linewidth]{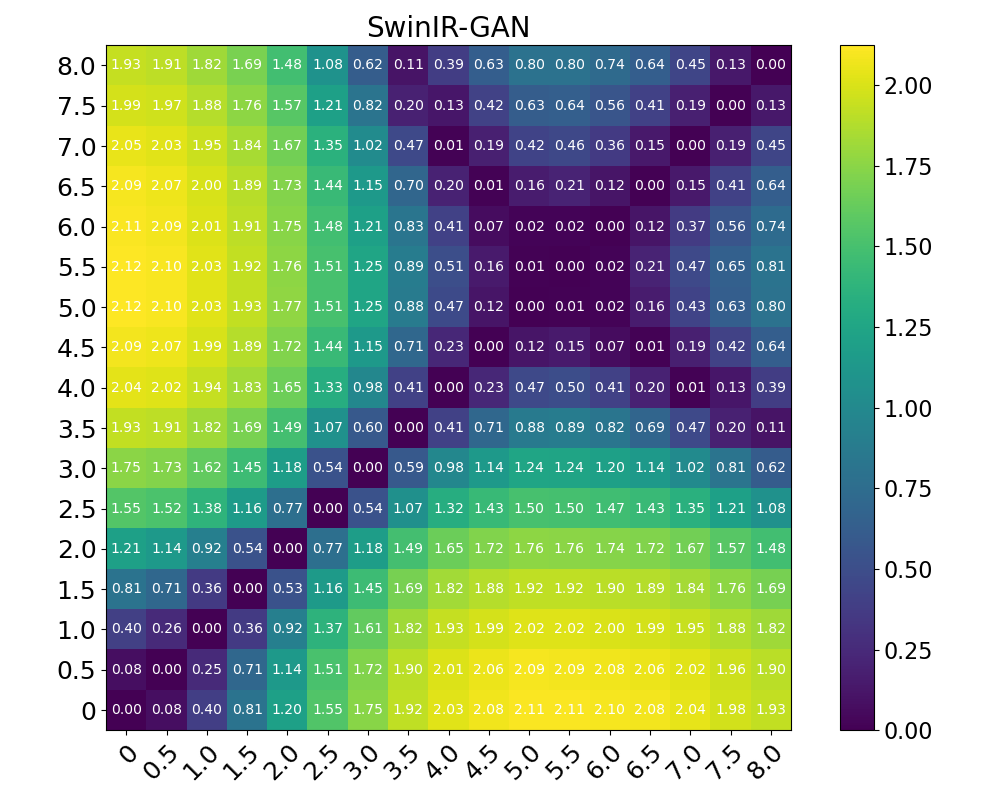}\\
			\scriptsize (f)
		\end{minipage}

	\end{center}
	\caption{\textcolor{black}{The SRGA matrices on blur degradation of (a) SRResNet (train: clean), (b) SRResNet (train: blur0-4), (c) DAN, (d) DASR, (e) SwinIR-PSNR, (f) SwinIR-GAN.}}
	
	\label{fig:SRGA_matrix_blur}
\end{figure*}

\subsection{Sanity Check for SRGA}
As there is no ground truth or human label for generalization ability, it is hard to evaluate the effectiveness of GA index directly. Nevertheless, we can construct some cases for sanity check.  \textit{Case 1}: The baseline model SRResNet trained only with clean LR data is supposed to have poor generalization ability to handle blur images. The PSNR value of this model reaches the highest on the clean input data, and then decreases rapidly as the blur degree increases (Figure \ref{fig:blur_PSNR}{(a)}). After involving blur LR data into training, SRResNet (train: blur0-4) is eligible to restore blur images: the PSNR value decreases slowly especially in the range of blur level $[0, 3]$. In Figure \ref{fig:blur_PSNR}{(b)}, we observe that the SRGA curves have the same trend with the PSNR curves in such cases. For example, the SRGA curve of SRResNet (train: clean) is steadily increasing, while the SRGA curve of SRResNet (train: blur0-4) grows slower and its values are all lower than that of SRResNet (train: clean). Especially in blur range $[0, 3]$, the SRGA values of SRResNet (train: blur0-4) only change a little within a small range. The SRGA curves successfully depict the model generalization ability in a quantitative manner. Similar results are also observed in baseline model SRResNet (train: clean) and SRResNet (train: noise0-20) (see Figure \ref{fig:blur_PSNR}{(d)}\&{(e)}). \textit{Case 2}: In another extreme case, we train a SRResNet model with only blur2 data. The SRGA curve can also successfully reflect the trend of generalization. \textit{Case 3}: SRResNet (train: clean) and SRResNet (train: noise0-20) have similar poor generalization trend on blur degradation; but SRResNet (train: noise0-20) has better generalization than SRResNet (train: clean) on noise degradation (see Table \ref{tab:benchmark}). The SRGA results are consistent with our common sense. In principal, SRGA is theoretically and experimentally sound. It provides us with a tool to quantitatively measure the generalization, which is more reliable than intuitive judgment.

\renewcommand\arraystretch{1.2} 
\begin{table*}[htbp]
	\centering
	\caption{Average results (mSRGA) of generalization ability of representative SR networks. The null value indicates that the method cannot handle this type of degradation. The reference dataset is excluded when calculating the average. PIES-Blur, PIES-Noise and PIES-BlurNoise each contain multiple subdatasets. Since models may perform inconsistently on different degrees of degraded data, the average value may not reflect the actual situation. We recommend to refer to the detailed results in the supplementary file. ($0\sim2$: well generalized; $2\sim3$: mediocre; $>3$: poor)}
	\begin{tabular}{|l|c|c|c|c|c|c|}
		\hline
		
		\multicolumn{1}{|c|}{\textbf{Methods}}         & \textbf{PIES-Blur} & \textbf{PIES-AnisoBlur} & \textbf{PIES-Noise} & \textbf{PIES-BlurNoise} & \textbf{PIES-RealCam} & \textbf{PIES-RealLQ} \\ \hline
		\textbf{SRResNet (train: clean)}                & $3.639_{\ (13)}$    & $3.615_{\ (13)}$   & $3.727_{\ (9)}$      & $3.454_{\ (10)}$           & $3.563_{\ (11)}$         & $3.825_{\ (8)}$        \\ \hline
		\textbf{SRResNet (train: blur0-4)}              & $2.967_{\ (7)}$    &  $2.880_{\ (8)}$  & -          & $3.366_{\ (9)}$           & $3.125_{\ (6)}$         & $3.811_{\ (6)}$        \\ \hline
		\textbf{SRResNet (train: noise0-20)}           & $3.623_{\ (12)}$     &   $3.605_{\ (12)}$   & $2.325_{\ (8)}$       & $3.167_{\ (8)}$           & $3.547_{\ (10)}$         & $3.827_{\ (9)}$        \\ \hline
		\textbf{SRResNet (train: blur0-4\&noise0-20)} & $2.978_{\ (8)}$   &  $2.820_{\ (7)}$   & $1.574_{\ (2)}$       & $2.901_{\ (6)}$           & $2.566_{\ (1)}$         & $3.736_{\ (3)}$        \\ \hline
		\textbf{IKC}                                   & $3.416_{\ (10)}$   & $3.375_{\ (11)}$    & -          & -              & $3.432_{\ (9)}$         & $3.845_{\ (10)}$        \\ \hline
		\textbf{DAN}                                   & $3.534_{\ (11)}$   &  $3.310_{\ (10)}$  & -          & -              & $3.937_{\ (13)}$         & $4.070_{\ (12)}$        \\ \hline
		\textbf{DASR}                                  & $3.248_{\ (9)}$  &  $3.157_{\ (9)}$    & -          & -              & $3.770_{\ (12)}$         & $4.033_{\ (11)}$        \\ \hline
		\textbf{Real-ESRGAN}                           & $2.480_{\ (4)}$   & $2.500_{\ (6)}$   & $1.791_{\ (4)}$       & $2.692_{\ (3)}$           & $3.301_{\ (7)}$         & $3.823_{\ (7)}$        \\ \hline
		\textbf{Real-ESRNet}                           & $2.336_{\ (3)}$   &  $2.456_{\ (5)}$   & $1.566_{\ (1)}$       & $2.900_{\ (5)}$           & $2.787_{\ (3)}$         & $3.770_{\ (4)}$        \\ \hline
		\textbf{BSRGAN}                                & $2.592_{\ (5)}$    & $2.397_{\ (4)}$   & $1.946_{\ (5)}$       & $2.850_{\ (4)}$           & $2.872_{\ (5)}$         & $3.796_{\ (5)}$        \\ \hline
		\textbf{BSRNet}                                & $2.598_{\ (6)}$  & $2.339_{\ (3)}$     & $2.254_{\ (7)}$       & $2.964_{\ (7)}$           & $2.686_{\ (2)}$         & $4.345_{\ (13)}$        \\ \hline
		\textbf{SwinIR-GAN}                            & $1.639_{\ (1)}$   & $1.852_{\ (2)}$    & $1.996_{\ (6)}$       & $2.435_{\ (2)}$           & $3.379_{\ (8)}$         & $3.662_{\ (2)}$        \\ \hline
		\textbf{SwinIR-PSNR}                           & $1.668_{\ (2)}$    & $1.727_{\ (1)}$   & $1.685_{\ (3)}$       & $2.297_{\ (1)}$           & $2.826_{\ (4)}$         & $3.655_{\ (1)}$        \\ \hline
	\end{tabular}

	\label{tab:benchmark}
\end{table*}

\begin{figure*}[htbp]
	\begin{center}
		\includegraphics[width=0.95\linewidth]{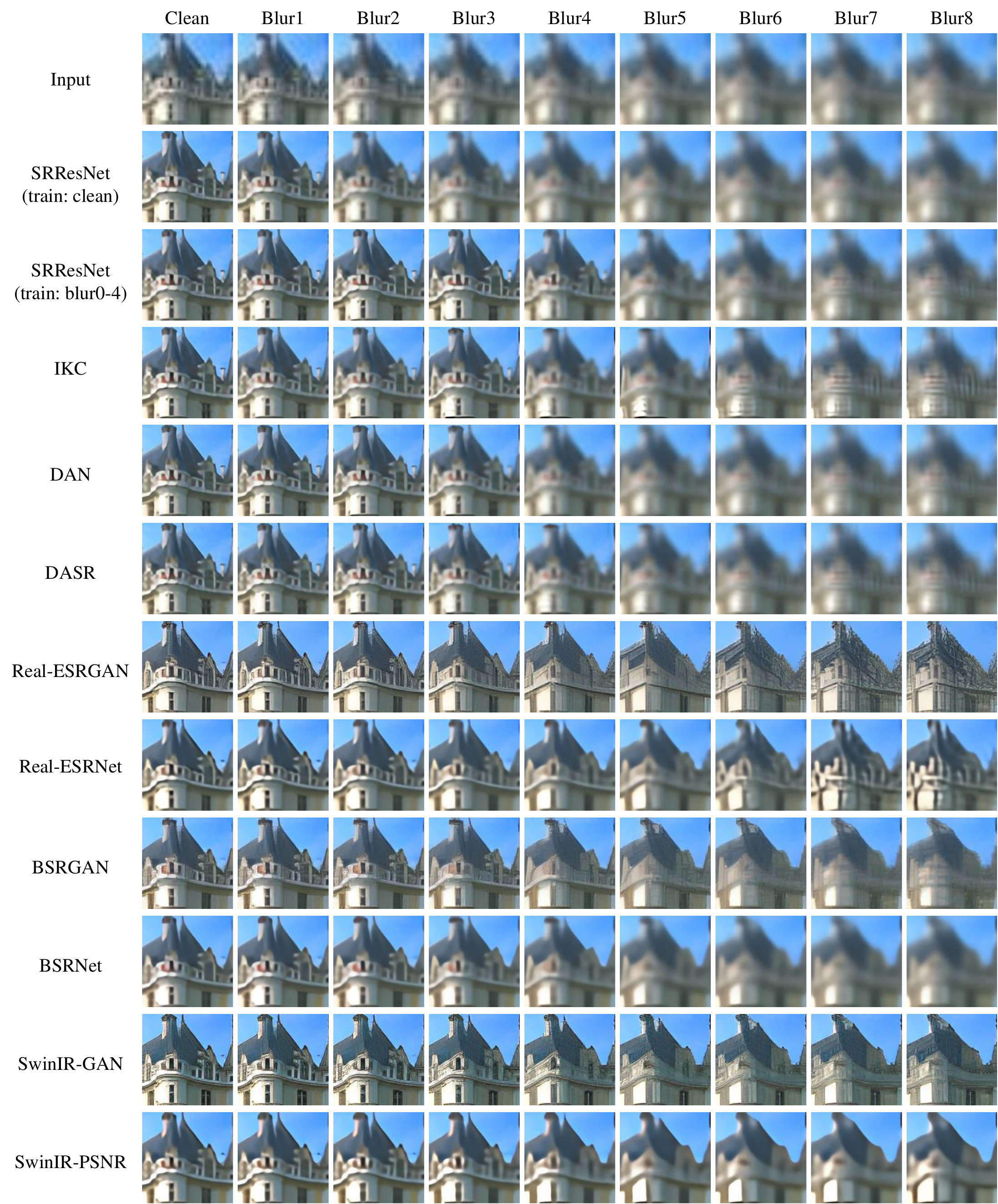}
	\end{center}
	\caption{Visual results of different models on PIES-Blur dataset.}
	\label{fig:blur_img1}
\end{figure*}

\begin{figure*}[htbp]
	\begin{center}
		\centering
		\includegraphics[width=0.95\textwidth]{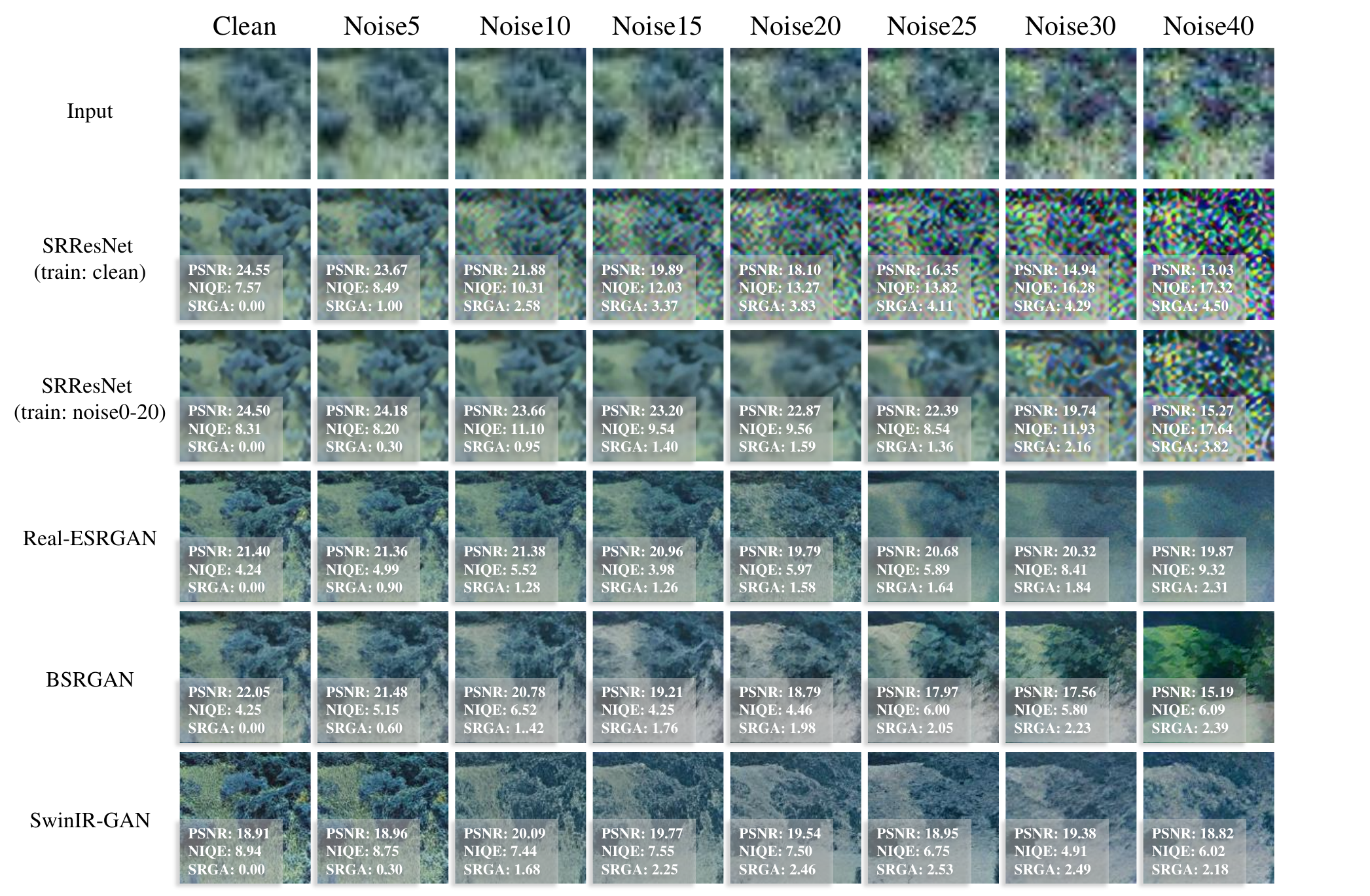}
		
	\end{center}
	\caption{Visual results of different models on PIES-Noise dataset. Note that PSNR and NIQE evaluate the \textbf{individual image quality}, while SRGA measures the model generalization (the consistency of processing effect) across different \textbf{degraded datasets}.}
	\label{fig:visual_noise2_sub}
	
\end{figure*}

\begin{figure*}[htbp]
	\begin{center}
		\includegraphics[width=0.9\linewidth,height=11cm]{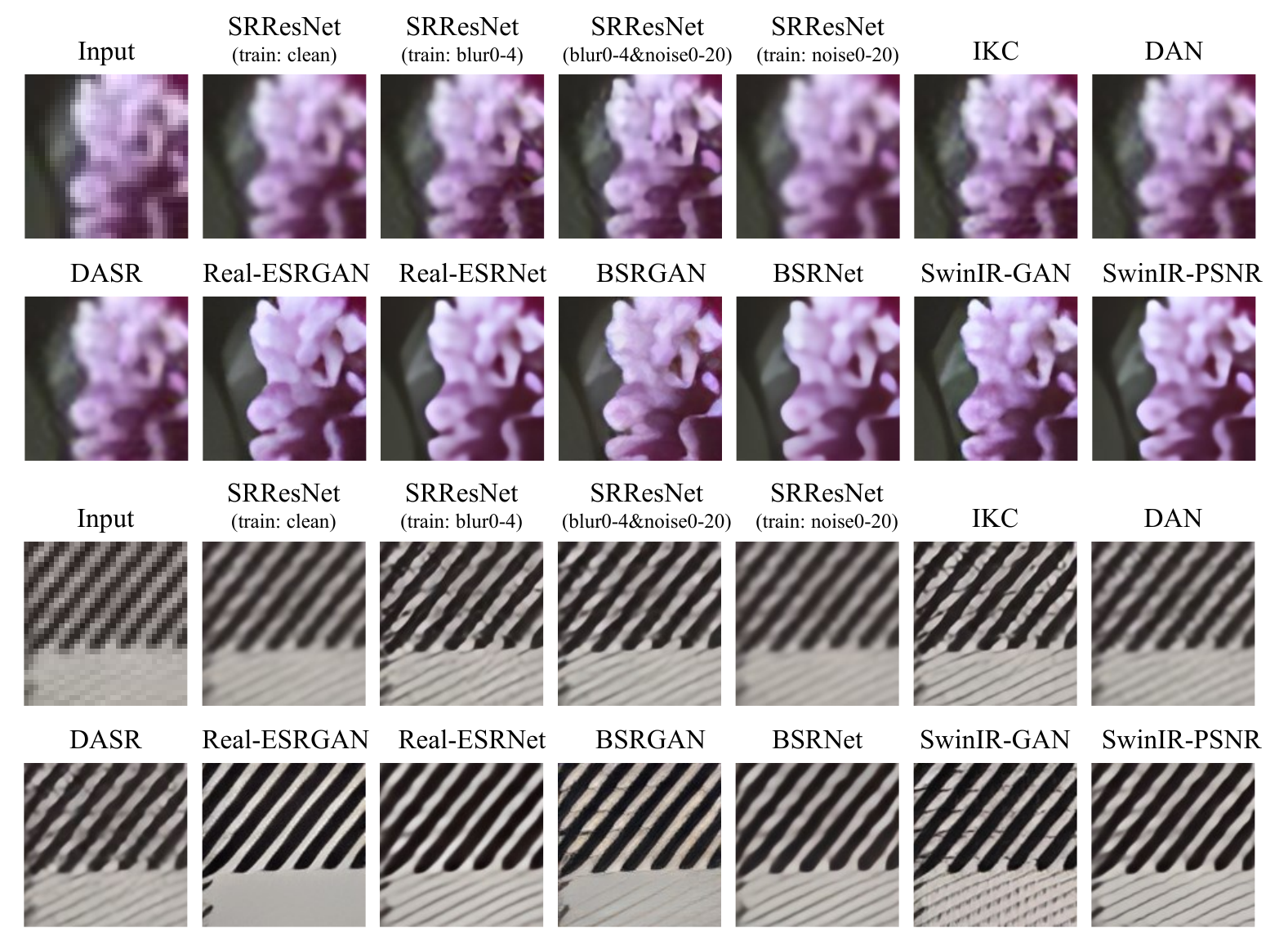}
	\end{center}
	\caption{Visual results of different models on PIES-RealCam dataset.}
	\label{fig:realcam_img}
\end{figure*}

\begin{figure*}[htbp]
	\begin{center}
		\includegraphics[width=0.9\linewidth,height=11cm]{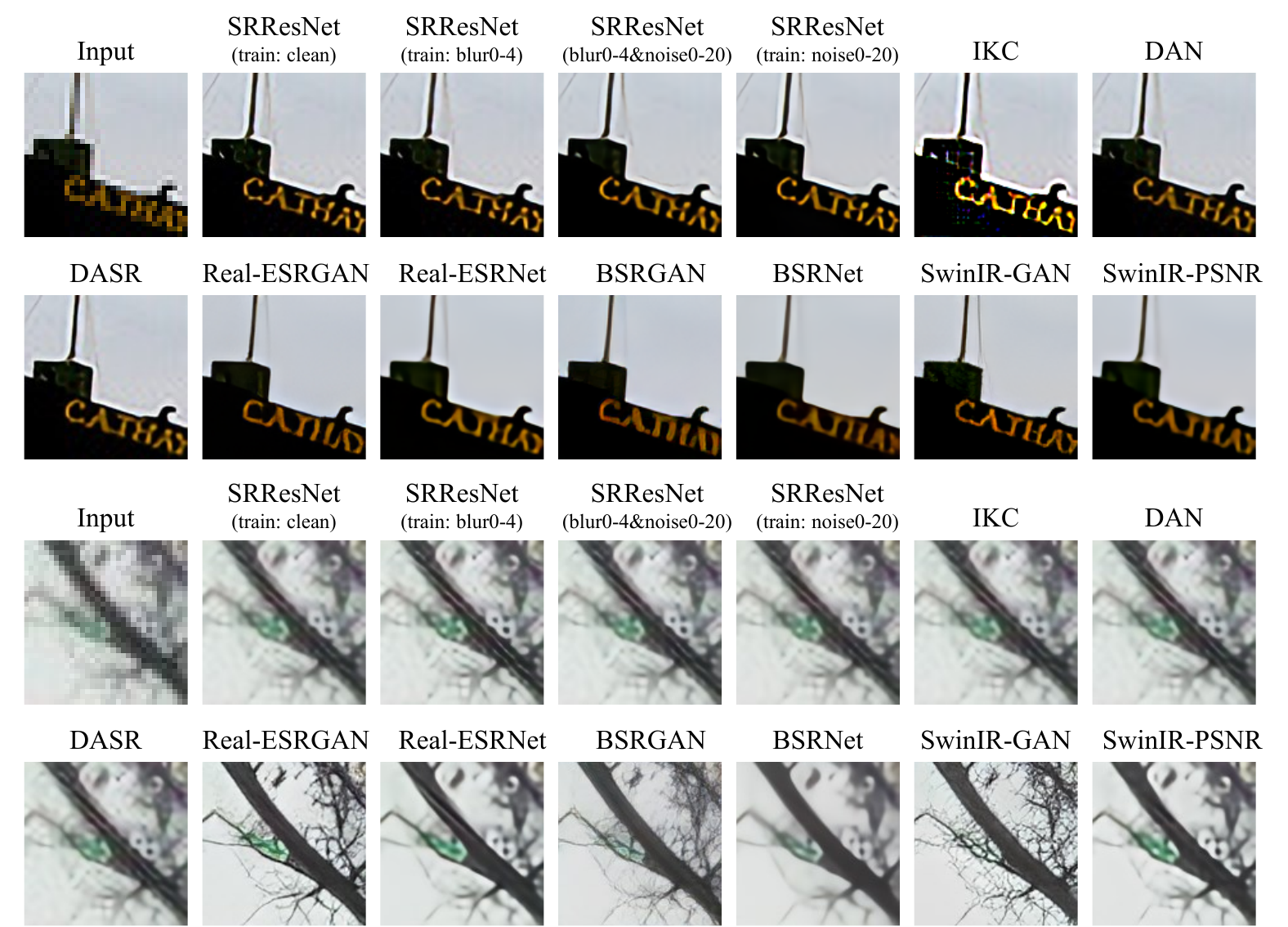}
	\end{center}
	\caption{Visual results of different models on PIES-RealLQ dataset.}
	\label{fig:reallq_img}
\end{figure*}

\subsection{Evaluating the Model Generalization Ability}
Model generalization ability measures the difference of processing effects on different types of input data. In the following, we analyze the model generalization results on blur, noise and real-world degraded input data. Note that the generalization ability is not equal to the output image quality. The benchmarking results are summarized in Table \ref{tab:benchmark}. The detailed quantitative results are described in the supplementary file.

\textbf{Blur degradation.} Figure \ref{fig:blur_KL}{(b)} summarizes the distances of the resulting feature distributions between clean LR input dataset and degraded input datasets with various isotropic Gaussian blur levels. Smaller value means the model can well generalize to the test dataset. Larger SRGA values imply the model processes the input degradation more differently than the chosen reference dataset. As can be seen, the distances of SRResNet (train: blur0-4) are continuously smaller than SRResNet (train: clean), which is intuitive and reasonable. The results can successfully describe the model performance change in a quantitative and analytical manner. Interestingly, for mild blur degradation (blur [0, 3]), DAN \cite{dan} has better model performance and generalization ability than SRResNet (train: blur0-2) and IKC. However, when the blur level becomes larger, its performance (PSNR, SSIM) and generalization (SRGA) deteriorate dramatically, indicating that DAN can only well generalized within mild degradation region. The proposed SRGA metric successfully describes such phenomenon. Another important observation is that methods like Real-ESRGAN \cite{realesrgan}, BSRGAN \cite{bsrgan} and SwinIR \cite{swinir} have relatively more stable and better generalization ability, especially for severe degraded data. Particularly, when the blur level is larger than 5, SRResNet, IKC \cite{ikc}, DAN \cite{dan} and DASR \cite{dasr} all fail to reconstruct the sharp images, while Real-ESRGAN, BSRGAN and SwinIR-GAN can still generate realistic images with sharp textures (see Figure \ref{fig:blur_img1}). It shows the superiority of such newly-proposed methods. This suggests that models trained with more degradations are likely to have better generalization despite that the absolute performance could deteriorate. For anisotropic blur degradation, as shown in Table \ref{tab:benchmark}, SRGA can also successfully embody the generalization ability of different models. Methods like BSRGAN, Real-ESRGAN and SwinIR still have good generalization on anisotropic blur degradation.

\textbf{Pairwise SRGA Matrix for Blur Degradation.} One of the key advantages of SRGA is its ability to eliminate the need for ground-truth images, making it a practical tool for real-world scenarios where ground-truth datasets are unavailable. This allows all candidate test datasets to be treated equally. Concretely, we can compute the SRGA score for each pair of datasets and obtain an SRGA matrix that provides a comprehensive view of a model's generalization ability. The resulting SRGA matrix provides a valuable tool for comparing and selecting models across a range of datasets, facilitating the identification of models with the best overall performance. Figure \ref{fig:SRGA_matrix_blur} presents the detailed SRGA matrices of different models under fine-grained blur levels. From Figure \ref{fig:SRGA_matrix_blur}(a), we can observe that, except for the diagonal elements, the SRGA values at other positions are considerably high, indicating that the model's performance in handling different degraded data significantly varies. This observation is consistent with our intuition and analysis that models trained only on clean data struggle with degraded data, leading to low generalization ability. Comparing Figure \ref{fig:SRGA_matrix_blur}(a) and Figure \ref{fig:SRGA_matrix_blur}(b), when adding blur data into training, the SRGA scores in low-level blur regions are relatively low, indicating that the model has similar processing effects on these datasets. The SRGA matrix provides a useful tool for describing the generalization distribution over different degradation data.

\textbf{Noise degradation.} As show in Figure \ref{fig:blur_KL}{(e)}, as the input noise level increases, the feature distributions become more divergent from that of the input clean data. When the noise level exceeds the range that the model can handle (denoise), the SRGA values increase dramatically. When noise level is extremely severe (\eg, noise level is larger than 30), most methods are unable to produce visually pleasing images. The results of SRResNet contains many noise residues and artifacts. The results of Real-ESRGAN (Real-ESRNet) are over-smoothed. The results of BSRGAN (BSRNet) are unsatisfactory with severe color distortion and artifacts. It is noteworthy that SwinIR-GAN shows peculiar characteristics when the noise level is large. The distance of the feature distributions between the clean input and severe noise input becomes smaller. The visual results of SwinIR-GAN can give a reasonable explanation and can well embody the superiority of the proposed distribution-based GA metric (see Figure \ref{fig:visual_noise2_sub}). SwinIR-GAN can produce relatively realistic output images by generating sharpened textures. Although the generated images are not consistent with the ground truth image, they look realistic with few noise residue and artifacts. SwinIR-GAN seems to treat the severe noisy input as a kind of texture and generate more textured details according to the noise distribution. This demonstrates SwinIR's powerful learning and generation capabilities. However, if we use the IQAs (especially reference-based IQA) to deduce the model generalization ability, such important phenomenon and observation cannot be perceived (see navy blur curves in Figure \ref{fig:blur_KL}{(d)}\&{(e)}). More discussions are in Section \ref{sec:comparing}. SRGA successfully helps us discover such an interesting fact. Specifically, the PSNR values continuously decrease when the noise level become larger, since it measures the pixel-wise distance between the SR results and the GT images. However, SRGA can help us find that the processing effect of SwinIR-GAN on severe noisy images is not continuously deteriorating. For severe noisy images, SwinIR-GAN can still produce visually plausible realistic results.

\textbf{Real-world degradation.} Another advantage of the proposed SRGA is that it does not require any paired reference images as ground-truth (GT). Thus, it can be applied to evaluate the model generalization ability on {real-world} images. As shown in Table \ref{tab:benchmark}, SwinIR \cite{swinir}, BSRGAN \cite{bsrgan} and Real-ESRGAN \cite{realesrgan} generalize well on PIES-RealCam dataset. PIES-RealCam dataset is collected from \cite{realsr}, which captures paired images by zooming the cameral focal lenth. The LR images mainly contain modest blur and noise degradations. Hence, it is reasonable that SRResNet (train: blur0-4\&noise0-20) has good generalization performance on this dataset as well. This suggests that models trained with synthetic blur and noise data are able to generalize to LR images simply caused by camera focal length. The visual results of PIES-RealCam dataset are shown in Figure \ref{fig:realcam_img}. For PIES-RealLQ dataset, the LR images are colleted from the Internet, which contain various degradations including blur, noise, compression, physical damage, \etc. The degree of degradation also ranges from mild to severe. As can be seen, IKC \cite{ikc}, DAN \cite{dan} and DASR \cite{dasr} have quite different processing effects on this dataset, compared with the chosen reference PIES-Clean dataset. This implies that although these models have good performance on clean LR inputs, they do not generalize well on the collected real-world images. On the contrary, Real-ESRGAN \cite{realesrgan} and SwinIR \cite{swinir} have a relatively good generalization ability on this dataset. This again suggests that exploiting abundant synthetic data into training is helpful for covering more realistic cases, especially for large models. The visual results of PIES-RealLQ dataset are shown in Figure \ref{fig:reallq_img}.

Note again, the generalization does not equal to model performance. It just measures the difference in the processing effect of the model on different inputs. To fully evaluate the merits of a model, we need to assess it comprehensively, including model performance, generalization ability, number of parameters, computational cost, \etc. SRGA only gives a new perspective for evaluating the generalization ability.

\begin{figure}[htbp]
	\begin{center}
		\includegraphics[width=0.9\linewidth]{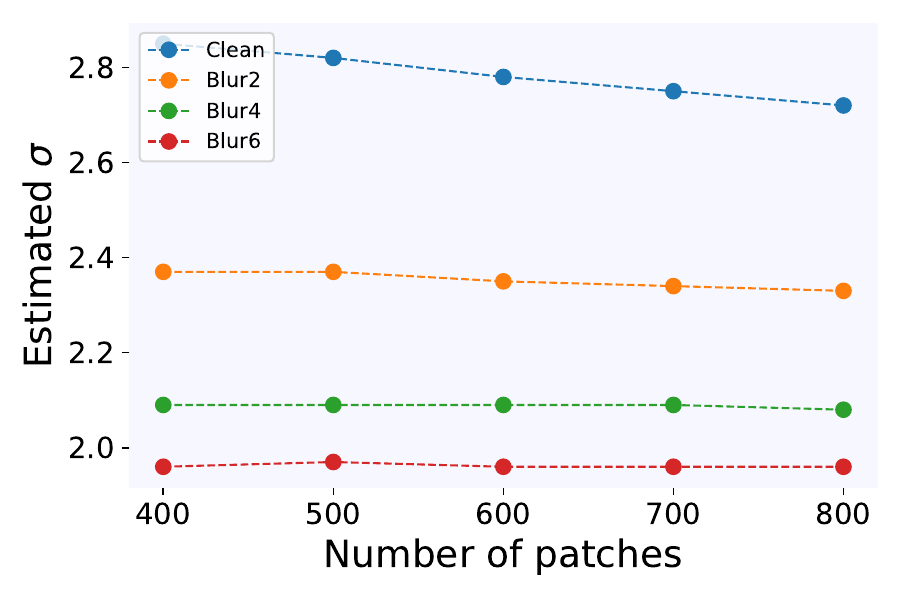}\\

	\end{center}
	\caption{As the number of patches increases, the estimated distribution parameters gradually converge.}
	
	\label{fig:property}
\end{figure}

\subsection{Comparing SRGA with IQA}\label{sec:comparing}
Notably, GA and IQA are conceptually different. \textit{They are two different but complementary facets of evaluating the model. We are not proposing GA to supplant IQA.} Previously, however, since there is not any viable metric for evaluating the model generalization, people tend to utilize IQA to measure the performance and to deduce the generalization. Notwithstanding, there are several limitations of IQA for measuring model generalization ability. In the following, we illustrate the advantages of SRGA and limitations of IQA in five aspects.

\begin{figure}[htbp]
	\begin{center}
		
		\includegraphics[width=0.9\linewidth]{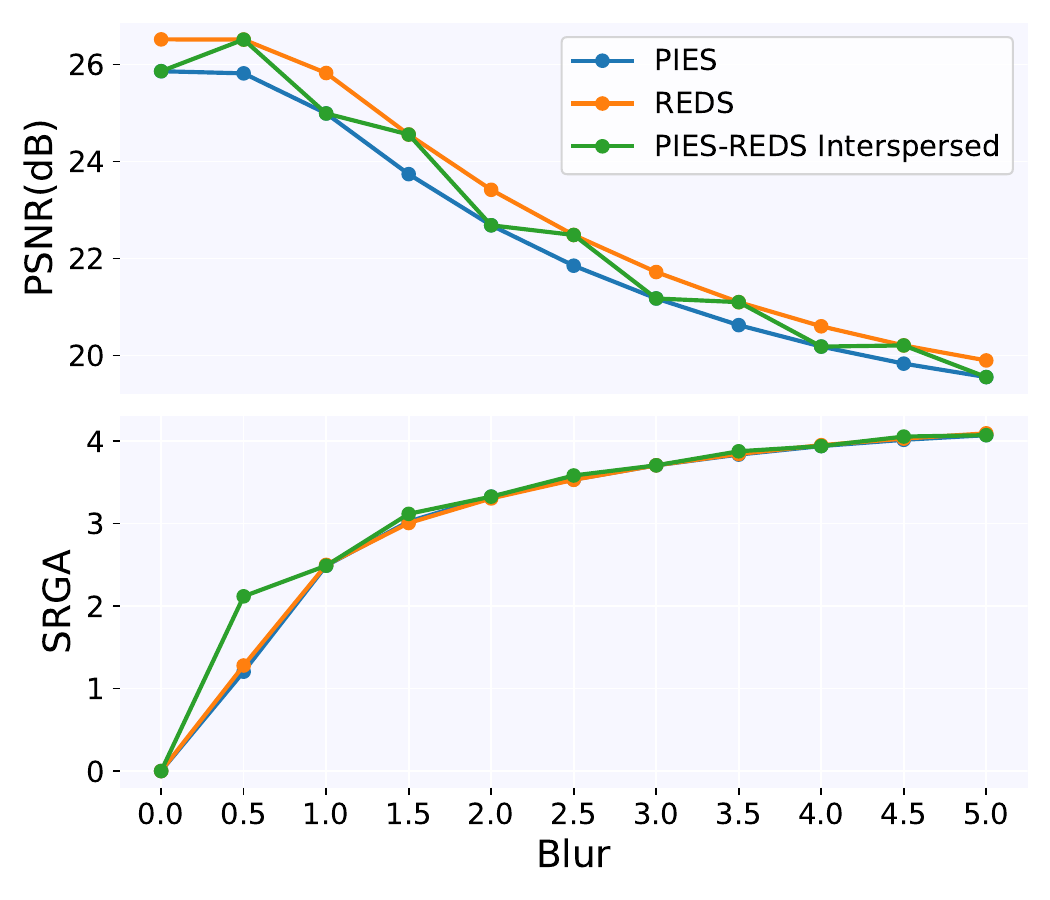}
		
	\end{center}
	
	\caption{Limitation of IQAs for measuring generalization ability. If the image contents are not aligned across degradations, PSNR values will fluctuate since it is heavily sensitive with the image contents while SRGA curves are more stable, focusing more on degradations.``PIES-REDS Interspersed'' means that the data in each blur level are interspersed by PIES and REDS.}
	\label{fig:limitations}
	
\end{figure}

\begin{figure*}[htbp]
	\begin{center}
		
		\includegraphics[width=1\linewidth]{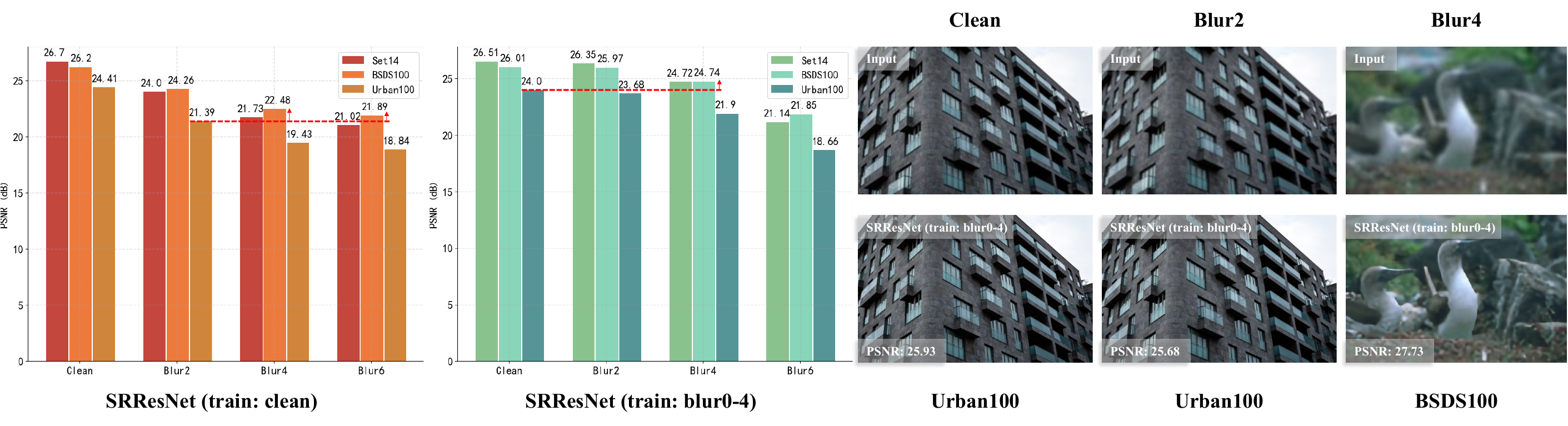}
		
	\end{center}
	\caption{IQA is highly sensitive to image content. Different datasets will have different absolute IQA values, even though they share the same degradation. For example, under the same degradation conditions, the average PSNR value of Urban100 dataset is consistently lower than Set14 and BSDS100 datasets. More counterintuitively, the PSNR value of Urban100 with blur level 2 is even worse than those of Set14 and BSDS100 with blur level 4 (marked with red dashed lines and arrows). If we use IQA to indicate the model generalization ability, we may come to a conclusion that the model has better generalization ability on the degradation of blur level 4 rather than blur level 2. However, the visual results do not support such a claim. For image restoration task, the model generalization ability should focus more on the input degradations rather than the specific contents. Concretely, for different datasets containing the same degradation type, the generalization ability of the model should be the same, \ie, models should have the same generalization on Urban100 with blur level 2 and BSDS100 with blur level 2. IQA fails to satisfy such a property thus is not appropriate for evaluating the model generalization ability.}
	\label{fig:IQA_content}
\end{figure*}

\textbf{SRGA \vs PSNR in sensitivity of image content.}
IQA is highly sensitive to image content, thus will have different values on different images. Different datasets will have different absolute IQA values, even though they share the same degradation. While in image restoration, we need to give a stable measurement on unseen degradations but not specific datasets. For different datasets containing the same degradation type, the generalization ability of the model should be the same. An illustration is depicted in Figure \ref{fig:limitations} and ~\ref{fig:IQA_content}. The proposed SRGA metric stems from the statistical characteristics of deep features of SR model: for a given model, different input degradation types will lead to different feature probability distributions, and such distributions are less sensitive to the image content but have higher response to image degradation, as illustrated in Section \ref{sec:Statistics}. Similar property is also observed in previous literatures \cite{mittal2012no,mittal2012making,DDR,DDP}. We further validate this property. We select different numbers of image patches from PIES-Blur dataset to fit the corresponding distribution parameters ($\sigma$ and $\alpha$) using SRResNet(train:clean) model. Thus, the contents of the selected image sets with different numbers of patches are inconsistent. The patches are randomly selected and each setting is repeated three times. As shown in Figure \ref{fig:property}, with the increase of patch number, the estimated parameters gradually converge. Moreover, we select different datasets at different blur levels, so that the contents of the images are not aligned. For example, the data for clean, blur1, and blur2 are from the PIES dataset, and the data for blur0.5, blur1.5, and blur2.5 are from the REDS \cite{nah2019ntire} dataset. Figure \ref{fig:limitations} plots the corresponding PSNR and SRGA curves. The PSNR values are largely affected by the image content, leading to a fluctuating PSNR curve (not monotonic). But the SRGA curve can still depict the generalization trend of the model, despite of the fact that we use different datasets for calculation.\footnote{In this experiment, PIES-Clean is selected as the reference dataset.} 

\begin{figure*}[htbp]
	\begin{center}
		\begin{minipage}[t]{0.48\linewidth}
			\centering
			\includegraphics[width=0.92\linewidth]{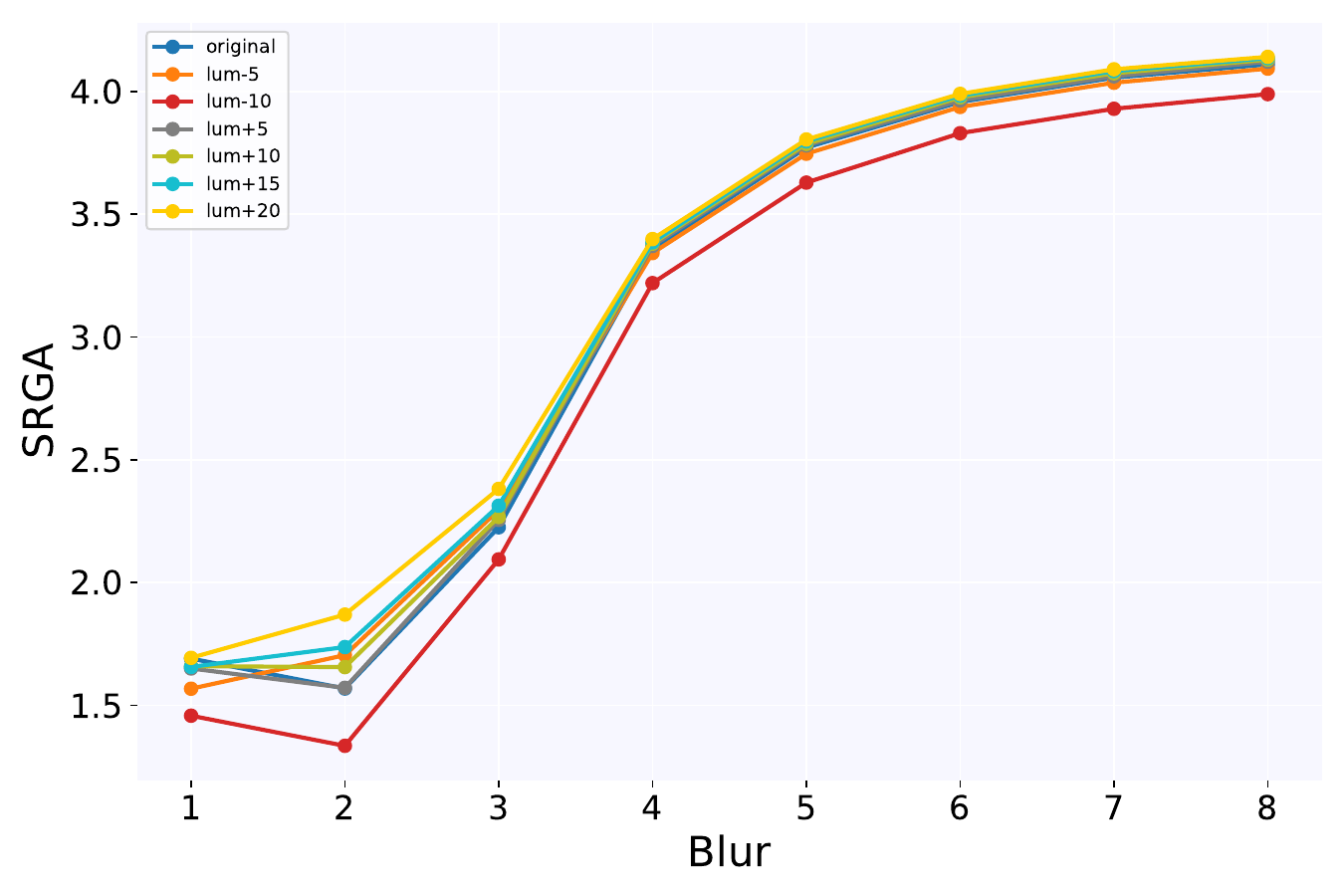}\\
			\scriptsize (a)
		\end{minipage}
		\begin{minipage}[t]{0.48\linewidth}
			\centering
			\includegraphics[width=0.92\linewidth]{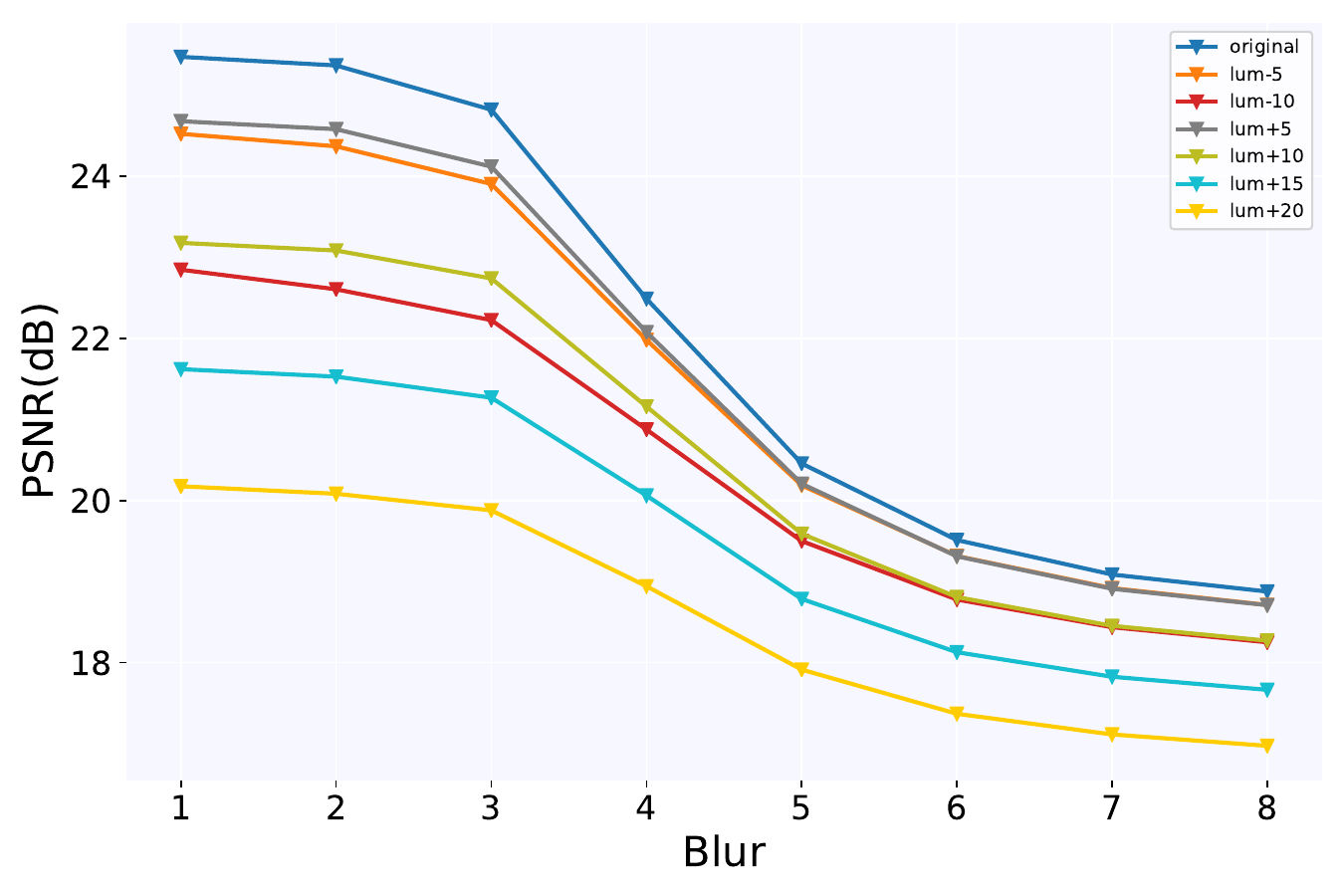}\\
			\scriptsize (b)
		\end{minipage}

	\end{center}
	\caption{When the global brightness between the test and reference images is not aligned (luminance jitter), SR models may still produce good SR results, but IQA will be severely affected, especially for the most commonly-used PSNR. Nevertheless, SRGA is more robust to such disturbances.}
	
	\label{fig:luminance}
\end{figure*}

\textbf{SRGA \vs PSNR in luminance jitter.}
Another advantageous property of SRGA is that it is more robust to image luminance jitter, which is a common photographic disturbance in real camera imaging. To validate the robustness of SRGA, we manually shift the luminance of the test PIES-Blur images and maintain the original luminance of the reference PIES-Clean dataset. As revealed in Figure \ref{fig:luminance}, when the global luminance jitter is within $[-10, +20]$, SRGA is basically unaffected and maintains good stability. However, PSNR is sensitive to such a disturbance, since it measures the absolute distance between paired pixels. This shows the great superiority of SRGA over IQA. SRGA can deal with such disturbances since the PCA operation has substracted the mean of the features beforehand. If the brightness changes too much, SRGA will also change, because the input distribution has altered a lot. This is equivalent to adding a new degradation to the input image, which is out of the scope of this discussion.

\begin{figure}[htbp]
	\begin{center}
		
		\includegraphics[width=0.9\linewidth]{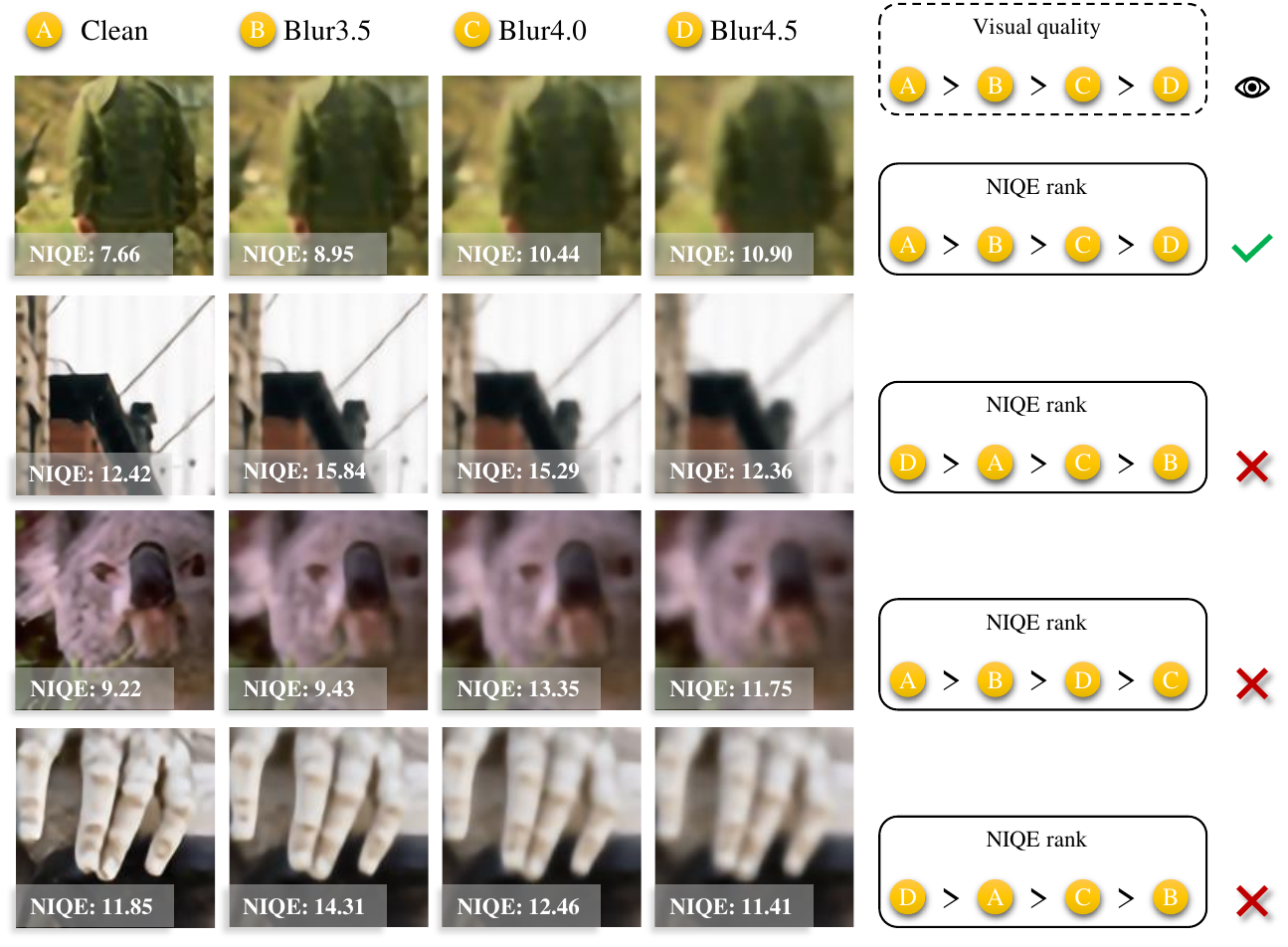}
		
	\end{center}
	
	\caption{When there is no corresponding ground-truth image, FR-IQAs like PSNR, SSIM and LPIPS are unavailable. Existing commonly-used NR-IQA like NIQE is instable and less accurate. We evaluate the results of SRResNet (train:blur0-4) on PIES-Blur dataset. The NIQE values cannot reflect the actual visual quality. For example, in the second row, the NIQE value of Blur3.5 data is even higher than that of Blur4.0 and Blur4.5. In the fourth row, the visual quality of Clean data is obvious better than others. However, the NIQE value of Clean data is worse than that of Blur4.5 data. Note that Lower NIQE values indicate better performance. The NIQE results often do not match subjective image quality evaluations. This reveals that NR-IQA is much less accurate in evaluating the model performance, let alone deducing the model generalization ability.}
	\label{fig:niqe_fail}
	
\end{figure}

\textbf{SRGA \vs NR-IQA in performance stability.}
For no-reference IQA, like NIQE, it is widely acknowledged that it is unstable and less accurate \cite{gu2021ntire,bsrgan}. As shown in Figure \ref{fig:blur_PSNR}(c), the NIQE curves of SRResNet (train: blur0-4) and SRResNet (train: blur0-4\&noise0-20) on PIES-Blur dataset are counterintuitive and abnormal. The NIQE value reaches the maximum at Blur4 data, and then drops suddenly. To better illustrate the abnormal behavior of NIQE, we show some visual results of SRResNet (train: blur0-4) in Figure \ref{fig:niqe_fail}. The visual examples reveal that the NIQE values cannot always reflect the actual image quality. In the second row, the NIQE value of Blur3.5 data is even higher than that of Blur4.0 and Blur4.5. However, the image quality of Blur3.5 is clearly better than Blur4.0 and Blur4.5 with sharper textures. Similar phenomena are observed in the third and fourth rows. This problem occurs not only in a few images, but also in a large amount of images, especially for images with blur level around 4.0. In addition, as shown in Figure \ref{fig:blur_PSNR}(f), NIQE values also fluctuate greatly on PIES-Noise dataset, compared with PSNR. In Figure \ref{fig:visual_noise2_sub}, the results of SwinIR-GAN on Noise$>20$ look visually similar, but their NIQE values differ a lot. This can be attributed to the fact that NIQE is simply determined by limited handcrafted features. Besides NIQE, Gu \etal \cite{jinjin2020pipal,gu2021ntire} have demonstrated that other existing NR-IQAs are also far from satisfactory. Inaccuracy and instability make NR-IQA unsuitable as a good GA.

\textbf{SRGA \vs IQA in real data without GT.} Full-reference IQA, like PSNR, SSIM and LPIPS, requires paired ground-truth (GT) images, which are impractical in real scenarios. No-reference IQA is inaccurate and instable as mentioned before. However, SRGA does not rely on paired reference images, since it considers the distance between statistical feature distributions. As shown in Table \ref{tab:benchmark}, SRGA can be successfully applied on real-world datasets.

\textbf{Failure cases of IQA.} IQA cannot precisely reflect the model generalization. For example, when the noise level is extremely severe (\eg, larger than 30), most methods are unable to produce visually pleasing images. However, SwinIR-GAN can produce relatively realistic output images by generating imaginary textures (see Figure \ref{fig:visual_noise2_sub}). This demonstrates SwinIR's powerful learning and generation capabilities. The PSNR curve of SwinIR continues to decline, while the SRGA curve well depicts the trend of model generalization (see navy blur curves in Figure \ref{fig:blur_KL}{(d)}\&{(e)}). This suggests the intrinsic superiority of SRGA.

These aforementioned issues have prevented IQA from being a qualified GA. In fact, IQA and SRGA are complementary. IQA can be used to evaluate the absolute model performance and help select the reference dataset for SRGA. If a model has extremely low IQA score, such as a randomly initialized network, it is meaningless to discuss its generalization ability, and there is no need to consider its SRGA value. Hence, IQA and SRGA should cooperate with each other.

\section{Prospects and Limitations}
\textcolor{black}{
	Although the SRGA metric is designed specifically for blind super-resolution tasks, it can also be applied to other low-level vision tasks that face similar generalization challenges. For example, denoising, deblurring, and dehazing models are trained on a set of degraded images and expected to perform well on a wide range of in/out-distribution degraded images. The SRGA metric can be used to evaluate the generalization ability of these models across various types and levels of degradations. By using the SRGA metric in these tasks, we can obtain a reliable measure of a model's generalization ability to handle real-world scenarios. In the future work, we will explore the possibility of extending the SRGA metric to other image restoration tasks and investigating its effectiveness in evaluating the generalization performance of other more models.
}

\textcolor{black}{
	Since we are the first to propose a generalization assessment metric for SR networks, our work has several limitations. First, SRGA requires a relatively large amount of computation and storage. Specifically, to estimate the GGD parameters of a dataset with 200 output features and dimensions of 128 $\times$ 128 $\times$ 64 (with input dimensions of 32 $\times$ 32 $\times$ 3), the time cost is approximately 120 seconds using two physical Intel(R) Xeon(R) Gold 6258R CPUs with 32GB RAM. Second, unlike IQA, it is hard to subjectively evaluate GA by user study, as GA requires to measure the processing consistency between degraded datasets, not individual images. Third, SRGA is a statistical method, which cannot be directly used as a loss function to optimize the network. But we can mimic its behavior to guide the design of future models.
}

\section{Conclusion}
In this paper, we present a novel perspective for measuring the SR model's generalization ability. The proposed generalization assessment index, named SRGA, utilizes the statistics of the internal deep features of SR networks. Specifically, we adopt generalized Gaussian distribution to model the deep features and then calculate the distance between different degraded input sets. Interestingly, SRGA does not require paired reference images or any learning process. The proposed SRGA and PIES datasets can help promote the development of blind SR methods, as well as other low-level vision problems.

\section*{Acknowledgment}
This work was supported in part by the National Key R\&D Program of China (NO. 2022ZD0160100), the National Natural Science Foundation of China under Grant (62276251, 62272450), the Joint Lab of CAS-HK, and in part by the Youth Innovation Promotion Association of Chinese Academy of Sciences (No. 2020356).

\bibliographystyle{IEEEtran}
\bibliography{egbib}

\begin{IEEEbiography}
	[{\includegraphics[width=1in,height=1.25in,clip,keepaspectratio]{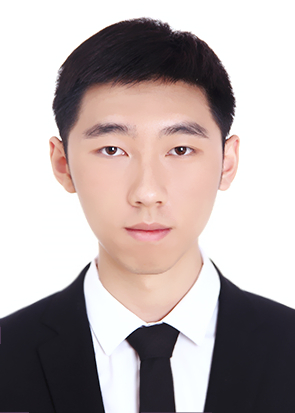}}]{Yihao Liu} received the B.S. and Ph.D. degrees from the University of Chinese Academy of Sciences in 2018 and 2023, respectively. During his Ph.D. studies, he was affiliated with the Shenzhen Institute of Advanced Technology, Chinese Academy of Sciences. Currently, he holds a research position at the Shanghai Artificial Intelligence Laboratory. His research interests include computer vision and image processing, with a distinct focus on image and video restoration and enhancement.
\end{IEEEbiography}
\begin{IEEEbiography}
	[{\includegraphics[width=1in,height=1.25in,clip,keepaspectratio]{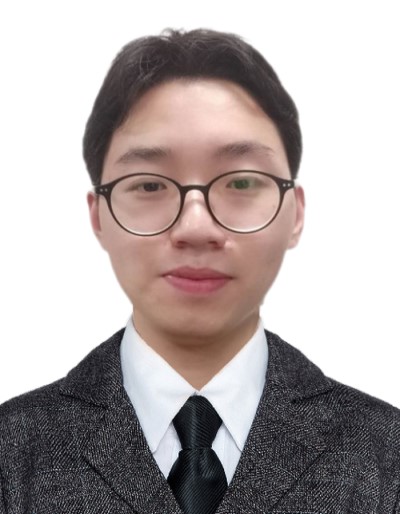}}]{Hengyuan Zhao} is a Ph.D. student at National University of Singapore supervised by Mike Zheng Shou. He worked as a research intern in
	Multimedia Laboratory, ShenZhen Institute of Advanced Technology, Chinese Academy of Sciences.
	He was supervised by Prof. Yu Qiao and Prof. Chao
	Dong. His research interests include computer vision
	and image/video processing.
\end{IEEEbiography}
\begin{IEEEbiography}
	[{\includegraphics[width=1in,height=1.25in,clip,keepaspectratio]{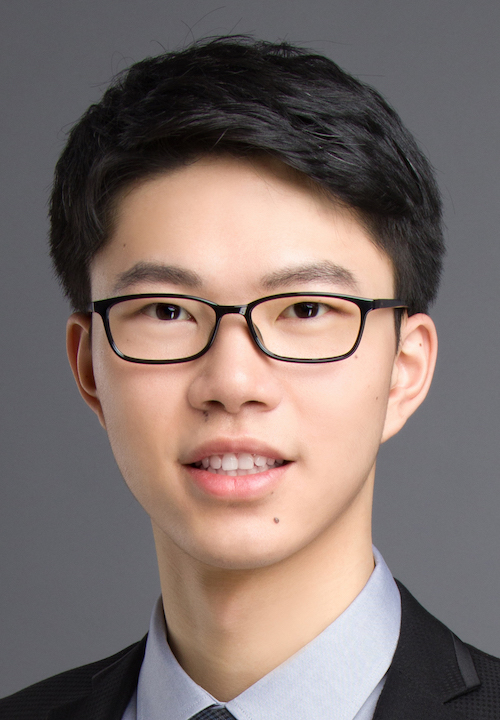}}]{Jinjin Gu} received a bachelor's degree in computer science and engineering from the Chinese University of Hong Kong (Shenzhen) in 2020, and is currently pursuing a doctoral degree in the School of Electrical and Information Engineering, University of Sydney, Australia. He has published more than 30 papers in well-known conferences and journals such as CVPR, ECCV, NeurIPS, IEEE TPAMI. He has received honors such as SenseTime Scholarship, Huawei Camera Academic Star, and the 2023 Yunfan Award.
\end{IEEEbiography}
\begin{IEEEbiography}
	[{\includegraphics[width=1in,height=1.25in,clip,keepaspectratio]{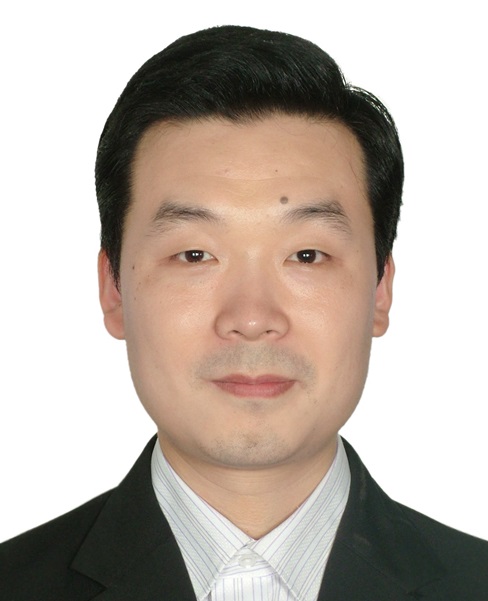}}]{Yu Qiao} (Senior Member, IEEE) is a professor with Shanghai AI Laboratory and the Shenzhen Institute of Advanced Technology (SIAT), Chinese Academy of Sciences. He has published more than 600 articles in international journals and conferences, including T-PAMI, IJCV, T-IP, T-SP, CVPR, and ICCV. His research interests include computer vision, deep learning, and bioinformation. He received the First Prize of the Guangdong Technological Invention Award, and the Jiaxi Lv Young Researcher Award from the Chinese Academy of Sciences. He is a recipient of the distinguished paper award in AAAI 2021. His group achieved the first runner-up at the ImageNet Large Scale Visual Recognition Challenge 2015 in scene recognition, and the winner at the ActivityNet Large Scale Activity Recognition Challenge 2016 in video classification. He served as the program chair of IEEE ICIST 2014.
\end{IEEEbiography}
\begin{IEEEbiography}
	[{\includegraphics[width=1in,height=1.25in,clip,keepaspectratio]{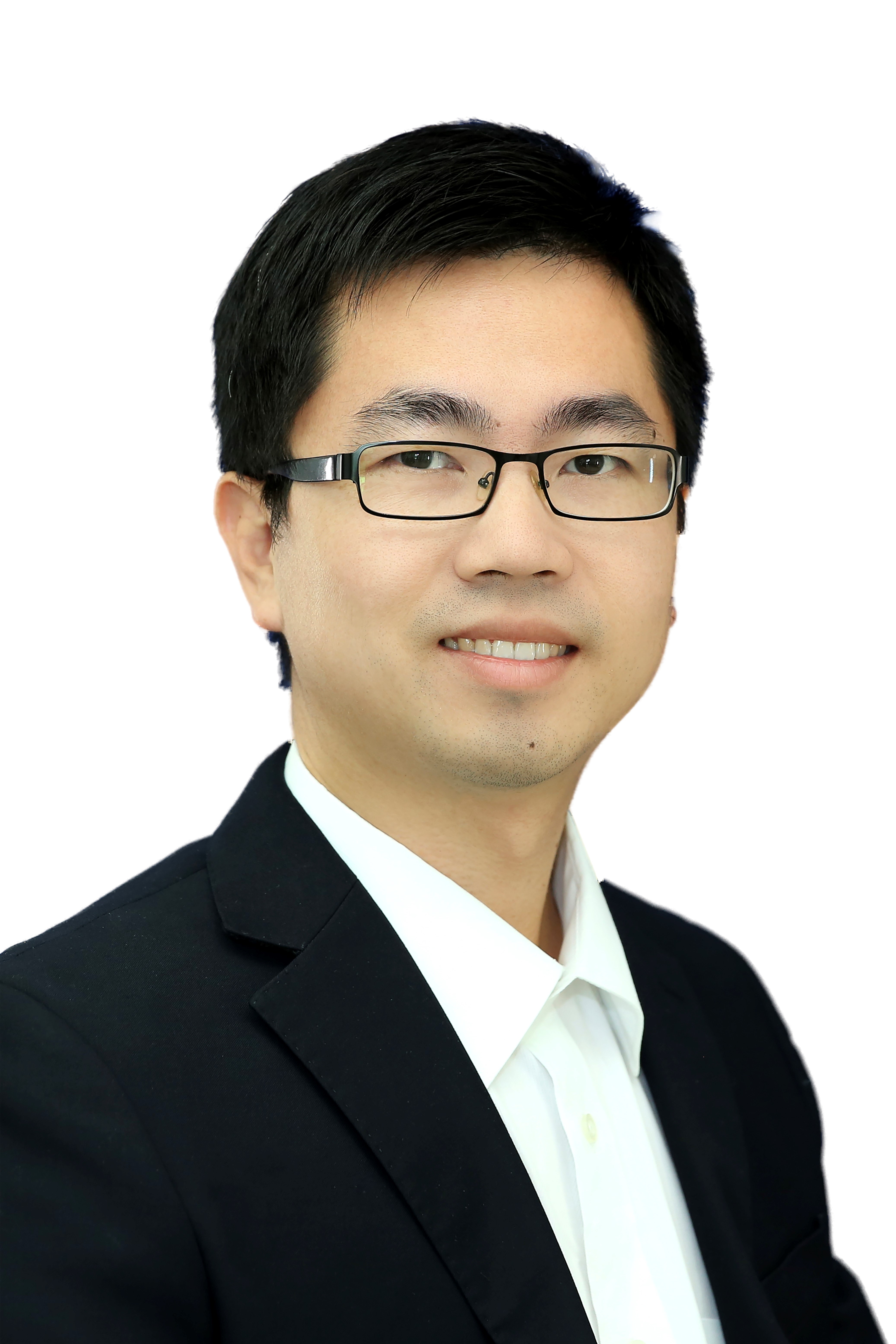}}]{Chao Dong} is a professor with Shenzhen Institute of Advanced Technology, Chinese Academy of Science (SIAT), and Shanghai AI Laboratory. In 2014, he first introduced deep learning method – SRCNN into the super-resolution field. This seminal work was chosen as one of the top ten “Most Popular Articles” of TPAMI in 2016. His team has won several championships in international challenges –NTIRE2018, PIRM2018, NTIRE2019, NTIRE2020 AIM2020 and NTIRE2022. He worked in SenseTime from 2016 to 2018, as the team leader of Super-Resolution Group. In 2021, he was chosen as one of the World’s Top 2\% Scientists. In 2022, he was recognized as the AI 2000 Most Influential Scholar Honorable Mention in computer vision. His current research interest focuses on low-level vision problems, such as image/video super-resolution, denoising, and enhancement.
\end{IEEEbiography}

\clearpage

\appendices
\section{Complete Benchmarking Results}
\begin{figure*}[htbp]
	\begin{center}
		\begin{minipage}[t]{0.45\linewidth}
			\centering
			\includegraphics[width=0.9\linewidth,height=0.7\linewidth]{figures/Blur_PSNR}\\
			\scriptsize (a)
		\end{minipage}
		\begin{minipage}[t]{0.45\linewidth}
			\centering
			\includegraphics[width=0.9\linewidth,height=0.7\linewidth]{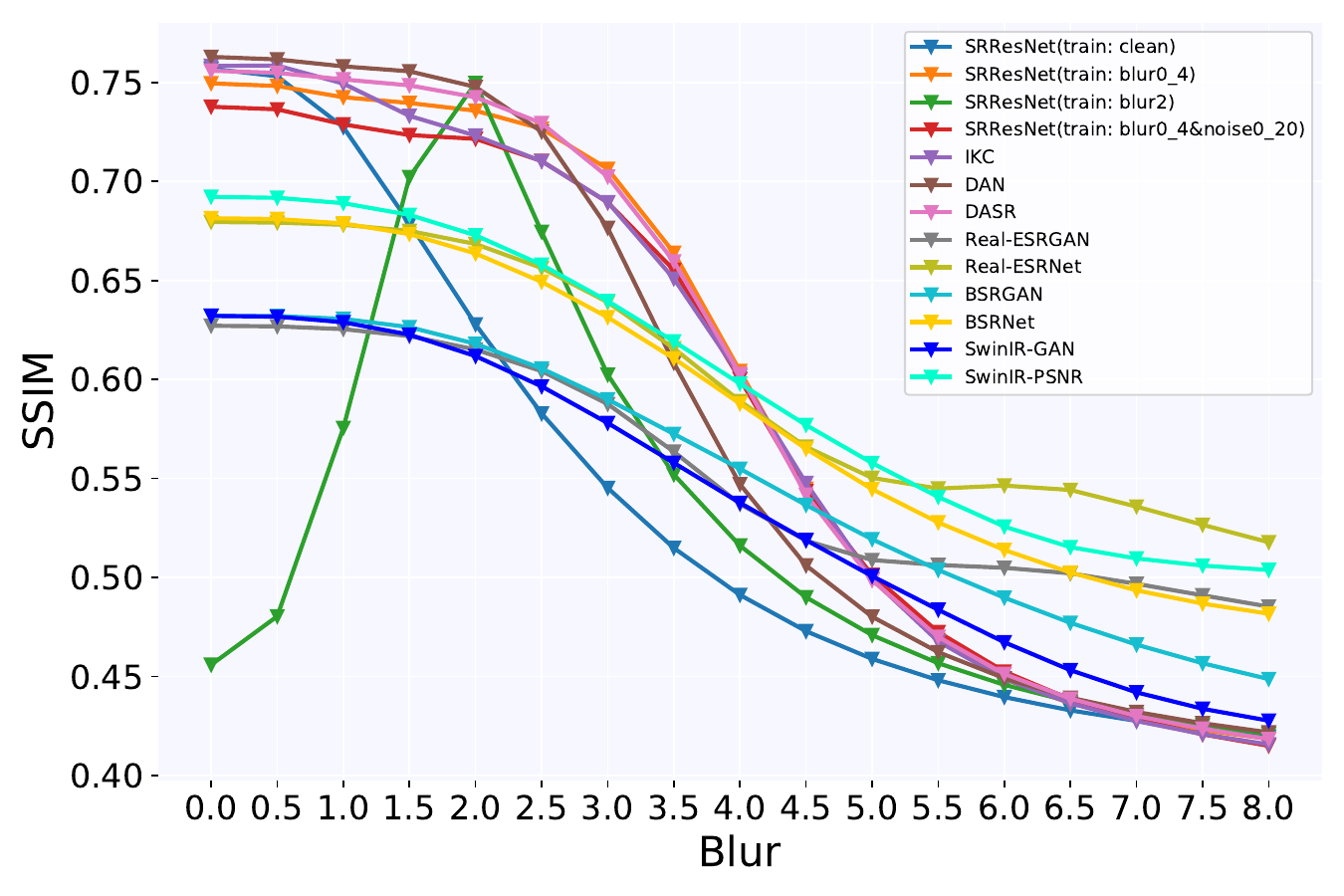}\\
			\scriptsize (b)
		\end{minipage}
		
		\begin{minipage}[t]{0.45\linewidth}
			\centering
			\includegraphics[width=0.9\linewidth,height=0.7\linewidth]{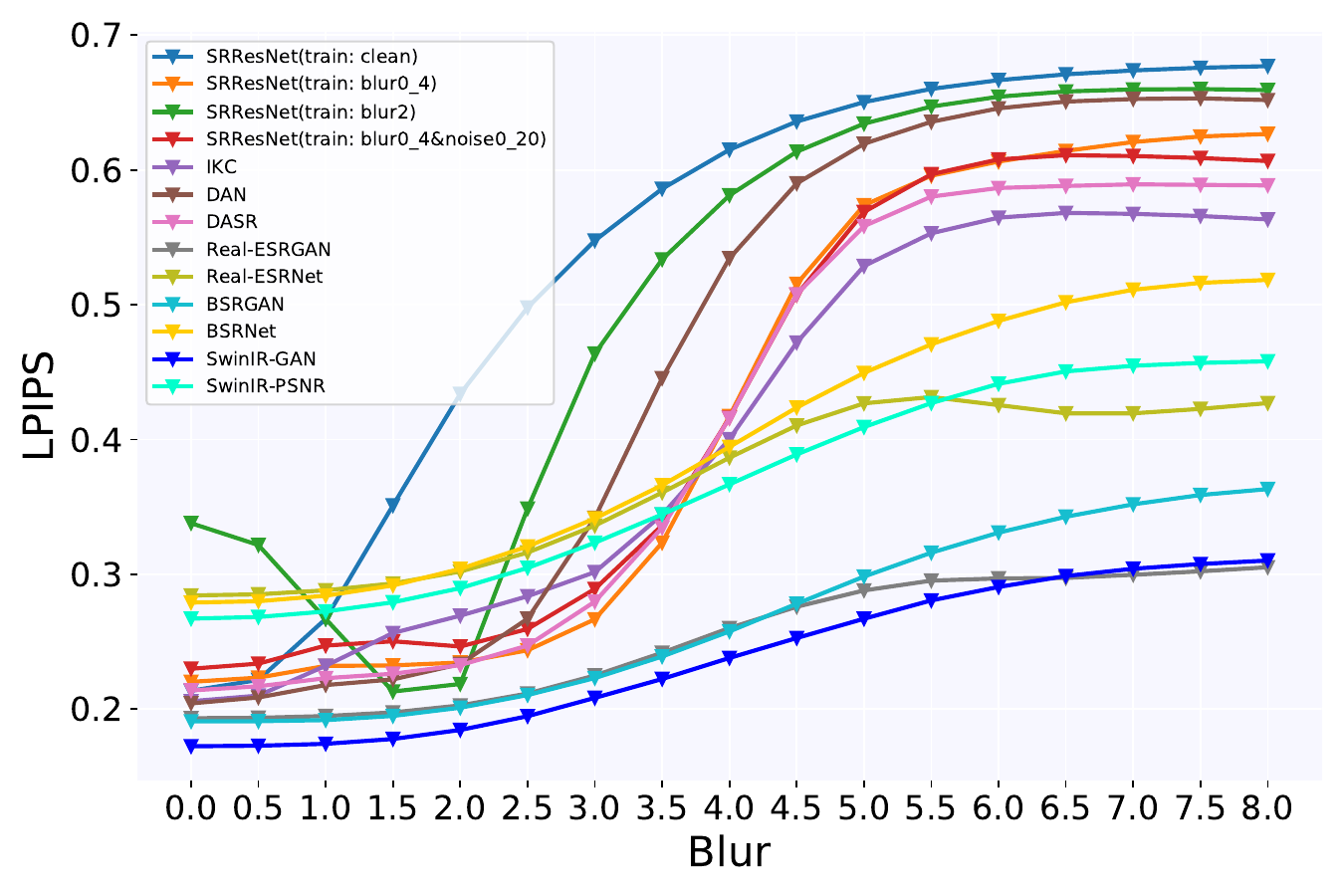}\\
			\scriptsize (c)
		\end{minipage}
		\begin{minipage}[t]{0.45\linewidth}
			\centering
			\includegraphics[width=0.9\linewidth,height=0.7\linewidth]{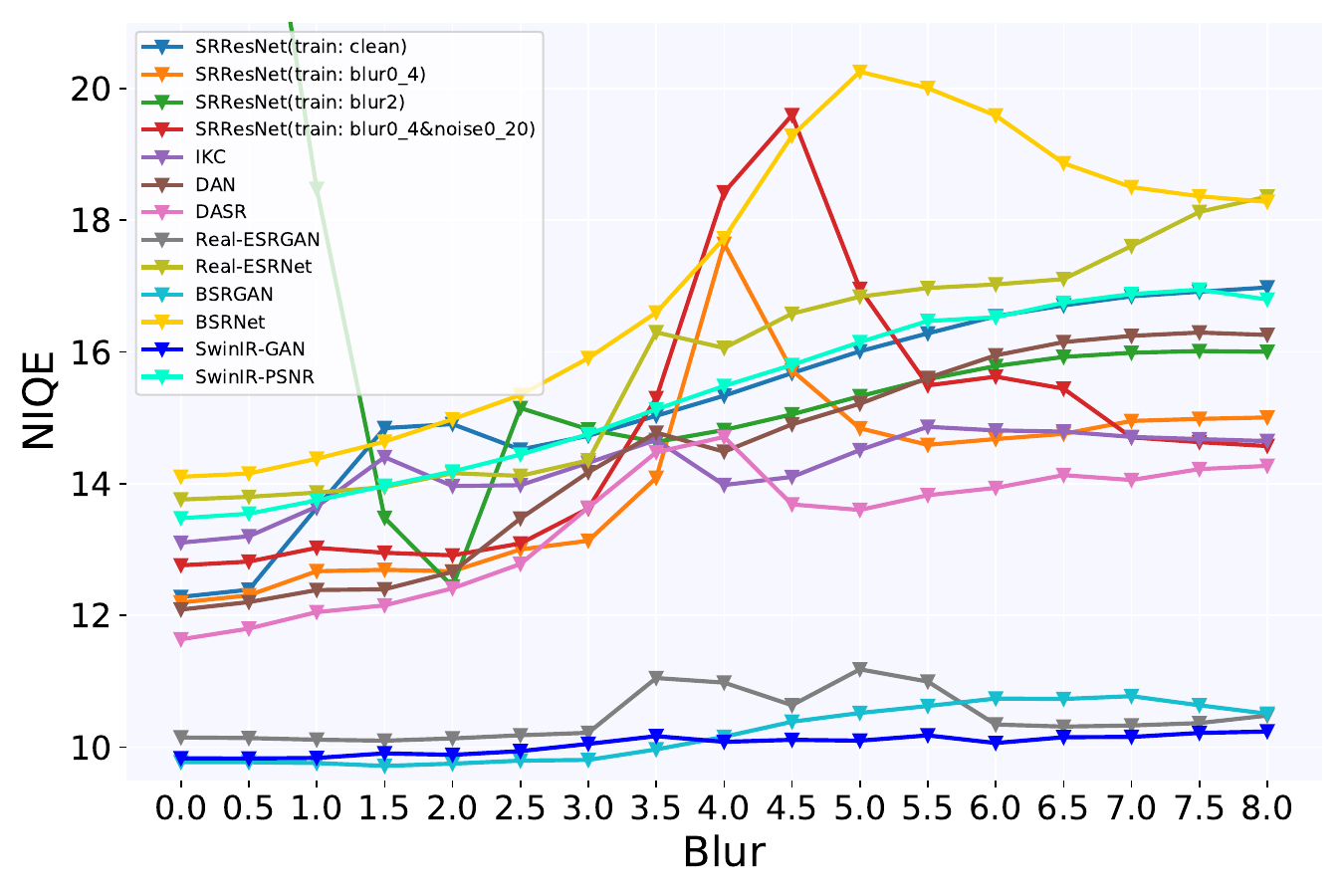}\\
			\scriptsize (d)
		\end{minipage}

	\end{center}
	\caption{The PSNR, SSIM, LPIPS and NIQE curves of different methods on PIES-Blur dataset.}
	\label{fig:blur_curves}
\end{figure*}

\begin{figure*}[htbp]
	\begin{center}
		\begin{minipage}[t]{0.45\linewidth}
			\centering
			\includegraphics[width=0.9\linewidth,height=0.7\linewidth]{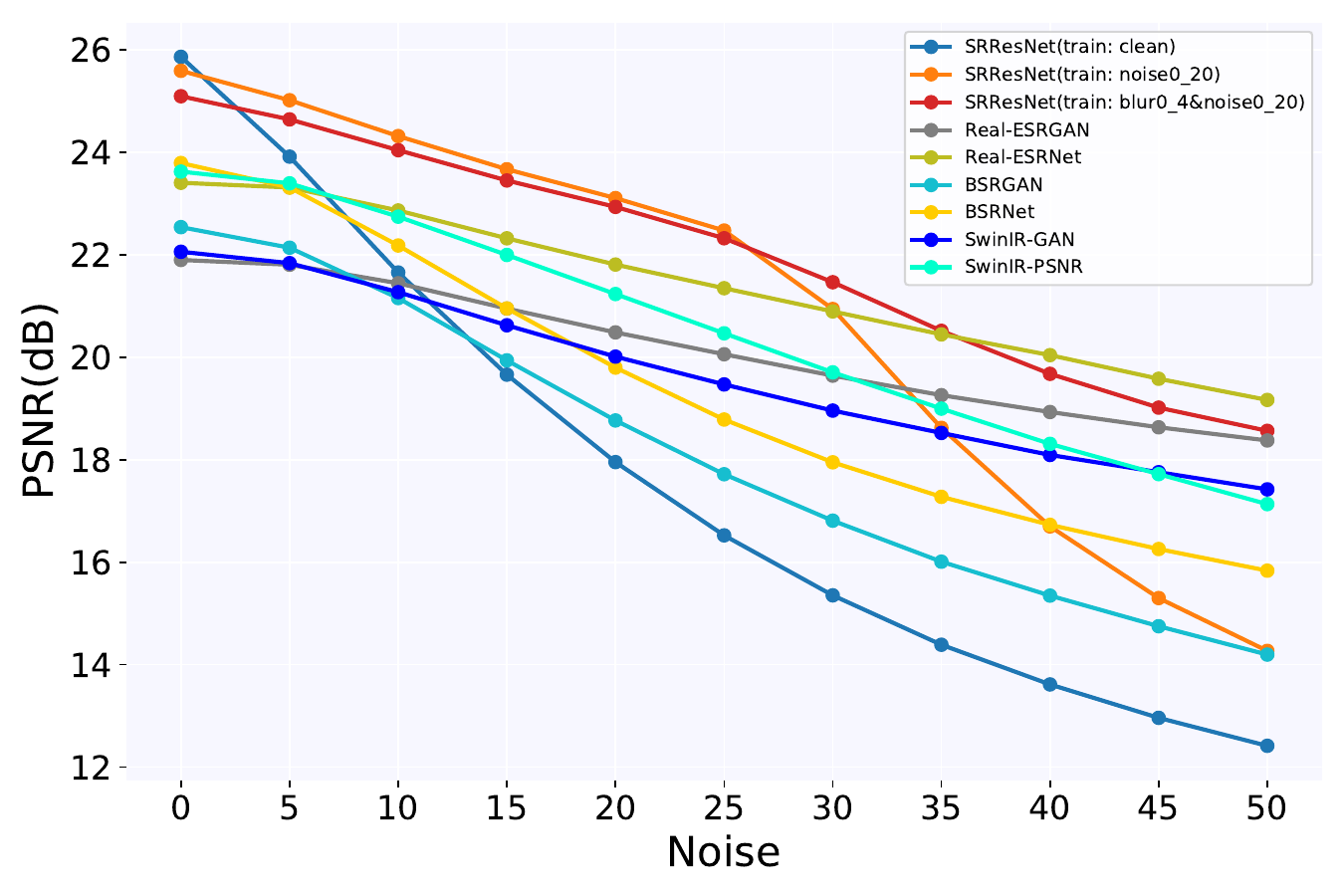}\\
			\scriptsize (a)
		\end{minipage}
		\begin{minipage}[t]{0.45\linewidth}
			\centering
			\includegraphics[width=0.9\linewidth,height=0.7\linewidth]{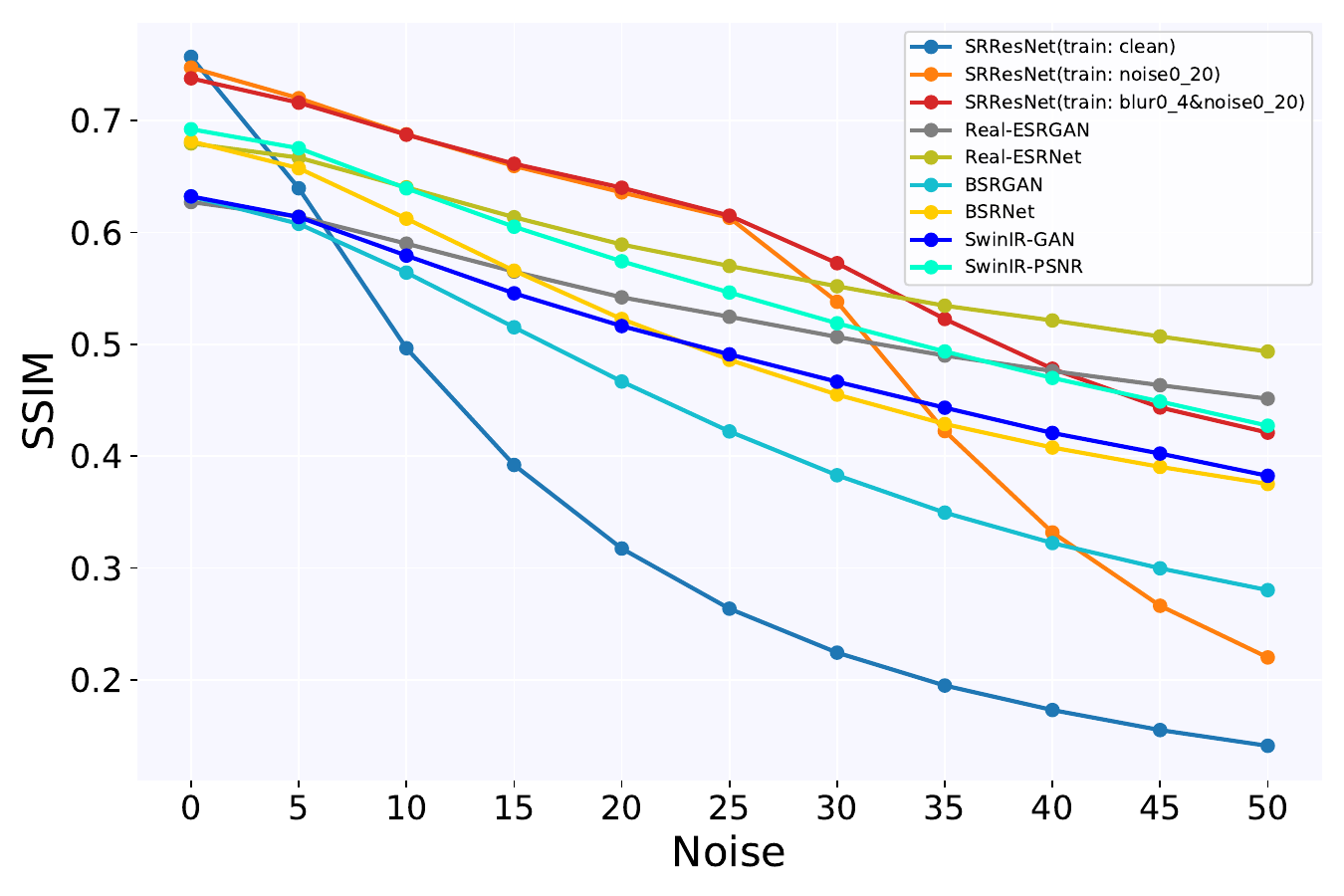}\\
			\scriptsize (b)
		\end{minipage}
		
		\begin{minipage}[t]{0.45\linewidth}
			\centering
			\includegraphics[width=0.9\linewidth,height=0.7\linewidth]{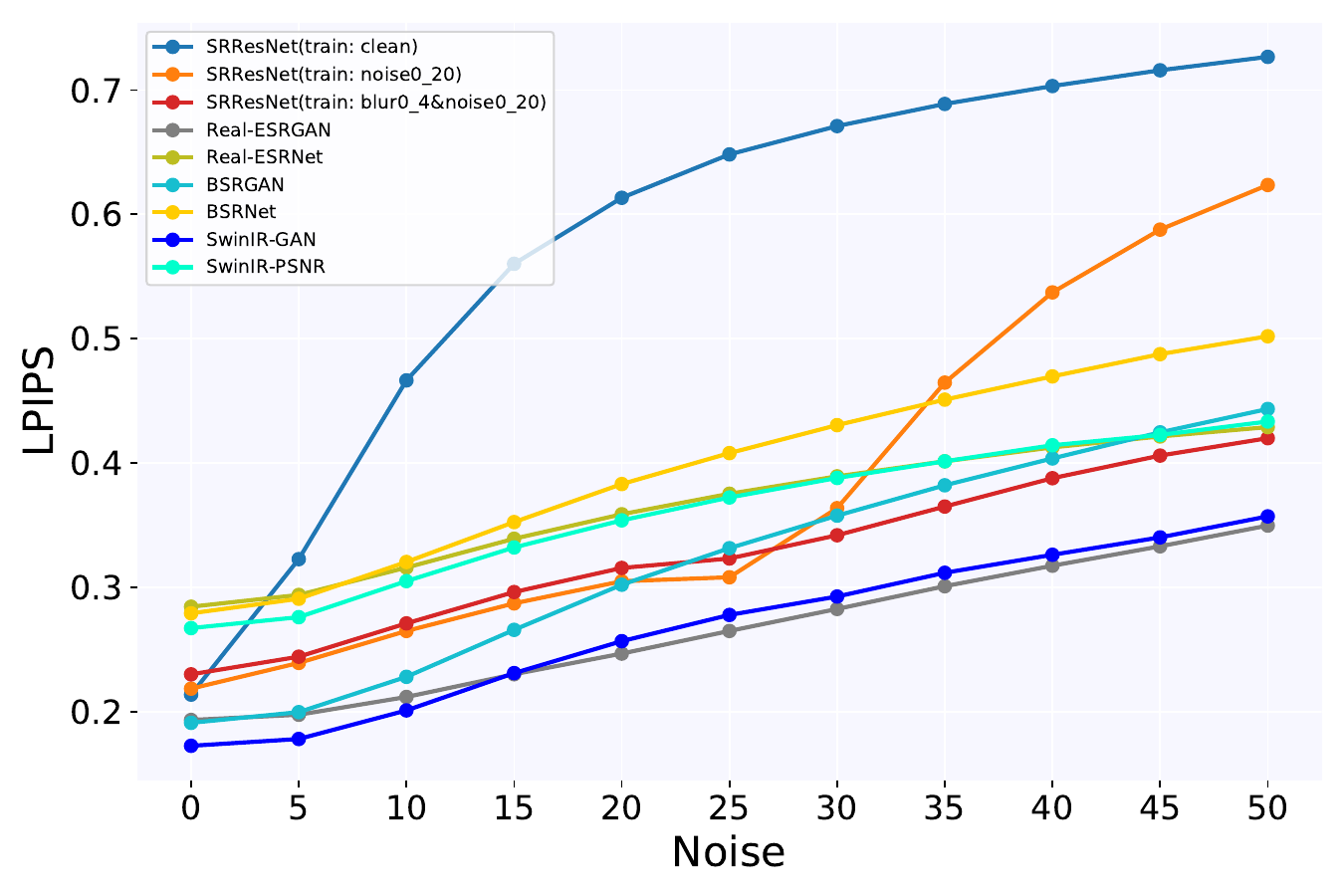}\\
			\scriptsize (c)
		\end{minipage}
		\begin{minipage}[t]{0.45\linewidth}
			\centering
			\includegraphics[width=0.9\linewidth,height=0.7\linewidth]{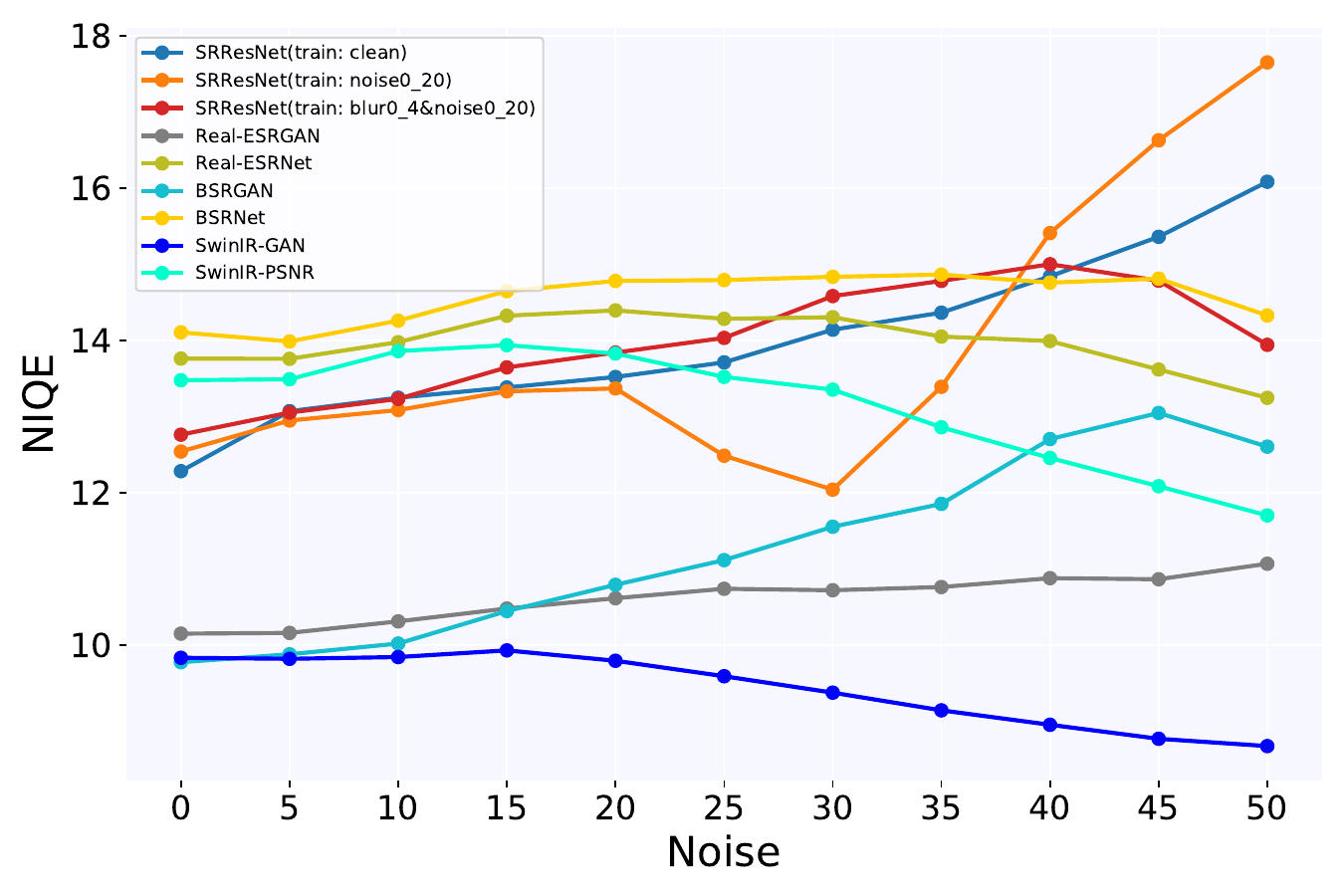}\\
			\scriptsize (d)
		\end{minipage}

	\end{center}
	\caption{The PSNR, SSIM, LPIPS and NIQE curves of different methods on PIES-Noise dataset.}
	\label{fig:noise_curves}
\end{figure*}

We present the completee quantitative results in this section. The benchmarking results of model generalization ability based on SRGA are summarized in Tab. \ref{tab:blur_srga}, \ref{tab:noise_srga}, \ref{tab:blurnoise_srga}, \ref{tab:real_srga} and \ref{tab:aniso_blur} . The corresponding SRGA curves are already shown in the main paper. In addition, we also show the PSNR, SSIM, LPIPS, NIQE values, which evaluate the output image quality and measure the model performance, as shown in Tab. \ref{tab:blur_psnr}, \ref{tab:blur_ssim}, \ref{tab:blur_lpips}, \ref{tab:blur_niqe}, \ref{tab:noise_psnr}, \ref{tab:noise_ssim}, \ref{tab:noise_lpips}, \ref{tab:noise_niqe}, \ref{tab:blurnoise_psnr}, \ref{tab:blurnoise_ssim}, \ref{tab:blurnoise_lpips}, \ref{tab:blurnoise_niqe} and \ref{tab:real}. The corresponding curves are depicted in Fig. \ref{fig:blur_curves} and \ref{fig:noise_curves}.

\section{Visualization of Output Images}
\begin{figure*}[htbp]
	\begin{center}
		\includegraphics[width=0.95\linewidth]{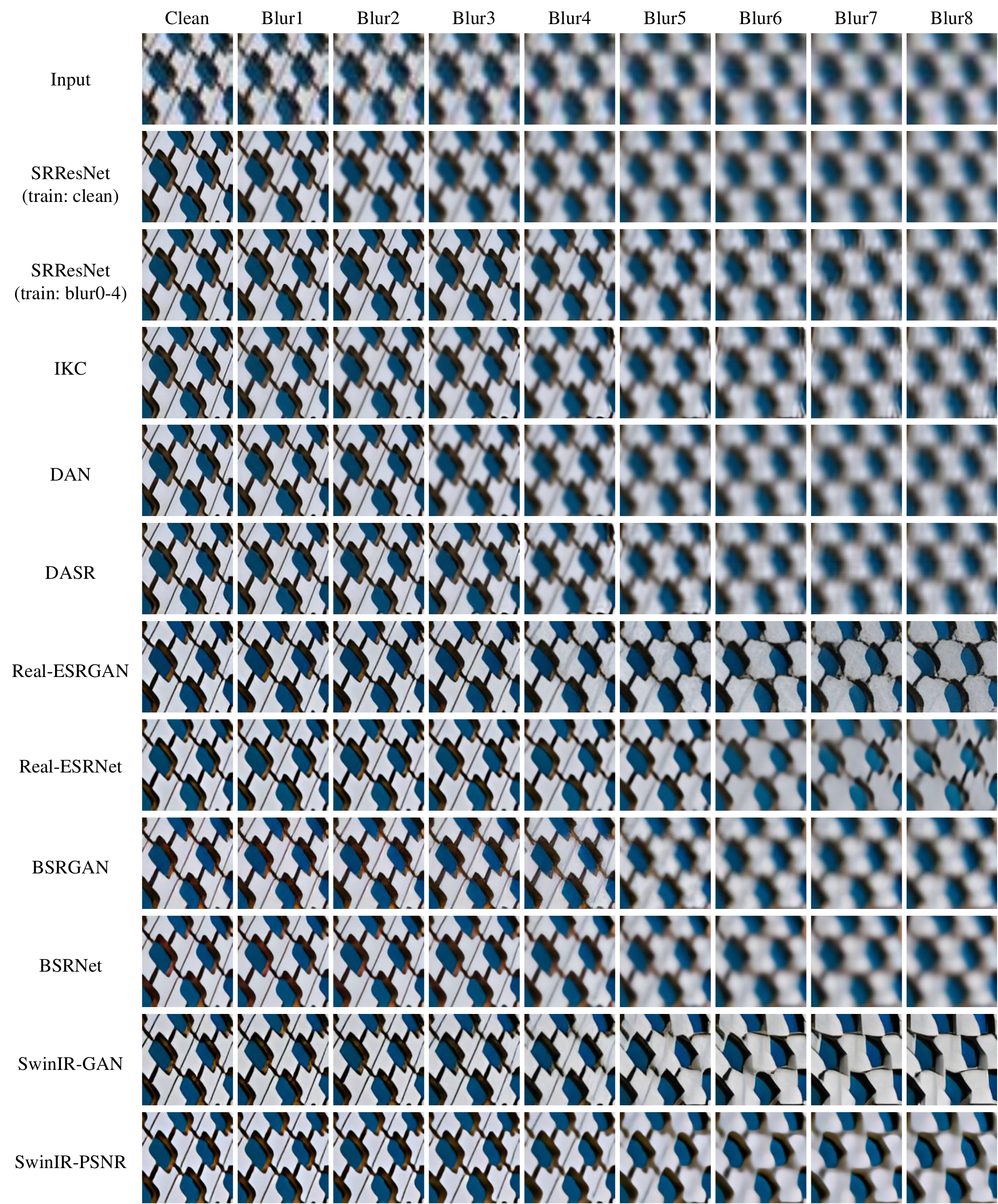}
	\end{center}
	\caption{Visual comparison on PIES-Blur dataset (2).}
	\label{fig:blur_img2}
\end{figure*}

\begin{figure*}[htbp]
	\begin{center}
		\includegraphics[width=0.95\linewidth]{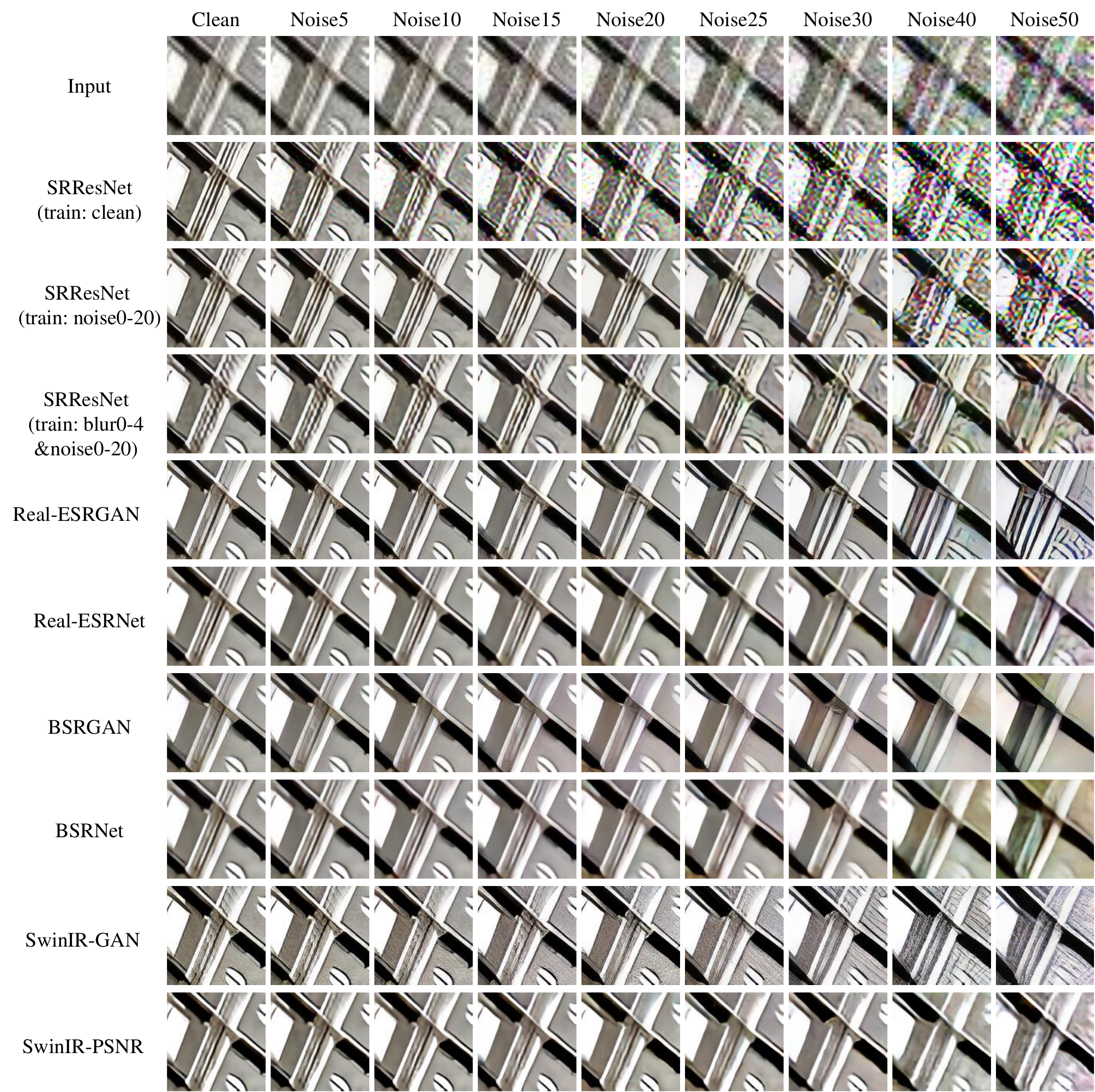}
	\end{center}
	\caption{Visual comparison on PIES-Noise dataset (1).}
	\label{fig:noise_img1}
\end{figure*}

\begin{figure*}[htbp]
	\begin{center}
		\includegraphics[width=0.95\linewidth]{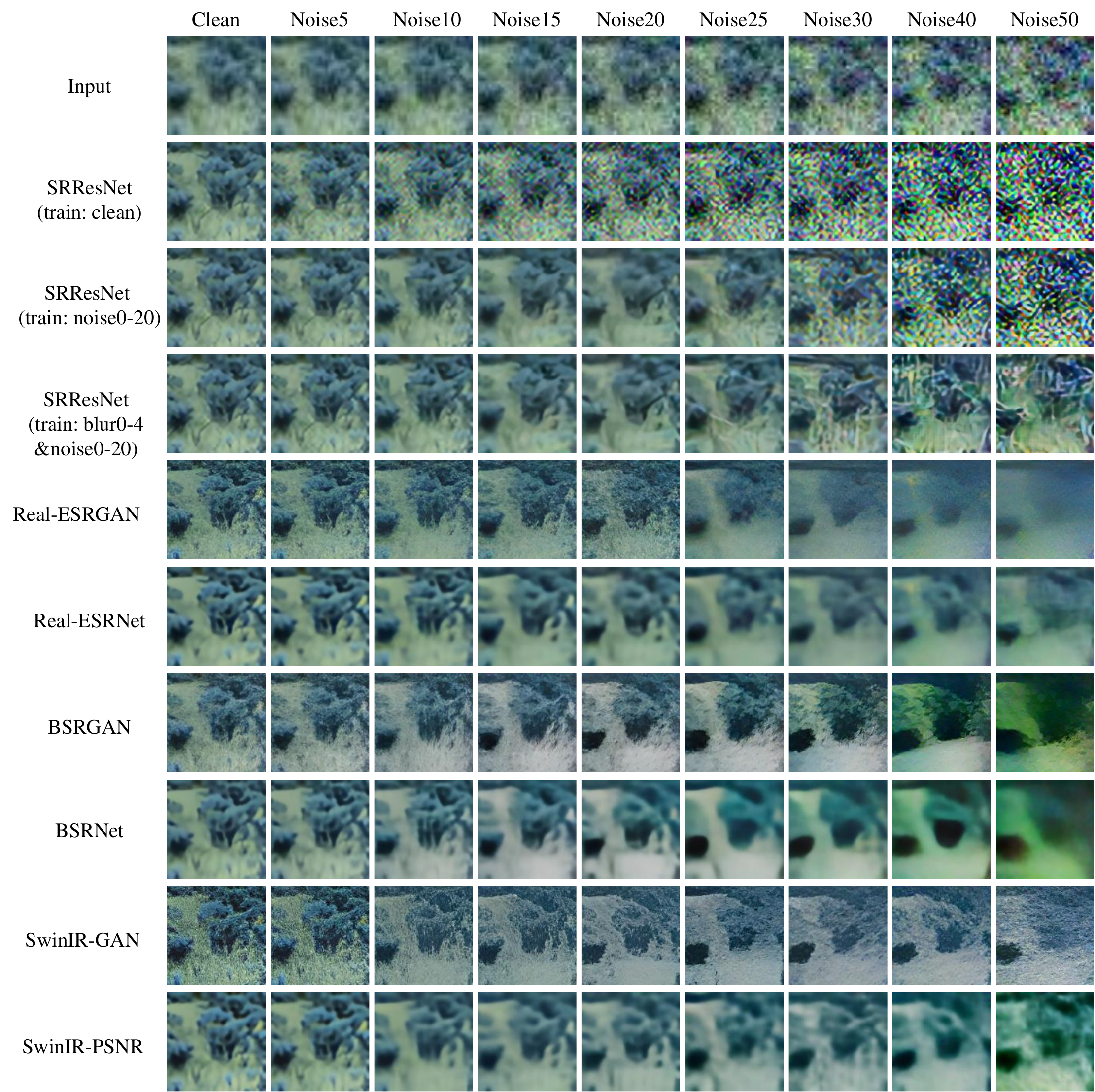}
	\end{center}
	\caption{Visual comparison on PIES-Noise dataset (2).}
	\label{fig:noise_img2}
\end{figure*}

\begin{figure*}[htbp]
	\begin{center}
		\includegraphics[width=0.95\linewidth]{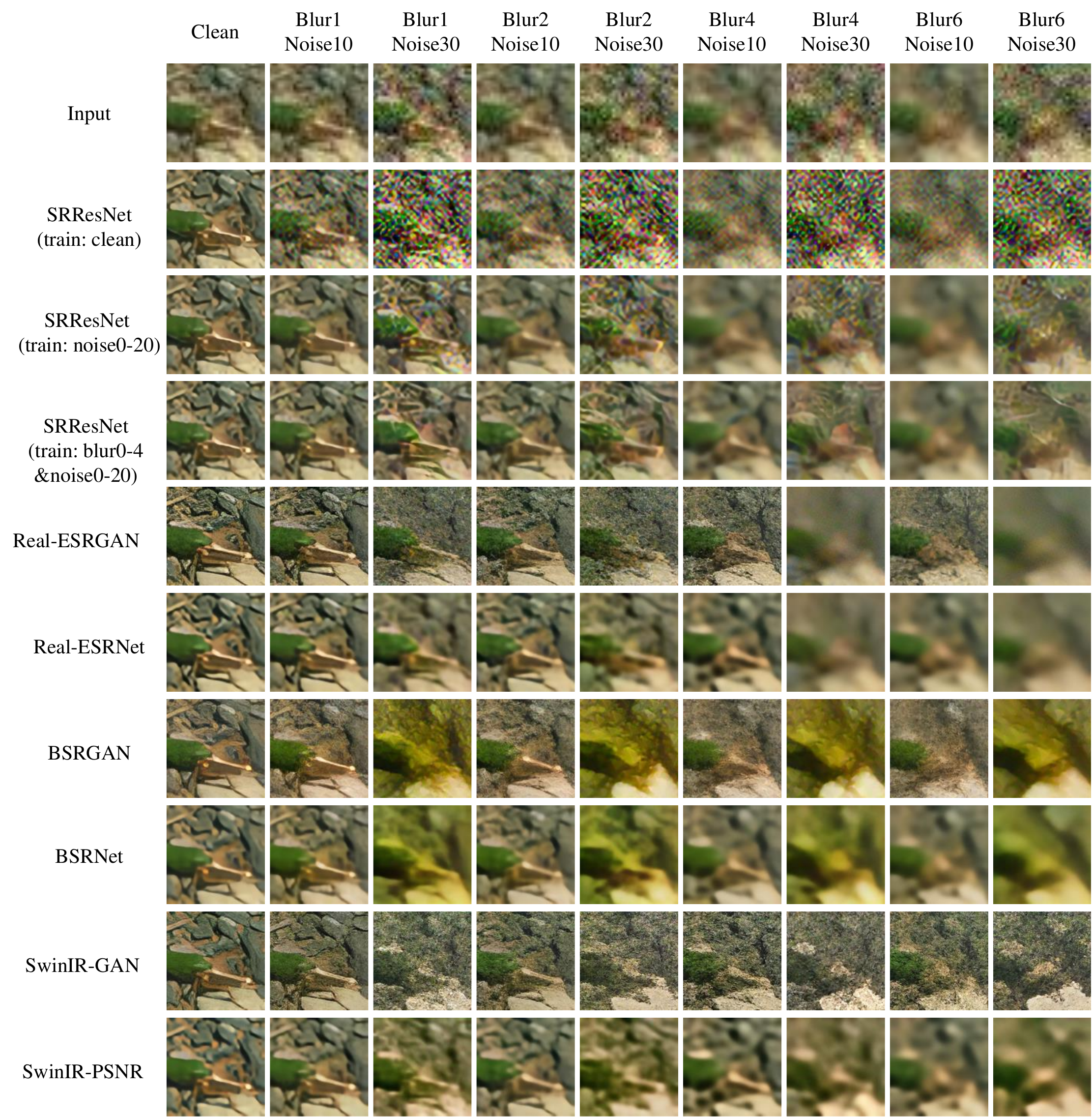}
	\end{center}
	\caption{Visual comparison on PIES-BlurNoise dataset.}
	\label{fig:blurnoise_img}
\end{figure*}

The visual results of the output images produced by representative SR models are shown in Fig. \ref{fig:blur_img1}, \ref{fig:blur_img2}, \ref{fig:noise_img1}, \ref{fig:noise_img2}, \ref{fig:blurnoise_img}, \ref{fig:realcam_img}, and \ref{fig:reallq_img}. From the visual results, we can see that Real-ESRGAN, BSRGAN, SwinIR-GAN achieve relatively better performance with visually pleasing output images.

\section{Samples of PIES Dataset}
\begin{figure*}[htbp]
	\begin{center}
		\includegraphics[width=0.9\linewidth,height=1.2\linewidth]{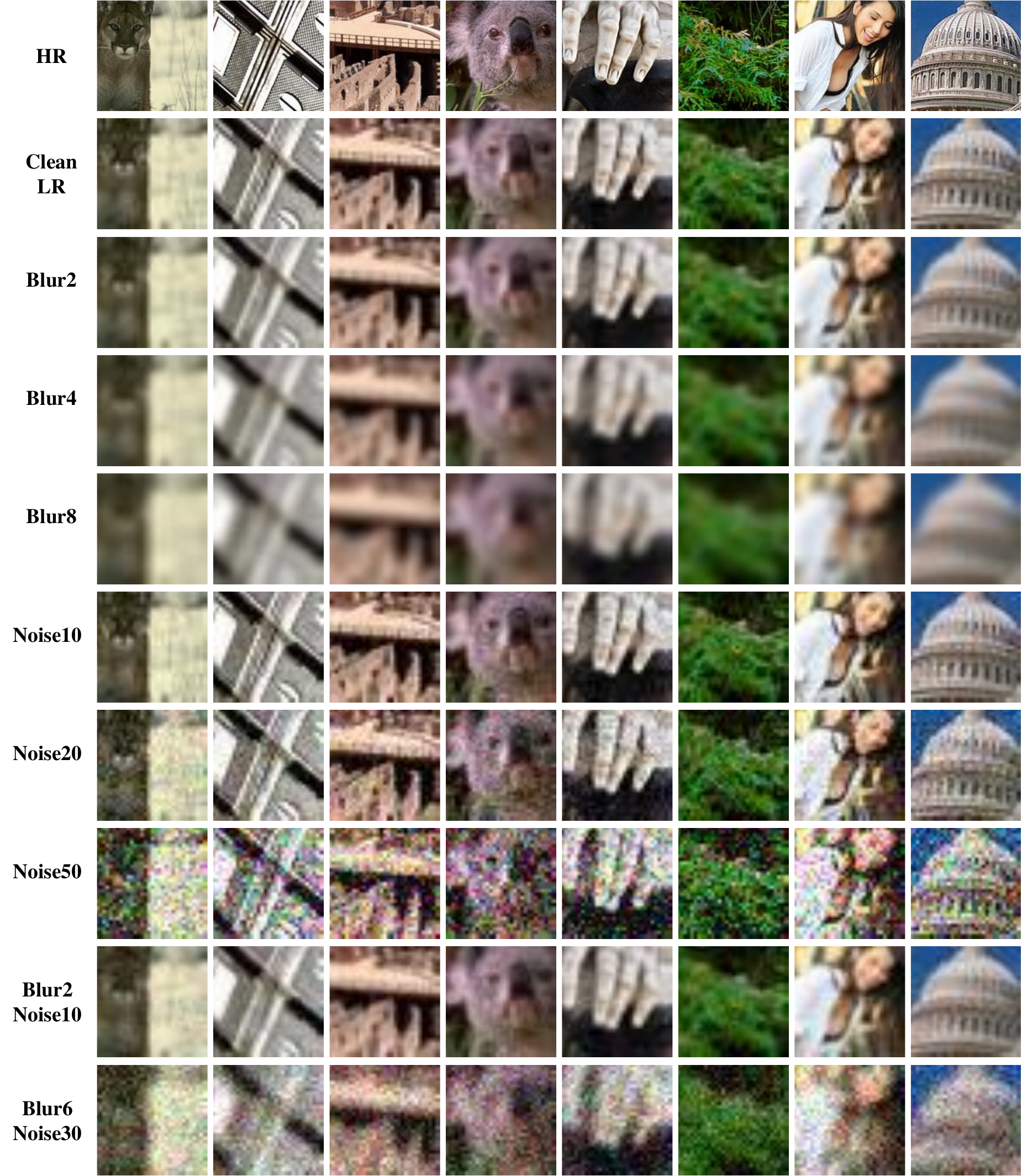}
	\end{center}
	\caption{Samples of PIES-Blur, PIES-Noise and PIES-BlurNoise datasets.}
	\label{fig:sample_PIES1}
\end{figure*}

\begin{figure*}[htbp]
	\begin{center}
		\includegraphics[width=0.95\linewidth]{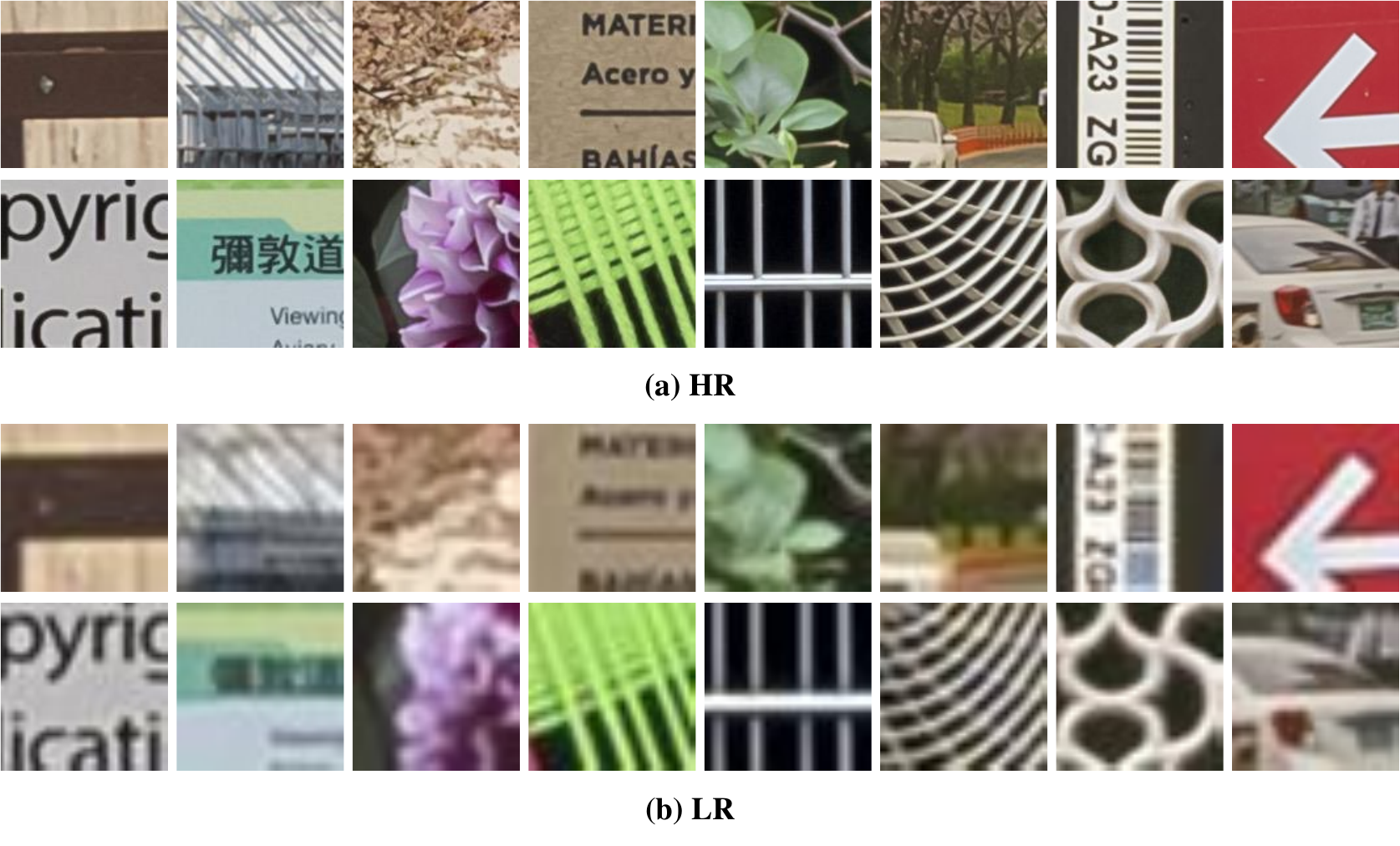}
	\end{center}
	\caption{Samples of PIES-RealCam dataset.}
	\label{fig:sample_PIES2}
\end{figure*}

\begin{figure*}[htbp]
	\begin{center}
		\includegraphics[width=0.95\linewidth]{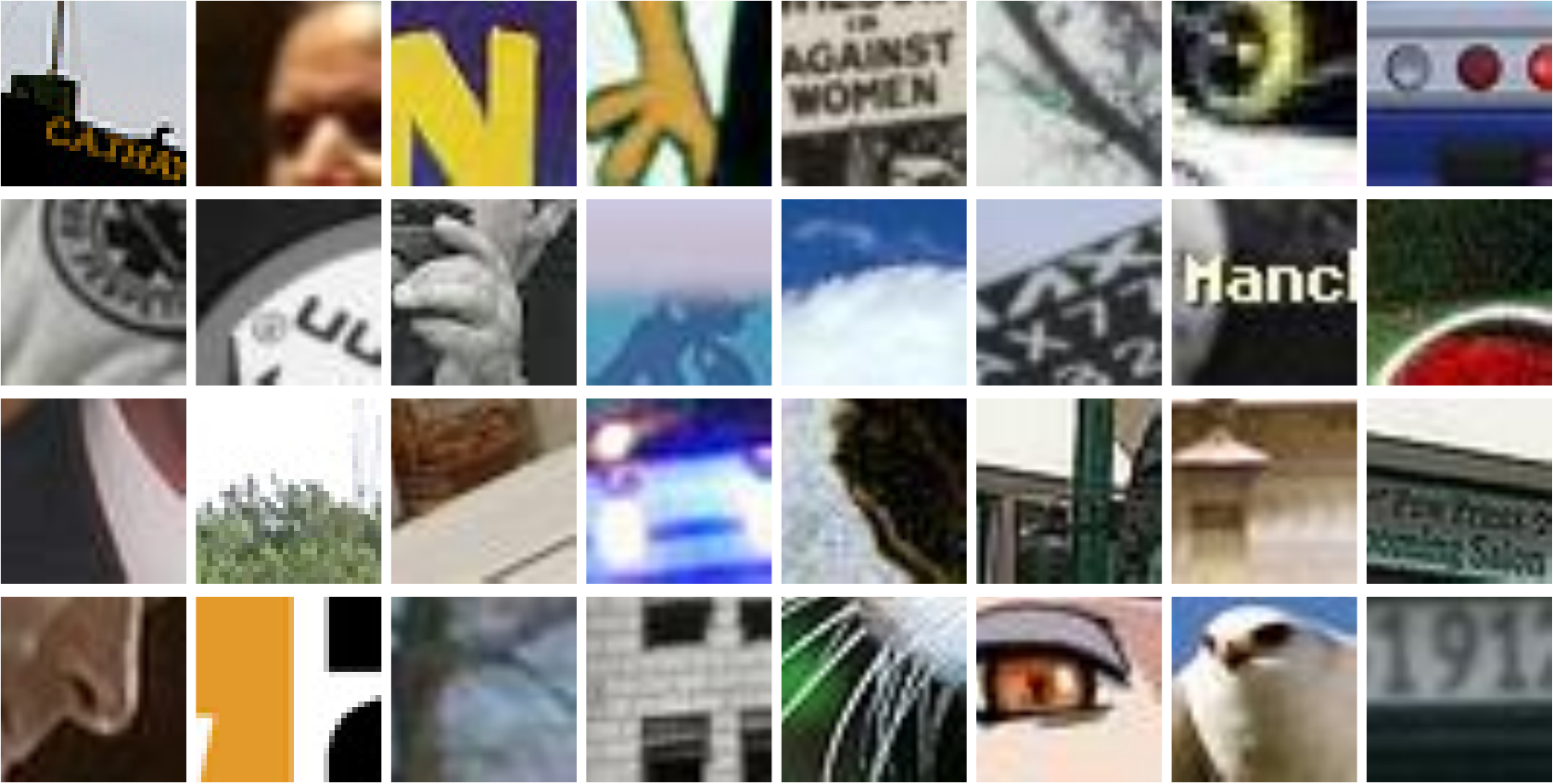}
	\end{center}
	\caption{Samples of PIES-RealLQ dataset.}
	\label{fig:sample_PIES3}
\end{figure*}

Fig. \ref{fig:sample_PIES1}, \ref{fig:sample_PIES2}, and \ref{fig:sample_PIES3} show some samples of PIES dataset, including PIES-Blur, PIES-Noise, PIES-BlurNoise, PIES-RealCam and PIES-RealLQ. It is easy to create your own datasets to test the model generalization ability using SRGA. The proposed PIES dataset only provide a unified platform to evaluate the model. The proposed metric is not restricted to this dataset.

\begin{table*}[htbp]
	\centering
	\caption{Model generalization ability (SRGA) on PIES-Blur dataset. Small SRGA value denotes better generalization.}
	\resizebox{\textwidth}{20mm}{
		\begin{tabular}{lccccccccccccccccc}
			\hline
			\multicolumn{1}{c}{\multirow{2}{*}{Methods}} & 
			\multicolumn{17}{c}{Blur \quad Level}\\ \cline{3-18}
			& Clean  & 0.5 & 1.0 & 1.5 &2.0 & 2.5 & 3.0 & 3.5 & 4.0  & 4.5 & 5.0 & 5.5 & 6.0 & 6.5 & 7.0 & 7.5 & 8.0 \\ \hline
			SRResNet (train: clean)               & 0.000 	& 1.204 &	2.489 &	3.024 &	3.325 &	3.532 &	3.705 &	3.835 &	3.938 &	4.014 &	4.070 &	4.118 &	4.151 &	4.177 &	4.196 &	4.213 &	4.229 \\
			SRResNet (train: blur0\_4)            & 0.000 &	0.903 &	1.690 &	1.633 &	1.568 &	1.681 &	2.225 &	2.930 &	3.358 &	3.607 &	3.772 &	3.880 &	3.958 &	4.011 &	4.057 &	4.089 &	4.111 \\
			SRResNet (train: blur2)               & 4.124 &	4.054 &	3.758 &	3.156 &	0.000 &	2.997 &	3.403 &	3.642 &	3.798 &	3.910 &	3.989 &	4.051 &	4.096 &	4.128 &	4.154 &	4.173 &	4.190 \\
			SRResNet (train: blur0\_4+noise0\_20) & 0.000 &	0.778 &	1.886 &	1.892 &	1.477 &	1.653 &	2.164 &	2.831 &	3.349 &	3.641 &	3.809 &	3.911 &	3.978 &	4.026 &	4.057 &	4.086 &	4.105 \\
			IKC                                   & 0.000 &	1.672 &	2.810 &	3.181 &	3.231 &	3.216 &	3.116 &	3.056 &	2.975 &	3.184 &	3.582 &	3.854 &	4.031 &	4.125 &	4.177 &	4.211 &	4.231 \\
			DAN                                   & 0.000 &	1.362 &	2.049 &	2.134 &	2.471 &	2.919 &	3.367 &	3.750 &	3.987 &	4.126 &	4.210 &	4.274 &	4.323 &	4.362 &	4.390 &	4.407 &	4.419 \\
			DASR                                  & 0.000 &	1.041 &	1.881 &	1.987 &	2.199 &	2.511 &	2.878 &	3.202 &	3.520 &	3.791 &	3.944 &	4.048 &	4.120 &	4.167 &	4.208 &	4.226 &	4.245 \\
			Real-ESRGAN                           & 0.000 &	0.301 &	1.114 &	1.672 &	2.025 &	2.279 &	2.458 &	2.612 &	2.722 &	2.836 &	2.934 &	3.015 &	3.068 &	3.120 &	3.151 &	3.180 &	3.192 \\
			Real-ESRNet                           & 0.000 &	0.000 &	0.778 &	1.322 &	1.763 &	2.000 &	2.127 &	2.281 &	2.480 &	2.725 &	2.866 &	3.052 &	3.170 &	3.187 &	3.191 &	3.206 &	3.230 \\
			BSRGAN                                & 0.000 &	0.000 &	0.903 &	1.556 &	2.041 &	2.356 &	2.610 &	2.814 &	2.947 &	3.061 &	3.147 &	3.217 &	3.280 &	3.333 &	3.373 &	3.403 &	3.437 \\
			BSRNet                                & 0.000 &	0.000 &	0.845 &	1.447 &	1.903 &	2.241 &	2.496 &	2.719 &	2.905 &	3.056 &	3.181 &	3.294 &	3.387 &	3.459 &	3.516 &	3.549 &	3.570 \\
			SwinIR-GAN                            & 0.000 &	0.000 &	0.477 &	0.778 &	1.204 &	1.556 &	1.748 &	1.929 &	2.037 &	2.090 &	2.124 &	2.121 &	2.107 &	2.090 &	2.049 &	1.987 &	1.934 \\
			SwinIR-PSNR                           & 0.000 &	0.000 &	0.000 &	0.477 &	0.954 &	1.301 &	1.531 &	1.778 &	1.914 &	2.053 &	2.143 &	2.260 &	2.365 &	2.433 &	2.471 &	2.498 &	2.504 \\ \hline
	\end{tabular}}
	
	\label{tab:blur_srga}
\end{table*}

\begin{table*}[htbp]
	\centering
	\caption{Model generalization ability (SRGA) on PIES-Noise dataset. Small SRGA value denotes better generalization.}
	\resizebox{\textwidth}{20mm}{
		
		\begin{tabular}{lccccccccccc}
			\hline
			\multicolumn{1}{c}{\multirow{2}{*}{Methods}} & 
			\multicolumn{11}{c}{Noise \quad Level}\\ \cline{3-12}
			& Clean & 5 & 10 & 15 & 20 & 25 & 30 & 35 & 40 & 45 & 50 \\ \hline
			SRResNet (train: clean)               & 0.000 &	1.000 &	2.579 &	3.368 &	3.826 &	4.108 &	4.291 &	4.417 &	4.501 &	4.564 &	4.612  \\
			SRResNet (train: noise0\_20)          & 0.000 &	0.301 &	0.954 &	1.398 &	1.591 &	1.362 &	2.161 &	3.323 &	3.818 &	4.087 &	4.253  \\
			SRResNet (train: blur0\_4+noise0\_20) & 0.000 &	0.172 &	0.077 &	0.778 &	0.699 &	0.477 &	1.857 &	2.534 &	2.900 &	3.111 &	3.226  \\
			Real-ESRGAN                           & 0.000 &	0.903 &	1.279 &	1.255 &	1.580 &	1.643 &	1.839 &	2.083 &	2.312 &	2.456 &	2.562   \\
			Real-ESRNet                           & 0.000 &	0.602 &	1.041 &	1.255 &	1.447 &	1.690 &	1.778 &	1.845 &	1.996 &	1.959 &	2.049   \\
			BSRGAN                                & 0.000 &	0.602 &	1.415 &	1.756 &	1.982 &	2.049 &	2.228 &	2.354 &	2.387 &	2.358 &	2.328  \\
			BSRNet                                & 0.000 &	0.903 &	1.708 &	2.057 &	2.288 &	2.371 &	2.498 &	2.571 &	2.648 &	2.700 &	2.800  \\
			SwinIR-GAN                            & 0.000 &	0.301 &	1.681 &	2.248 &	2.461 &	2.533 &	2.494 &	2.417 &	2.182 &	1.964 &	1.681  \\
			SwinIR-PSNR                           & 0.000 &	0.477 &	1.114 &	1.301 &	1.491 &	1.643 &	1.857 &	2.049 &	2.179 &	2.320 &	2.423  \\ \hline
	\end{tabular}}
	
	\label{tab:noise_srga}
\end{table*}

\begin{table*}[htbp]
	\centering
	\caption{Model generalization ability (SRGA) on PIES-BlurNoise dataset. Small SRGA value denotes better generalization.}
	\resizebox{\textwidth}{20mm}{
		
		\begin{tabular}{lccccccccccccc}
			\hline
			\multicolumn{1}{c}{\multirow{2}{*}{Methods}} &       & \multicolumn{3}{c}{Blur1} & \multicolumn{3}{c}{Blur2} & \multicolumn{3}{c}{Blur4} & \multicolumn{3}{c}{Blur6} \\ \cline{3-14} 
			\multicolumn{1}{c}{}                         & Clean & 10   & 20  & 30  & 10   & 20  & 30  & 10   & 20  & 30  & 10   & 20  & 30  \\ \hline
			SRResNet (train: clean)                       & 0.000 &	1.771 &	3.744 &	4.282 &	1.934 &	3.665 &	4.260 &	3.037 &	3.535 &	4.239 &	3.266 &	3.476 &	4.224   \\
			SRResNet (train: blur0\_4)                   & 0.000 &	1.114 &	3.632 &	4.184 &	2.297 &	3.535 &	4.161 &	3.179 &	3.361 &	4.125 &	3.412 &	3.279 &	4.115   \\
			SRResNet (train: noise0\_20)                 & 0.000 &	2.459 &	2.539 &	1.079 &	3.253 &	3.261 &	2.630 &	3.833 &	3.827 &	3.427 &	4.037 &	4.030 &	3.630   \\
			SRResNet (train: blur0\_4+noise0\_20)        & 0.000 &	1.964 &	2.061 &	0.954 &	2.307 &	2.584 &	2.797 &	3.299 &	3.534 &	3.642 &	3.870 &	3.920 &	3.874   \\
			Real-ESRGAN                                  & 0.000 &	1.875 &	1.857 &	1.633 &	2.442 &	2.452 &	2.382 &	3.039 &	3.114 &	3.232 &	3.270 &	3.419 &	3.585   \\
			Real-ESRNet                                  & 0.000 &	1.663 &	2.025 &	2.253 &	2.332 &	2.624 &	2.868 &	3.093 &	3.383 &	3.575 &	3.448 &	3.687 &	3.852   \\
			BSRGAN                                       & 0.000 &	1.892 &	2.389 &	2.569 &	2.425 &	2.668 &	2.727 &	3.121 &	3.177 &	3.135 &	3.378 &	3.379 &	3.339   \\
			BSRNet                                       & 0.000 &	1.982 &	2.490 &	2.654 &	2.450 &	2.805 &	2.885 &	3.134 &	3.288 &	3.339 &	3.468 &	3.520 &	3.554  \\
			SwinIR-GAN                                   & 0.000 &	1.863 &	2.548 &	2.520 &	2.104 &	2.611 &	2.550 &	2.444 &	2.649 &	2.531 &	2.330 &	2.515 &	2.554  \\
			SwinIR-PSNR                                  & 0.000 &	1.322 &	1.806 &	2.004 &	1.839 &	2.086 &	2.301 &	2.358 &	2.604 &	2.814 &	2.550 &	2.825 &	3.053   \\ \hline
	\end{tabular}}
	
	\label{tab:blurnoise_srga}
\end{table*}

\begin{table*}[]
	\centering
	\caption{Model generalization ability (SRGA) on PIES-RealCam and PIES-RealLQ datasets. Small SRGA value denotes better generalization.}
	\label{tab:real_srga}
	\begin{tabular}{lccc}
		\hline
		\multicolumn{1}{c}{Methods}                 & PIES-Clean & PIES-RealCam & PIES-RealLQ \\ \hline
		SRResNet (train: clean)               & 0.000          & 3.563 &	3.825 \\
		SRResNet (train: blur0\_4)            & 0.000          & 3.125 &	3.811 \\
		SRResNet (train: noise0\_20)          & 0.000          & 3.547 &	3.827 \\
		SRResNet (train: blur0\_4+noise0\_20) & 0.000          & 2.566 &	3.736 \\
		IKC                                   & 0.000          & 3.432 &	3.845 \\
		DAN                                   & 0.000          & 3.937 &	4.070 \\
		DASR                                  & 0.000          & 3.770 &	4.033 \\
		Real-ESRGAN                           & 0.000          & 3.301 &	3.823 \\
		Real-ESRNet                           & 0.000          & 2.787 &	3.770 \\
		BSRGAN                                & 0.000          & 2.872 &	3.796 \\
		BSRNet                                & 0.000          & 2.686 &	4.345 \\
		SwinIR-GAN                            & 0.000          & 3.379 &	3.662 \\
		SwinIR-PSNR                           & 0.000          & 2.826 &	3.655 \\ \hline
	\end{tabular}
	
\end{table*}

\begin{table*}
	\centering
	\caption{Model performance (PNSR, SSIM, NIQE) and model generalization ability (SRGA) on PIES-AnisoBlur dataset.}
	\label{tab:aniso_blur}
	\begin{tabular}{lcccc}
		\hline
		Methods                                     & PSNR & SSIM & NIQE & SRGA \\ \hline
		SRResNet (train: clean)               & 21.54           & 0.5692 & 15.14 & 3.615          \\
		SRResNet (train: blur0\_4)            & 23.24           & 0.6564 & 15.27 & 2.880          \\
		SRResNet (train: noise0\_20)          & 21.52           & 0.5665 & 15.18 & 3.605          \\
		SRResNet (train: blur0\_4+noise0\_20) & 23.19           & 0.6513 & 15.48 & 2.820          \\
		IKC                                         & 23.16     & 0.6537 & 16.23 & 3.375          \\
		DAN                                         & 23.43           & 0.6636 & 15.80 & 3.310          \\
		DASR                                 & 23.49           & 0.6658 & 15.92 & 3.157                     \\
		Real-ESRGAN                                 & 21.31           & 0.5771 & 10.84 & 2.500          \\
		Real-ESRNet                                 & 22.46           & 0.6247 & 15.32 & 2.456          \\
		BSRGAN                                      & 21.89           & 0.5831 & 9.93 & 2.397          \\
		BSRNet                                      & 22.98           & 0.6315 & 15.91 & 2.339          \\
		SwinIR-GAN                                  & 21.21           & 0.5762 & 10.36 & 1.852          \\
		SwinIR-PSNR                                 & 22.57           & 0.6347 & 15.16 & 1.727         \\ \hline
	\end{tabular}
	
\end{table*}

\begin{table*}[htbp]
	\centering
	\caption{Model performance (PSNR) on PIES-Blur dataset. Higher PSNR value denotes that the output images are closer to the ground truth images in content.}
	\label{tab:blur_psnr}
	\resizebox{\textwidth}{20mm}{
		\begin{tabular}{lccccccccccccccccc}
			\hline
			
			\multicolumn{1}{c}{\multirow{2}{*}{Methods}} & 
			\multicolumn{17}{c}{Blur \quad Level}\\ \cline{3-18}
			
			& Clean  & 0.5 & 1.0 & 1.5 &2.0 & 2.5 & 3.0 & 3.5 & 4.0  & 4.5 & 5.0 & 5.5 & 6.0 & 6.5 & 7.0 & 7.5 & 8.0
			
			\\ \hline
			
			SRResNet (train: clean) & 25.87 & 25.82 & 24.99 & 23.74 & 22.69 & 21.85 & 21.18 & 20.63 & 20.19 & 19.84 & 19.56 & 19.34 & 19.17 & 19.03 & 18.91 & 18.82 & 18.74   \\ 
			SRResNet (train: blur0\_4)   & 25.48 & 25.52 & 25.47 & 25.43 & 25.37 & 25.19 & 24.82 & 23.83 & 22.49 & 21.31 & 20.46 & 19.89 & 19.51 & 19.26 & 19.09 & 18.97 & 18.88   \\ 
			SRResNet (train: blur2)      & 16.20 & 16.69 & 18.80 & 22.48 & 25.80 & 23.74 & 22.27 & 21.34 & 20.69 & 20.21 & 19.85 & 19.57 & 19.36 & 19.19 & 19.05 & 18.94 & 18.85   \\ 
			SRResNet (train: blur0\_4+noise0\_20) & 25.10 & 25.15 & 25.08 & 24.95 & 24.87 & 24.70 & 24.38 & 23.75 & 22.50 & 21.32 & 20.51 & 19.97 & 19.61 & 19.34 & 19.15 & 19.00 & 18.89   \\ 
			IKC                                                                                   & 25.50 & 25.67 & 25.63 & 25.26 & 24.96 & 24.52 & 23.84 & 22.80 & 21.66 & 20.77 & 20.21 & 19.84 & 19.59 & 19.38 & 19.21 & 19.08 & 18.97   \\ 
			DAN                                                                                   & 25.94 & 25.95 & 25.90 & 25.86 & 25.59 & 24.88 & 23.70 & 22.39 & 21.28 & 20.56 & 20.09 & 19.75 & 19.48 & 19.27 & 19.12 & 18.99 & 18.90   \\ 
			DASR                                                                                & 25.67 & 25.70 & 25.70 & 25.65 & 25.49 & 25.08 & 24.35 & 23.40 & 22.27 & 21.19 & 20.47 & 19.99 & 19.66 & 19.41 & 19.23 & 19.09 & 18.97   \\
			Real-ESRGAN                                                                                 & 21.90 & 21.92 & 21.99 & 22.03 & 22.01 & 21.89 & 21.66 & 21.30 & 20.87 & 20.58 & 20.41 & 20.36 & 20.32 & 20.27 & 20.16 & 20.04 & 19.93   \\ 
			Real-ESRNet                                                                                 & 23.41 & 23.43 & 23.49 & 23.53 & 23.50 & 23.31 & 22.98 & 22.53 & 22.05 & 21.65 & 21.35 & 21.27 & 21.30 & 21.22 & 21.02 & 20.80 & 20.60   \\ 
			BSRGAN                                                                                     & 22.54 & 22.56 & 22.60 & 22.60 & 22.53 & 22.36 & 22.13 & 21.83 & 21.51 & 21.17 & 20.84 & 20.55 & 20.29 & 20.05 & 19.85 & 19.68 & 19.53   \\ 
			BSRNet                                                                                     & 23.79 & 23.80 & 23.83 & 23.80 & 23.67 & 23.43 & 23.08 & 22.66 & 22.19 & 21.72 & 21.31 & 20.97 & 20.70 & 20.48 & 20.30 & 20.17 & 20.06   \\ 
			
			SwinIR-GAN                                                                                 & 22.06 & 22.07 & 22.06 & 22.01 & 21.87 & 21.66 & 21.39 & 21.06 & 20.71 & 20.38 & 20.05 & 19.73 & 19.42 & 19.14 & 18.90 & 18.71 & 18.56   \\ 
			SwinIR-PSNR                                                                                & 23.62 & 23.63 & 23.63 & 23.57 & 23.43 & 23.21 & 22.89 & 22.50 & 22.07 & 21.64 & 21.25 & 20.91 & 20.62 & 20.41 & 20.28 & 20.20 & 20.13   \\ \hline
	\end{tabular}}
	
\end{table*}

\begin{table*}[htbp]
	\centering
	\caption{Model performance (SSIM) on PIES-Blur dataset. Higher SSIM value denotes that the output images are structurally closer to the ground truth images.}
	\label{tab:blur_ssim}
	\resizebox{\textwidth}{20mm}{
		\begin{tabular}{lccccccccccccccccc}
			\hline
			\multicolumn{1}{c}{\multirow{2}{*}{Methods}} & 
			\multicolumn{17}{c}{Blur \quad Level}\\ \cline{3-18}
			& Clean  & 0.5 & 1.0 & 1.5 &2.0 & 2.5 & 3.0 & 3.5 & 4.0  & 4.5 & 5.0 & 5.5 & 6.0 & 6.5 & 7.0 & 7.5 & 8.0
			\\ \hline
			SRResNet (train: clean)                                                           & 0.7569 & 0.7531 & 0.7274 & 0.6789 & 0.6278 & 0.5828 & 0.5452 & 0.5149 & 0.4912 & 0.4729 & 0.4589 & 0.4481 & 0.4396 & 0.4329 & 0.4276 & 0.4232 & 0.4196 \\
			SRResNet (train: blur0\_4)                                                            & 0.7495 & 0.7482 & 0.7425 & 0.7396 & 0.7357 & 0.7266 & 0.7063 & 0.6640 & 0.6043 & 0.5448 & 0.4990 & 0.4684 & 0.4487 & 0.4364 & 0.4286 & 0.4229 & 0.4185 \\
			SRResNet (train: blur2) & 0.4559 & 0.4804 & 0.5755 & 0.7022 & 0.7497 & 0.6745 & 0.6024 & 0.5520 & 0.5161 & 0.4900 & 0.4709 & 0.4567 & 0.4458 & 0.4373 & 0.4307 & 0.4253 & 0.4210 \\
			SRResNet (train: blur0\_4+noise0\_20) & 0.7377 & 0.7364 & 0.7288 & 0.7234 & 0.7215 & 0.7103 & 0.6894 & 0.6553 & 0.6002 & 0.5435 & 0.5013 & 0.4724 & 0.4525 & 0.4385 & 0.4283 & 0.4208 & 0.4150 \\
			IKC                                                                              & 0.7583 & 0.7585 & 0.7494 & 0.7331 & 0.7231 & 0.7102 & 0.6895 & 0.6509 & 0.6018 & 0.5477 & 0.4992 & 0.4677 & 0.4488 & 0.4365 & 0.4275 & 0.4209 & 0.4157 \\
			DAN                                                                                              & 0.7629 & 0.7616 & 0.7581 & 0.7556 & 0.7477 & 0.7251 & 0.6765 & 0.6084 & 0.5471 & 0.5061 & 0.4802 & 0.4623 & 0.4491 & 0.4393 & 0.4321 & 0.4264 & 0.4218 \\
			DASR   & 0.7559 & 0.7548 & 0.7515 & 0.7485 & 0.7425 & 0.7294 & 0.7024 & 0.6596 & 0.6027 & 0.5418 & 0.4986 & 0.4700 & 0.4513 & 0.4388 & 0.4300 & 0.4237 & 0.4184 \\
			Real-ESRGAN              & 0.6271 & 0.6268 & 0.6254 & 0.6219 & 0.6151 & 0.6042 & 0.5876 & 0.5635 & 0.5371 & 0.5186 & 0.5088 & 0.5063 & 0.5049 & 0.5021 & 0.4968 & 0.4910 & 0.4853 \\
			Real-ESRNet             & 0.6795 & 0.6792 & 0.6781 & 0.6750 & 0.6684 & 0.6563 & 0.6390 & 0.6157 & 0.5891 & 0.5660 & 0.5503 & 0.5449 & 0.5464 & 0.5442 & 0.5358 & 0.5266 & 0.5178 \\
			BSRGAN                                                                                           & 0.6321 & 0.6320 & 0.6306 & 0.6264 & 0.6180 & 0.6055 & 0.5899 & 0.5725 & 0.5549 & 0.5366 & 0.5193 & 0.5038 & 0.4898 & 0.4771 & 0.4661 & 0.4567 & 0.4487 \\
			BSRNet                                                                                           & 0.6814 & 0.6810 & 0.6788 & 0.6734 & 0.6636 & 0.6493 & 0.6314 & 0.6106 & 0.5878 & 0.5650 & 0.5446 & 0.5278 & 0.5139 & 0.5025 & 0.4935 & 0.4868 & 0.4817 \\
			SwinIR-GAN                                                                                       & 0.6321 & 0.6317 & 0.6289 & 0.6226 & 0.6118 & 0.5965 & 0.5780 & 0.5579 & 0.5377 & 0.5189 & 0.5007 & 0.4837 & 0.4673 & 0.4532 & 0.4420 & 0.4337 & 0.4277 \\
			SwinIR-PSNR                                                                                      & 0.6922 & 0.6917 & 0.6890 & 0.6831 & 0.6728 & 0.6579 & 0.6396 & 0.6192 & 0.5979 & 0.5770 & 0.5578 & 0.5407 & 0.5257 & 0.5153 & 0.5096 & 0.5060 & 0.5038 \\
			\hline
	\end{tabular}}
	
\end{table*}

\begin{table*}[]
	\centering
	\caption{Model performance (LPIPS) on PIES-Blur dataset. Lower LPIPS value denotes that the output images are perceptually closer to the ground truth images.}
	\label{tab:blur_lpips}
	\resizebox{\textwidth}{20mm}{
		\begin{tabular}{lccccccccccccccccc}
			\hline
			\multicolumn{1}{c}{\multirow{2}{*}{Methods}} & 
			\multicolumn{17}{c}{Blur \quad Level}\\ \cline{3-18}
			& Clean  & 0.5 & 1.0 & 1.5 &2.0 & 2.5 & 3.0 & 3.5 & 4.0  & 4.5 & 5.0 & 5.5 & 6.0 & 6.5 & 7.0 & 7.5 & 8.0 \\ \hline
			SRResNet (train: clean)               & 0.2136 & 0.2216 & 0.2669 & 0.3511 & 0.4339 & 0.4979 & 0.5476 & 0.5860 & 0.6150 & 0.6359 & 0.6503 & 0.6601 & 0.6666 & 0.6709 & 0.6738 & 0.6758 & 0.6770 \\
			SRResNet (train: blur0\_4)            & 0.2202 & 0.2233 & 0.2320 & 0.2324 & 0.2345 & 0.2436 & 0.2666 & 0.3234 & 0.4173 & 0.5154 & 0.5738 & 0.5957 & 0.6062 & 0.6141 & 0.6207 & 0.6248 & 0.6267 \\
			SRResNet (train: blur2)               & 0.3379 & 0.3217 & 0.2669 & 0.2130 & 0.2186 & 0.3487 & 0.4637 & 0.5336 & 0.5811 & 0.6135 & 0.6343 & 0.6470 & 0.6543 & 0.6582 & 0.6597 & 0.6600 & 0.6592 \\
			SRResNet (train: blur0\_4+noise0\_20) & 0.2299 & 0.2337 & 0.2471 & 0.2503 & 0.2464 & 0.2594 & 0.2891 & 0.3363 & 0.4160 & 0.5073 & 0.5688 & 0.5968 & 0.6078 & 0.6109 & 0.6103 & 0.6088 & 0.6066 \\
			IKC                                   & 0.2058 & 0.2100 & 0.2321 & 0.2564 & 0.2693 & 0.2837 & 0.3016 & 0.3442 & 0.4003 & 0.4719 & 0.5287 & 0.5530 & 0.5646 & 0.5681 & 0.5674 & 0.5658 & 0.5633 \\
			DAN                                   & 0.2042 & 0.2086 & 0.2178 & 0.2220 & 0.2334 & 0.2670 & 0.3418 & 0.4454 & 0.5344 & 0.5900 & 0.6194 & 0.6357 & 0.6457 & 0.6507 & 0.6526 & 0.6531 & 0.6518 \\
			DASR                                  & 0.2138 & 0.2169 & 0.2229 & 0.2262 & 0.2326 & 0.2471 & 0.2797 & 0.3346 & 0.4159 & 0.5077 & 0.5583 & 0.5802 & 0.5866 & 0.5881 & 0.5893 & 0.5889 & 0.5885 \\
			Real-ESRGAN                           & 0.1932 & 0.1935 & 0.1948 & 0.1975 & 0.2027 & 0.2115 & 0.2249 & 0.2418 & 0.2601 & 0.2759 & 0.2880 & 0.2953 & 0.2969 & 0.2973 & 0.2996 & 0.3022 & 0.3052 \\
			Real-ESRNet                           & 0.2843 & 0.2851 & 0.2881 & 0.2932 & 0.3020 & 0.3161 & 0.3361 & 0.3606 & 0.3865 & 0.4104 & 0.4268 & 0.4315 & 0.4255 & 0.4194 & 0.4194 & 0.4227 & 0.4270 \\
			BSRGAN                                & 0.1909 & 0.1910 & 0.1918 & 0.1949 & 0.2010 & 0.2104 & 0.2230 & 0.2390 & 0.2576 & 0.2781 & 0.2983 & 0.3160 & 0.3309 & 0.3427 & 0.3519 & 0.3588 & 0.3632 \\
			BSRNet                                & 0.2790 & 0.2800 & 0.2841 & 0.2917 & 0.3040 & 0.3206 & 0.3414 & 0.3661 & 0.3945 & 0.4237 & 0.4494 & 0.4706 & 0.4880 & 0.5018 & 0.5111 & 0.5162 & 0.5184 \\
			SwinIR-GAN                            & 0.1724 & 0.1728 & 0.1742 & 0.1778 & 0.1845 & 0.1947 & 0.2083 & 0.2223 & 0.2379 & 0.2527 & 0.2670 & 0.2806 & 0.2906 & 0.2987 & 0.3041 & 0.3076 & 0.3102 \\
			SwinIR-PSNR                           & 0.2671 & 0.2683 & 0.2724 & 0.2792 & 0.2897 & 0.3047 & 0.3234 & 0.3444 & 0.3667 & 0.3889 & 0.4092 & 0.4270 & 0.4413 & 0.4505 & 0.4547 & 0.4568 & 0.4580 \\ \hline
	\end{tabular}}
	
\end{table*}

\begin{table*}[]
	\centering
	\caption{Model performance (NIQE) on PIES-Blur dataset. Lower NIQE value denotes better perceptual quality.}
	\label{tab:blur_niqe}
	\resizebox{\textwidth}{20mm}{
		\begin{tabular}{lccccccccccccccccc}
			\hline
			\multicolumn{1}{c}{\multirow{2}{*}{Methods}} & 
			\multicolumn{17}{c}{Blur \quad Level}\\ \cline{3-18}
			& Clean  & 0.5 & 1.0 & 1.5 &2.0 & 2.5 & 3.0 & 3.5 & 4.0  & 4.5 & 5.0 & 5.5 & 6.0 & 6.5 & 7.0 & 7.5 & 8.0 \\ \hline
			SRResNet (train: clean)               & 12.28 & 12.39 & 13.64 & 14.84 & 14.90 & 14.52 & 14.73 & 15.03 & 15.34 & 15.68 & 16.01 & 16.28 & 16.54 & 16.71 & 16.84 & 16.91 & 16.98 \\
			SRResNet (train: blur0\_4)            & 12.20 & 12.31 & 12.67 & 12.69 & 12.67 & 13.00 & 13.13 & 14.09 & 17.64 & 15.72 & 14.84 & 14.59 & 14.68 & 14.76 & 14.95 & 14.98 & 15.01 \\
			SRResNet (train: blur2)               & 26.71 & 24.98 & 18.47 & 13.48 & 12.44 & 15.15 & 14.82 & 14.63 & 14.82 & 15.05 & 15.33 & 15.59 & 15.79 & 15.92 & 15.99 & 16.01 & 16.00 \\
			SRResNet (train: blur0\_4+noise0\_20) & 12.76 & 12.82 & 13.03 & 12.95 & 12.91 & 13.09 & 13.62 & 15.30 & 18.42 & 19.59 & 16.96 & 15.49 & 15.62 & 15.44 & 14.70 & 14.63 & 14.57 \\
			IKC                                   & 13.11 & 13.20 & 13.65 & 14.40 & 13.96 & 13.98 & 14.32 & 14.66 & 13.98 & 14.10 & 14.51 & 14.86 & 14.81 & 14.79 & 14.71 & 14.68 & 14.65 \\
			DAN                                   & 12.09 & 12.20 & 12.39 & 12.40 & 12.66 & 13.47 & 14.17 & 14.78 & 14.49 & 14.90 & 15.21 & 15.60 & 15.95 & 16.15 & 16.24 & 16.29 & 16.26 \\
			DASR                                  & 11.64 & 11.80 & 12.05 & 12.15 & 12.41 & 12.78 & 13.64 & 14.48 & 14.71 & 13.69 & 13.60 & 13.83 & 13.94 & 14.13 & 14.06 & 14.22 & 14.27 \\
			Real-ESRGAN                           & 10.15 & 10.14 & 10.12 & 10.10 & 10.14 & 10.18 & 10.22 & 11.05 & 10.98 & 10.64 & 11.18 & 11.00 & 10.35 & 10.31 & 10.33 & 10.37 & 10.48 \\
			Real-ESRNet                           & 13.76 & 13.80 & 13.87 & 13.95 & 14.16 & 14.12 & 14.35 & 16.30 & 16.06 & 16.58 & 16.84 & 16.97 & 17.02 & 17.10 & 17.61 & 18.12 & 18.36 \\
			BSRGAN                                & 9.77  & 9.77  & 9.76  & 9.72  & 9.75  & 9.80  & 9.81  & 9.97  & 10.16 & 10.39 & 10.52 & 10.63 & 10.74 & 10.73 & 10.78 & 10.64 & 10.51 \\
			BSRNet                                & 14.11 & 14.15 & 14.38 & 14.64 & 14.98 & 15.35 & 15.90 & 16.60 & 17.72 & 19.28 & 20.25 & 20.00 & 19.59 & 18.86 & 18.50 & 18.36 & 18.28 \\
			SwinIR-GAN                            & 9.83  & 9.83  & 9.84  & 9.91  & 9.89  & 9.94  & 10.05 & 10.17 & 10.08 & 10.11 & 10.10 & 10.18 & 10.07 & 10.15 & 10.16 & 10.22 & 10.24 \\
			SwinIR-PSNR                           & 13.48 & 13.55 & 13.75 & 13.97 & 14.18 & 14.44 & 14.75 & 15.13 & 15.48 & 15.80 & 16.15 & 16.47 & 16.53 & 16.75 & 16.88 & 16.95 & 16.80 \\ \hline
	\end{tabular}}
	
\end{table*}

\begin{table*}[]
	\centering
	\caption{Model performance (PSNR) on PIES-Noise dataset. Higher PSNR value denotes that the output images are closer to the ground truth images in content.}
	\label{tab:noise_psnr}
	\resizebox{\textwidth}{20mm}{
		
		\begin{tabular}{lccccccccccc}
			\hline
			\multicolumn{1}{c}{\multirow{2}{*}{Methods}} & 
			\multicolumn{11}{c}{Noise \quad Level}\\ \cline{3-12} 
			
			& Clean & 5 & 10 & 15 & 20 & 25 & 30 & 35 & 40 & 45 & 50 \\ \hline
			SRResNet (train: clean)               & 25.87 & 23.92 & 21.66 & 19.66 & 17.95 & 16.52 & 15.35 & 14.39 & 13.61 & 12.96 & 12.42 \\
			SRResNet (train: noise0\_20)          & 25.59 & 25.02 & 24.32 & 23.67 & 23.11 & 22.47 & 20.94 & 18.62 & 16.70 & 15.30 & 14.27 \\
			SRResNet (train: blur0\_4+noise0\_20) & 25.10 & 24.64 & 24.04 & 23.45 & 22.94 & 22.32 & 21.47 & 20.52 & 19.68 & 19.02 & 18.56 \\
			Real-ESRGAN                           & 21.90 & 21.81 & 21.45 & 20.95 & 20.49 & 20.06 & 19.64 & 19.26 & 18.93 & 18.63 & 18.38 \\
			Real-ESRNet                           & 23.41 & 23.31 & 22.87 & 22.32 & 21.81 & 21.35 & 20.90 & 20.45 & 20.04 & 19.58 & 19.17 \\
			BSRGAN                                & 22.54 & 22.14 & 21.16 & 19.94 & 18.77 & 17.72 & 16.81 & 16.01 & 15.35 & 14.75 & 14.20 \\
			BSRNet                                & 23.79 & 23.32 & 22.18 & 20.95 & 19.80 & 18.78 & 17.95 & 17.28 & 16.73 & 16.26 & 15.83 \\
			SwinIR-GAN                            & 22.06 & 21.84 & 21.27 & 20.63 & 20.01 & 19.47 & 18.96 & 18.52 & 18.09 & 17.76 & 17.42 \\
			SwinIR-PSNR                           & 23.62 & 23.40 & 22.75 & 22.00 & 21.24 & 20.47 & 19.71 & 19.00 & 18.31 & 17.72 & 17.13   \\ \hline
	\end{tabular}}
	
\end{table*}

\begin{table*}[]
	
	\centering
	\caption{Model performance (SSIM) on PIES-Noise dataset. Higher SSIM value denotes that the output images are structurally closer to the ground truth images.}
	\label{tab:noise_ssim}
	\resizebox{\textwidth}{20mm}{
		
		\begin{tabular}{lccccccccccc}
			\hline
			\multicolumn{1}{c}{\multirow{2}{*}{Methods}} & 
			\multicolumn{11}{c}{Noise \quad Level}\\ \cline{3-12} 
			& Clean & 5 & 10 & 15 & 20 & 25 & 30 & 35 & 40 & 45 & 50 \\ \hline
			SRResNet (train: clean)               & 0.7569 & 0.6394 & 0.4964 & 0.3920 & 0.3173 & 0.2635 & 0.2242 & 0.1948 & 0.1729 & 0.1550 & 0.1409 \\
			SRResNet (train: noise0\_20)          & 0.7473 & 0.7199 & 0.6876 & 0.6593 & 0.6357 & 0.6130 & 0.5379 & 0.4223 & 0.3317 & 0.2661 & 0.2199 \\
			SRResNet (train: blur0\_4+noise0\_20) & 0.7377 & 0.7159 & 0.6873 & 0.6613 & 0.6399 & 0.6149 & 0.5724 & 0.5225 & 0.4780 & 0.4434 & 0.4209 \\
			Real-ESRGAN                           & 0.6271 & 0.6140 & 0.5899 & 0.5649 & 0.5419 & 0.5245 & 0.5065 & 0.4898 & 0.4760 & 0.4633 & 0.4512 \\
			Real-ESRNet                           & 0.6795 & 0.6668 & 0.6402 & 0.6135 & 0.5891 & 0.5699 & 0.5518 & 0.5344 & 0.5212 & 0.5069 & 0.4934 \\
			BSRGAN                                & 0.6321 & 0.6076 & 0.5640 & 0.5151 & 0.4666 & 0.4220 & 0.3828 & 0.3494 & 0.3222 & 0.2996 & 0.2801 \\
			BSRNet                                & 0.6814 & 0.6573 & 0.6122 & 0.5657 & 0.5225 & 0.4860 & 0.4549 & 0.4286 & 0.4076 & 0.3903 & 0.3750 \\
			SwinIR-GAN                            & 0.6321 & 0.6137 & 0.5792 & 0.5455 & 0.5162 & 0.4909 & 0.4664 & 0.4432 & 0.4206 & 0.4022 & 0.3823 \\
			SwinIR-PSNR                           & 0.6922 & 0.6752 & 0.6393 & 0.6050 & 0.5741 & 0.5462 & 0.5187 & 0.4935 & 0.4698 & 0.4487 & 0.4271                    \\ \hline
	\end{tabular}}
	
\end{table*}

\begin{table*}[htbp]
	
	\centering
	\caption{Model performance (LPIPS) on PIES-Noise dataset. Lower LPIPS value denotes that the output images are perceptually closer to the ground truth images.}
	\label{tab:noise_lpips}
	\resizebox{\textwidth}{20mm}{
		
		\begin{tabular}{lccccccccccc}
			\hline
			\multicolumn{1}{c}{\multirow{2}{*}{Methods}} & 
			\multicolumn{11}{c}{Noise \quad Level}\\ \cline{3-12} 
			& Clean & 5 & 10 & 15 & 20 & 25 & 30 & 35 & 40 & 45 & 50 \\ \hline
			SRResNet (train: clean)               & 0.2136 & 0.3226 & 0.4664 & 0.5602 & 0.6133 & 0.6482 & 0.6710 & 0.6888 & 0.7032 & 0.7159 & 0.7267 \\
			SRResNet (train: noise0\_20)          & 0.2184 & 0.2391 & 0.2648 & 0.2870 & 0.3048 & 0.3080 & 0.3636 & 0.4646 & 0.5372 & 0.5877 & 0.6236 \\
			SRResNet (train: blur0\_4+noise0\_20) & 0.2299 & 0.2441 & 0.2709 & 0.2961 & 0.3155 & 0.3231 & 0.3418 & 0.3649 & 0.3877 & 0.4059 & 0.4199 \\
			Real-ESRGAN                           & 0.1932 & 0.1974 & 0.2117 & 0.2300 & 0.2466 & 0.2648 & 0.2825 & 0.3008 & 0.3173 & 0.3329 & 0.3496 \\
			Real-ESRNet                           & 0.2843 & 0.2938 & 0.3158 & 0.3390 & 0.3587 & 0.3752 & 0.3892 & 0.4013 & 0.4125 & 0.4213 & 0.4288 \\
			BSRGAN                                & 0.1909 & 0.1995 & 0.2279 & 0.2657 & 0.3020 & 0.3314 & 0.3576 & 0.3820 & 0.4036 & 0.4246 & 0.4434 \\
			BSRNet                                & 0.2790 & 0.2907 & 0.3203 & 0.3524 & 0.3830 & 0.4079 & 0.4304 & 0.4509 & 0.4696 & 0.4875 & 0.5019 \\
			DASR-N-Ani                            & 0.2248 & 0.2429 & 0.2674 & 0.2937 & 0.3149 & 0.3301 & 0.3371 & 0.3519 & 0.3797 & 0.4057 & 0.4293 \\
			SwinIR-GAN                            & 0.1724 & 0.1779 & 0.2009 & 0.2309 & 0.2566 & 0.2777 & 0.2925 & 0.3116 & 0.3261 & 0.3401 & 0.3570 \\
			SwinIR-PSNR                           & 0.2671 & 0.2759 & 0.3049 & 0.3320 & 0.3538 & 0.3722 & 0.3879 & 0.4013 & 0.4142 & 0.4225 & 0.4333                    \\ \hline
	\end{tabular}}
	
\end{table*}

\begin{table*}[htbp]
	
	\centering
	\caption{Model performance (NIQE) on PIES-Noise dataset. Lower NIQE value denotes better perceptual quality.}
	\label{tab:noise_niqe}
	\resizebox{\textwidth}{20mm}{
		
		\begin{tabular}{lccccccccccc}
			\hline
			\multicolumn{1}{c}{\multirow{2}{*}{Methods}} & 
			\multicolumn{11}{c}{Noise \quad Level}\\ \cline{3-12} 
			& Clean & 5 & 10 & 15 & 20 & 25 & 30 & 35 & 40 & 45 & 50 \\ \hline
			SRResNet (train: clean)               & 12.28 & 13.07 & 13.25 & 13.38 & 13.52 & 13.71 & 14.14 & 14.36 & 14.84 & 15.36 & 16.08 \\
			SRResNet (train: noise0\_20)          & 12.54 & 12.95 & 13.08 & 13.33 & 13.37 & 12.49 & 12.04 & 13.39 & 15.41 & 16.63 & 17.65 \\
			SRResNet (train: blur0\_4+noise0\_20) & 12.76 & 13.05 & 13.23 & 13.64 & 13.84 & 14.03 & 14.58 & 14.78 & 15.00 & 14.78 & 13.94 \\
			Real-ESRGAN                           & 10.15 & 10.16 & 10.31 & 10.48 & 10.61 & 10.74 & 10.72 & 10.76 & 10.88 & 10.86 & 11.07 \\
			Real-ESRNet                           & 13.76 & 13.76 & 13.98 & 14.32 & 14.39 & 14.28 & 14.30 & 14.05 & 13.99 & 13.62 & 13.25 \\
			BSRGAN                                & 9.77  & 9.88  & 10.02 & 10.45 & 10.79 & 11.11 & 11.55 & 11.85 & 12.70 & 13.05 & 12.60 \\
			BSRNet                                & 14.11 & 13.99 & 14.26 & 14.65 & 14.78 & 14.79 & 14.83 & 14.86 & 14.76 & 14.81 & 14.33 \\
			DASR-N-Ani                            & 12.12 & 12.68 & 13.14 & 14.01 & 14.34 & 14.38 & 14.10 & 14.23 & 14.10 & 14.04 & 13.74 \\
			SwinIR-GAN                            & 9.83  & 9.82  & 9.84  & 9.93  & 9.79  & 9.59  & 9.38  & 9.14  & 8.95  & 8.77  & 8.67  \\
			SwinIR-PSNR                           & 13.48 & 13.49 & 13.86 & 13.94 & 13.83 & 13.52 & 13.35 & 12.86 & 12.46 & 12.08 & 11.70                    \\ \hline
	\end{tabular}}
	
\end{table*}

\clearpage

\begin{table*}[htbp]
	\centering
	\caption{Model performance (PSNR) on PIES-BlurNoise dataset. Higher PSNR value denotes that the output images are closer to the ground truth images in content. n10 represents noise level 10.}
	\label{tab:blurnoise_psnr}
	\resizebox{\textwidth}{20mm}{
		
		\begin{tabular}{lccccccccccccc}
			\hline
			\multicolumn{1}{c}{\multirow{2}{*}{Methods}} &       & \multicolumn{3}{c}{Blur1} & \multicolumn{3}{c}{Blur2} & \multicolumn{3}{c}{Blur4} & \multicolumn{3}{c}{Blur6} \\ \cline{3-14} 
			\multicolumn{1}{c}{}                         & clean & n10   & n20  & n30  & n10   & n20  & n30  & n10   & n20  & n30  & n10   & n20  & n30  \\ \hline
			SRResNet(train: clean)                       & 25.87 & 21.50   & 17.93  & 15.32  & 20.58   & 17.55  & 15.09  & 18.91   & 16.61  & 14.48  & 18.13   & 16.11  & 14.14  \\
			SRResNet (train: blur0\_4)                   & 25.48 & 21.68   & 18.12  & 15.66  & 20.74   & 17.73  & 15.40  & 19.03   & 16.77  & 14.75  & 18.23   & 16.25  & 14.39  \\
			SRResNet (train: noise0\_20)                 & 25.59 & 23.88   & 22.80  & 20.85  & 22.12   & 21.56  & 20.18  & 19.91   & 19.65  & 18.72  & 19.01   & 18.83  & 18.01  \\
			SRResNet (train: blur0\_4+noise0\_20)        & 25.10 & 23.90   & 22.75  & 21.48  & 23.14   & 22.10  & 20.98  & 21.38   & 20.47  & 19.39  & 19.55   & 19.18  & 18.56  \\
			Real-ESRGAN                                  & 21.90 & 21.45   & 20.45  & 19.60  & 21.25   & 20.22  & 19.38  & 20.12   & 19.31  & 18.78  & 19.33   & 18.76  & 18.37  \\
			Real-ESRNet                                  & 23.41 & 22.82   & 21.72  & 20.81  & 22.49   & 21.36  & 20.48  & 21.00   & 20.15  & 19.53  & 20.08   & 19.43  & 18.92  \\
			BSRGAN                                       & 22.54 & 21.13   & 18.72  & 16.73  & 20.93   & 18.52  & 16.57  & 19.99   & 17.78  & 16.06  & 19.20   & 17.30  & 15.75  \\
			BSRNet                                       & 23.79 & 22.11   & 19.71  & 17.87  & 21.79   & 19.43  & 17.68  & 20.64   & 18.61  & 17.13  & 19.74   & 18.08  & 16.79  \\
			SwinIR-GAN                                   & 22.06 & 21.20   & 19.91  & 18.83  & 20.81   & 19.51  & 18.46  & 19.49   & 18.34  & 17.45  & 18.44   & 17.55  & 16.92  \\
			SwinIR-PSNR                                  & 23.62 & 22.63   & 21.11  & 19.58  & 22.19   & 20.68  & 19.23  & 20.78   & 19.50  & 18.28  & 19.77   & 18.78  & 17.72   \\ \hline
	\end{tabular}}
	
\end{table*}

\begin{table*}[htbp]
	\centering
	\caption{Model performance (SSIM) on PIES-BlurNoise dataset. Higher SSIM value denotes that the output images are structurally closer to the ground truth images.}
	\label{tab:blurnoise_ssim}
	\resizebox{\textwidth}{20mm}{
		
		\begin{tabular}{lccccccccccccc}
			\hline
			\multicolumn{1}{c}{\multirow{2}{*}{Methods}} &       & \multicolumn{3}{c}{Blur1} & \multicolumn{3}{c}{Blur2} & \multicolumn{3}{c}{Blur4} & \multicolumn{3}{c}{Blur6} \\ \cline{3-14} 
			\multicolumn{1}{c}{}                         & clean & n10   & n20  & n30  & n10   & n20  & n30  & n10   & n20  & n30  & n10   & n20  & n30  \\ \hline
			SRResNet (train: clean)                & 0.7569 & 0.4762  & 0.3020  & 0.2110  & 0.4122  & 0.2574  & 0.1779  & 0.3087  & 0.1839  & 0.1235  & 0.2698  & 0.1557  & 0.1032  \\
			SRResNet (train: blur0\_4)            & 0.7495 & 0.4926  & 0.3160  & 0.2256  & 0.4286  & 0.2709  & 0.1909  & 0.3233  & 0.1949  & 0.1337  & 0.2834  & 0.1655  & 0.1120  \\
			SRResNet (train: noise0\_20)          & 0.7473 & 0.6623  & 0.6144  & 0.5227  & 0.5821  & 0.5524  & 0.4744  & 0.4718  & 0.4591  & 0.3859  & 0.4313  & 0.4237  & 0.3484  \\
			SRResNet (train: blur0\_4+noise0\_20) & 0.7377 & 0.6719  & 0.6238  & 0.5630  & 0.6327  & 0.5853  & 0.5257  & 0.5388  & 0.4992  & 0.4374  & 0.4546  & 0.4411  & 0.3979  \\
			Real-ESRGAN                           & 0.6271 & 0.5834  & 0.5346  & 0.4991  & 0.5605  & 0.5124  & 0.4781  & 0.4925  & 0.4565  & 0.4364  & 0.4517  & 0.4284  & 0.4149  \\
			Real-ESRNet                           & 0.6795 & 0.6323  & 0.5795  & 0.5418  & 0.6060  & 0.5523  & 0.5159  & 0.5285  & 0.4889  & 0.4633  & 0.4856  & 0.4569  & 0.4359  \\
			BSRGAN                                & 0.6321 & 0.5583  & 0.4611  & 0.3773  & 0.5382  & 0.4419  & 0.3637  & 0.4769  & 0.3968  & 0.3324  & 0.4352  & 0.3719  & 0.3163  \\
			BSRNet                                & 0.6814 & 0.6047  & 0.5143  & 0.4478  & 0.5808  & 0.4927  & 0.4316  & 0.5134  & 0.4436  & 0.3966  & 0.4676  & 0.4160  & 0.3789  \\
			SwinIR-GAN                            & 0.6321 & 0.5722  & 0.5082  & 0.4564  & 0.5436  & 0.4828  & 0.4342  & 0.4714  & 0.4225  & 0.3809  & 0.4214  & 0.3844  & 0.3535  \\
			SwinIR-PSNR                           & 0.6922 & 0.6305  & 0.5642  & 0.5085  & 0.6015  & 0.5369  & 0.4839  & 0.5305  & 0.4768  & 0.4324  & 0.4870  & 0.4442  & 0.4060   \\ \hline
	\end{tabular}}
	
\end{table*}

\begin{table*}[htbp]
	\centering
	\caption{Model performance (LPIPS) on PIES-BlurNoise dataset. Lower LPIPS value denotes that the output images are perceptually closer to the ground truth images.}
	\label{tab:blurnoise_lpips}
	\resizebox{\textwidth}{20mm}{
		
		\begin{tabular}{lccccccccccccc}
			\hline
			\multicolumn{1}{c}{\multirow{2}{*}{Methods}} &       & \multicolumn{3}{c}{Blur1} & \multicolumn{3}{c}{Blur2} & \multicolumn{3}{c}{Blur4} & \multicolumn{3}{c}{Blur6} \\ \cline{3-14} 
			\multicolumn{1}{c}{}                         & clean & n10   & n20  & n30  & n10   & n20  & n30  & n10   & n20  & n30  & n10   & n20  & n30  \\ \hline
			SRResNet (train: clean)                & 0.2136 & 0.5064  & 0.6379 & 0.6815 & 0.5853  & 0.6773 & 0.7000 & 0.6587  & 0.7054 & 0.7092 & 0.6741  & 0.7093 & 0.7097 \\
			SRResNet (train: blur0\_4)            & 0.2202 & 0.4911  & 0.6363 & 0.6858 & 0.5705  & 0.6761 & 0.7058 & 0.6508  & 0.7078 & 0.7181 & 0.6674  & 0.7133 & 0.7176 \\
			SRResNet (train: noise0\_20)          & 0.2184 & 0.3043  & 0.3384 & 0.3842 & 0.4369  & 0.4397 & 0.4410 & 0.6027  & 0.5888 & 0.5157 & 0.6530  & 0.6388 & 0.5344 \\
			SRResNet (train: blur0\_4+noise0\_20) & 0.2299 & 0.2944  & 0.3396 & 0.3564 & 0.3420  & 0.3883 & 0.4009 & 0.4691  & 0.5149 & 0.4886 & 0.6088  & 0.6092 & 0.5184 \\
			Real-ESRGAN                           & 0.1932 & 0.2175  & 0.2572 & 0.2937 & 0.2360  & 0.2798 & 0.3224 & 0.3080  & 0.3604 & 0.4100 & 0.3623  & 0.4240 & 0.4818 \\
			Real-ESRNet                           & 0.2843 & 0.3268  & 0.3729 & 0.4037 & 0.3599  & 0.4088 & 0.4409 & 0.4505  & 0.4901 & 0.5202 & 0.5049  & 0.5379 & 0.5666 \\
			BSRGAN                                & 0.1909 & 0.2329  & 0.3071 & 0.3647 & 0.2506  & 0.3266 & 0.3822 & 0.3127  & 0.3816 & 0.4257 & 0.3668  & 0.4163 & 0.4538 \\
			BSRNet                                & 0.2790 & 0.3303  & 0.3945 & 0.4414 & 0.3601  & 0.4242 & 0.4667 & 0.4490  & 0.4942 & 0.5249 & 0.5155  & 0.5349 & 0.5566 \\
			SwinIR-GAN                            & 0.1724 & 0.2080  & 0.2654 & 0.3031 & 0.2305  & 0.2855 & 0.3220 & 0.2870  & 0.3318 & 0.3674 & 0.3218  & 0.3610 & 0.3919 \\
			SwinIR-PSNR                           & 0.2671 & 0.3158  & 0.3684 & 0.4027 & 0.3476  & 0.4004 & 0.4366 & 0.4243  & 0.4685 & 0.5036 & 0.4718  & 0.5072 & 0.5371   \\ \hline
	\end{tabular}}
	
\end{table*}

\begin{table*}[htbp]
	\centering
	\caption{Model performance (NIQE) on PIES-BlurNoise dataset. Lower NIQE value denotes better perceptual quality.}
	\label{tab:blurnoise_niqe}
	\resizebox{\textwidth}{20mm}{
		
		\begin{tabular}{lccccccccccccc}
			\hline
			\multicolumn{1}{c}{\multirow{2}{*}{Methods}} &       & \multicolumn{3}{c}{Blur1} & \multicolumn{3}{c}{Blur2} & \multicolumn{3}{c}{Blur4} & \multicolumn{3}{c}{Blur6} \\ \cline{3-14} 
			\multicolumn{1}{c}{}                         & clean & n10   & n20  & n30  & n10   & n20  & n30  & n10   & n20  & n30  & n10   & n20  & n30  \\ \hline
			SRResNet (train: clean)                       & 12.28 & 12.21   & 12.23  & 13.18  & 8.86    & 10.30  & 11.80  & 7.45    & 9.86   & 11.80  & 7.54    & 10.26  & 12.24  \\
			SRResNet (train: blur0\_4)                   & 12.20 & 11.91   & 12.06  & 12.68  & 8.80    & 10.27  & 11.56  & 7.17    & 9.65   & 11.60  & 7.22    & 10.10  & 12.04  \\
			SRResNet (train: noise0\_20)                 & 12.54 & 13.80   & 13.91  & 11.98  & 15.93   & 16.64  & 11.87  & 15.65   & 15.52  & 9.31   & 16.36   & 15.84  & 9.21   \\
			SRResNet (train: blur0\_4+noise0\_20)        & 12.76 & 13.60   & 14.14  & 13.95  & 14.38   & 14.70  & 13.21  & 17.03   & 18.15  & 10.72  & 16.73   & 15.68  & 10.32  \\
			Real-ESRGAN                                  & 10.15 & 10.24   & 10.73  & 10.83  & 10.30   & 10.57  & 10.76  & 10.94   & 11.46  & 11.37  & 11.05   & 11.36  & 12.11  \\
			Real-ESRNet                                  & 13.76 & 14.31   & 14.83  & 14.56  & 14.83   & 15.26  & 15.26  & 17.68   & 16.95  & 15.92  & 18.28   & 17.14  & 17.08  \\
			BSRGAN                                       & 9.77  & 10.05   & 10.86  & 11.70  & 10.10   & 10.78  & 11.78  & 10.93   & 11.13  & 11.81  & 10.59   & 11.13  & 11.94  \\
			BSRNet                                       & 14.11 & 14.57   & 15.25  & 15.15  & 15.16   & 15.76  & 15.71  & 18.58   & 17.13  & 17.22  & 19.37   & 19.62  & 18.58  \\
			SwinIR-GAN                                   & 9.83  & 10.00   & 10.06  & 9.36   & 10.09   & 9.84   & 9.39   & 10.17   & 9.72   & 9.19   & 10.05   & 9.83   & 9.14   \\
			SwinIR-PSNR                                  & 13.48 & 14.14   & 14.14  & 13.50  & 14.66   & 14.75  & 14.13  & 16.09   & 16.03  & 14.92  & 17.03   & 16.58  & 16.18   \\ \hline
	\end{tabular}}
	
\end{table*}

\begin{table*}
	\centering
	\caption{Model performance (NIQE) on PIES-RealCam and PIES-RealLQ datasets. Lower NIQE value denotes better perceptual quality.}
	\label{tab:real}
	\begin{tabular}{lcc}
		\hline
		Methods                                     & PIES800-RealCam & PIES800-RealLQ \\ \hline
		SRResNet (train: clean)               & 22.40           & 14.40          \\
		SRResNet (train: blur0\_4)            & 22.40           & 14.03          \\
		SRResNet (train: noise0\_20)          & 23.71           & 15.67          \\
		SRResNet (train: blur0\_4+noise0\_20) & 23.63           & 14.02          \\
		IKC                                         & 21.60           & 13.77          \\
		DAN                                         & 22.86           & 14.00          \\
		DASR                                 & 22.74           & 13.44                     \\
		Real-ESRGAN                                 & 18.94           & 13.53          \\
		Real-ESRNet                                 & 26.86           & 18.02          \\
		BSRGAN                                      & 20.30           & 13.09          \\
		BSRNet                                      & 32.71           & 18.39          \\
		SwinIR-GAN                                  & 17.75           & 12.84          \\
		SwinIR-PSNR                                 & 27.82           & 17.78         \\ \hline
	\end{tabular}
	
\end{table*}

\linespread{1.2}

\begin{table*}
	\centering
	\caption{The estimated GGD parameters $\sigma$ and $\alpha$ of representative methods with different degraded input.}
	\label{tab:GGD}
	\resizebox{\textwidth}{50mm}{

		\begin{tabular}{llccccccccccccccccc}
			\hline
			\multicolumn{2}{c}{\multirow{2}{*}{Methods}}               & \multicolumn{17}{c}{Blur \quad Level}                                                                                                       \\ \cline{4-19} 
			\multicolumn{2}{c}{}                                    & Clean & 0.5   & 1.0   & 1.5   & 2.0   & 2.5   & 3.0   & 3.5   & 4.0     & 4.5   & 5.0   & 5.5   & 6.0   & 6.5   & 7.0   & 7.5   & 8.0   \\ \hline
			\multirow{2}{*}{SRResNet (train: clean)}               & $\sigma$ & 2.718 & 2.668 & 2.532 & 2.418 & 2.333 & 2.260 & 2.194 & 2.138 & 2.083 & 2.042 & 2.007 & 1.983 & 1.965 & 1.948 & 1.933 & 1.920 & 1.911 \\ 
			& $\alpha$ & 0.687 & 0.684 & 0.661 & 0.628 & 0.596 & 0.568 & 0.539 & 0.514 & 0.494 & 0.478 & 0.466 & 0.454 & 0.446 & 0.440 & 0.436 & 0.432 & 0.428 \\ \hline
			\multirow{2}{*}{SRResNet (train: blur0\_4)}            & $\sigma$ & 2.866 & 2.840 & 2.783 & 2.787 & 2.795 & 2.785 & 2.735 & 2.599 & 2.465 & 2.370 & 2.299 & 2.245 & 2.203 & 2.170 & 2.142 & 2.122 & 2.101 \\ 
			& $\alpha$ & 0.691 & 0.686 & 0.682 & 0.683 & 0.683 & 0.682 & 0.667 & 0.632 & 0.591 & 0.555 & 0.526 & 0.505 & 0.489 & 0.478 & 0.468 & 0.461 & 0.457 \\ \hline
			\multirow{2}{*}{IKC}                                   & $\sigma$  & 1.291 & 1.245 & 1.133 & 1.062 & 1.048 & 1.049 & 1.068 & 1.082 & 1.100 & 1.066 & 0.977 & 0.895 & 0.834 & 0.800 & 0.781 & 0.768 & 0.760 \\
			& $\alpha$  & 0.770 & 0.774 & 0.778 & 0.774 & 0.779 & 0.786 & 0.798 & 0.795 & 0.787 & 0.764 & 0.718 & 0.678 & 0.643 & 0.619 & 0.604 & 0.595 & 0.588 \\ \hline
			\multirow{2}{*}{DAN}                                   & $\sigma$  & 4.278 & 4.152 & 4.025 & 3.986 & 3.849 & 3.610 & 3.273 & 2.902 & 2.669 & 2.542 & 2.465 & 2.404 & 2.357 & 2.317 & 2.291 & 2.276 & 2.264 \\
			& $\alpha$  & 0.668 & 0.675 & 0.673 & 0.680 & 0.690 & 0.695 & 0.689 & 0.660 & 0.605 & 0.559 & 0.529 & 0.506 & 0.488 & 0.474 & 0.463 & 0.456 & 0.452 \\ \hline
			\multirow{2}{*}{RealESRGAN}                            & $\sigma$  & 5.022 & 5.007 & 4.952 & 4.879 & 4.794 & 4.728 & 4.645 & 4.575 & 4.513 & 4.452 & 4.393 & 4.348 & 4.312 & 4.270 & 4.248 & 4.234 & 4.230 \\
			& $\alpha$  & 0.790 & 0.788 & 0.784 & 0.781 & 0.781 & 0.775 & 0.778 & 0.776 & 0.776 & 0.772 & 0.768 & 0.761 & 0.758 & 0.757 & 0.755 & 0.750 & 0.747 \\ \hline
			\multirow{2}{*}{RealESRNet}                            & $\sigma$  & 3.718 & 3.707 & 3.678 & 3.645 & 3.613 & 3.579 & 3.562 & 3.550 & 3.512 & 3.462 & 3.421 & 3.351 & 3.269 & 3.251 & 3.233 & 3.213 & 3.197 \\
			& $\alpha$  & 0.739 & 0.739 & 0.738 & 0.734 & 0.725 & 0.721 & 0.717 & 0.708 & 0.699 & 0.683 & 0.673 & 0.659 & 0.656 & 0.657 & 0.661 & 0.663 & 0.662 \\ \hline
			\multirow{2}{*}{BSRGAN}                                & $\sigma$  & 8.921 & 8.900 & 8.814 & 8.689 & 8.531 & 8.360 & 8.183 & 8.026 & 7.888 & 7.781 & 7.695 & 7.615 & 7.553 & 7.493 & 7.456 & 7.409 & 7.373 \\
			& $\alpha$  & 0.762 & 0.762 & 0.760 & 0.756 & 0.749 & 0.744 & 0.737 & 0.726 & 0.720 & 0.710 & 0.701 & 0.694 & 0.685 & 0.678 & 0.671 & 0.668 & 0.662 \\ \hline
			\multirow{2}{*}{BSRNet}                                & $\sigma$  & 5.339 & 5.328 & 5.289 & 5.234 & 5.165 & 5.092 & 5.015 & 4.939 & 4.857 & 4.788 & 4.734 & 4.678 & 4.631 & 4.584 & 4.552 & 4.527 & 4.510 \\
			& $\alpha$  & 0.693 & 0.693 & 0.689 & 0.684 & 0.677 & 0.668 & 0.659 & 0.647 & 0.635 & 0.622 & 0.608 & 0.594 & 0.581 & 0.571 & 0.562 & 0.557 & 0.554 \\ \hline
			\multirow{2}{*}{DASR}                                  & $\sigma$  & 6.416 & 6.293 & 6.107 & 6.058 & 5.948 & 5.747 & 5.445 & 5.087 & 4.691 & 4.342 & 4.155 & 4.020 & 3.908 & 3.837 & 3.773 & 3.747 & 3.713 \\
			& $\alpha$  & 0.679 & 0.682 & 0.683 & 0.686 & 0.693 & 0.705 & 0.715 & 0.727 & 0.704 & 0.649 & 0.603 & 0.571 & 0.551 & 0.537 & 0.524 & 0.518 & 0.513 \\ \hline
			\multirow{2}{*}{SRResNet (train: blur0\_4+noise0\_20)} & $\sigma$  & 3.171 & 3.137 & 3.064 & 3.059 & 3.100 & 3.084 & 3.020 & 2.893 & 2.735 & 2.612 & 2.529 & 2.477 & 2.435 & 2.406 & 2.380 & 2.358 & 2.342 \\
			& $\alpha$  & 0.696 & 0.694 & 0.682 & 0.683 & 0.689 & 0.687 & 0.678 & 0.646 & 0.595 & 0.552 & 0.521 & 0.500 & 0.486 & 0.475 & 0.469 & 0.463 & 0.459 \\ \hline
			\multirow{2}{*}{SwinIR-GAN}                            & $\sigma$  & 5.178 & 5.164 & 5.140 & 5.110 & 5.069 & 5.036 & 5.002 & 4.965 & 4.929 & 4.903 & 4.886 & 4.884 & 4.891 & 4.906 & 4.926 & 4.941 & 4.953 \\
			& $\alpha$  & 0.740 & 0.742 & 0.742 & 0.742 & 0.742 & 0.736 & 0.734 & 0.732 & 0.733 & 0.736 & 0.738 & 0.739 & 0.738 & 0.735 & 0.733 & 0.734 & 0.735 \\ \hline
			\multirow{2}{*}{SwinIR-PSNR}                           & $\sigma$  & 4.279 & 4.275 & 4.267 & 4.254 & 4.239 & 4.218 & 4.184 & 4.145 & 4.107 & 4.067 & 4.046 & 4.025 & 3.997 & 3.983 & 3.987 & 3.986 & 3.991 \\
			& $\alpha$  & 0.747 & 0.746 & 0.746 & 0.744 & 0.740 & 0.737 & 0.737 & 0.736 & 0.739 & 0.741 & 0.740 & 0.735 & 0.732 & 0.728 & 0.722 & 0.719 & 0.717 \\ \hline
	\end{tabular}}

\end{table*}

\begin{figure*}[htbp]
	\begin{center}
		\includegraphics[width=0.95\linewidth]{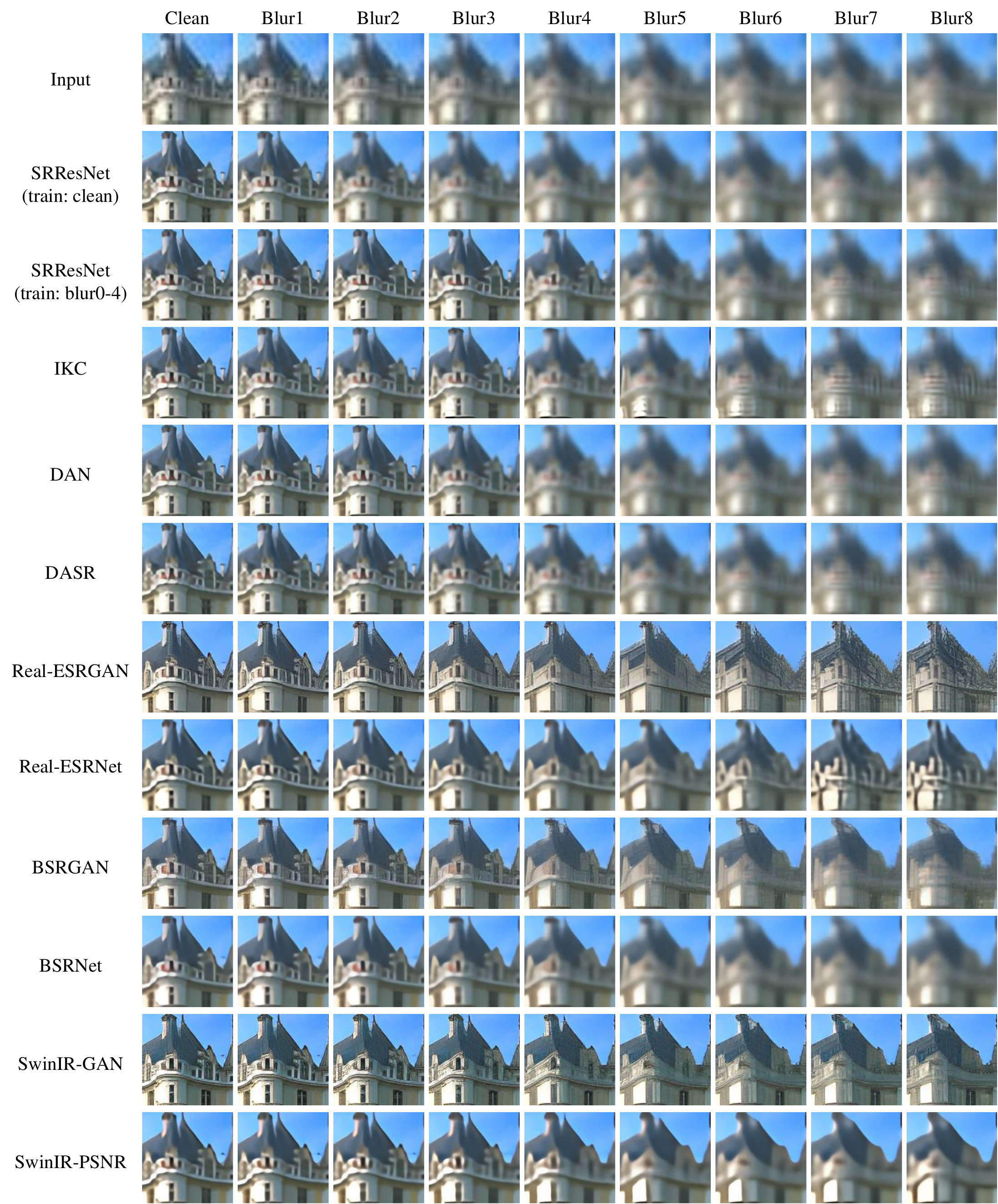}
	\end{center}
	\caption{Visual comparison on PIES-Blur dataset (1).}
	\label{fig:blur_img1}
\end{figure*}

\begin{figure*}[htbp]
	\begin{center}
		\includegraphics[width=0.9\linewidth,height=1.2\linewidth]{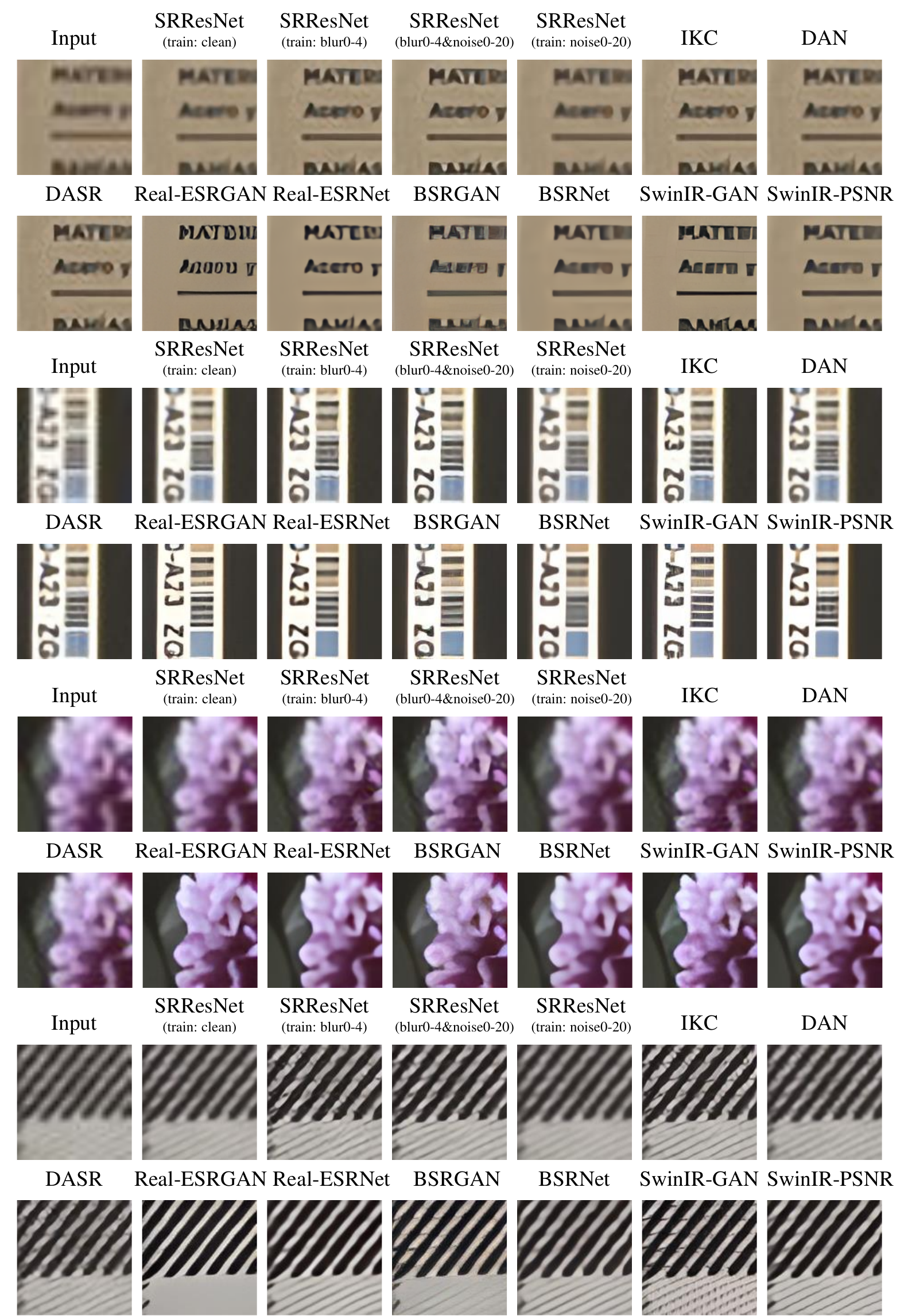}
	\end{center}
	\caption{Visual comparison on PIES-RealCam dataset.}
	\label{fig:realcam_img}
\end{figure*}

\begin{figure*}[htbp]
	\begin{center}
		\includegraphics[width=0.9\linewidth,height=1.2\linewidth]{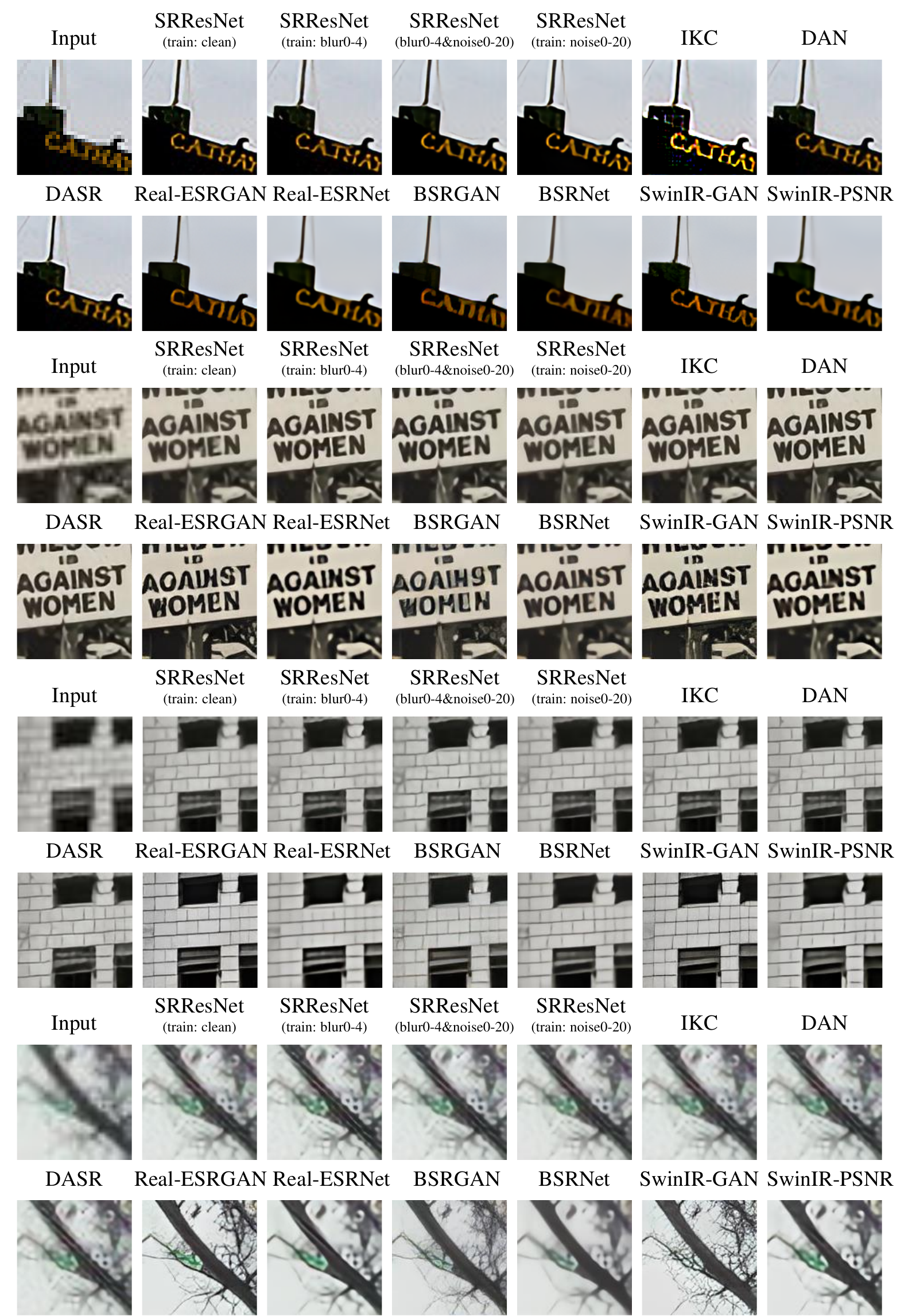}
	\end{center}
	\caption{Visual comparison on PIES-RealLQ dataset.}
	\label{fig:reallq_img}
\end{figure*}

%





\ifCLASSOPTIONcaptionsoff
  \newpage
\fi

\end{document}